\documentclass[aps, pra, reprint, amsmath, amssymb, superscriptaddress, onecolumn, longbibliography, nofootinbib, notitlepage]{revtex4-2}
\usepackage{mathtools}
\usepackage{siunitx}
\usepackage{svg}
\usepackage{caption}
\usepackage{float}
\usepackage{subcaption}
\usepackage[braket, qm]{qcircuit}
\usepackage{bbm}
\usepackage{braket}
\usepackage{physics}
\usepackage{lipsum} % for placeholder text
\usepackage{chngcntr}
\usepackage{booktabs}
\usepackage{nicematrix}
\usepackage{diagbox}
\usepackage{titletoc}
\usepackage{algorithm}
\usepackage{listings}
\usepackage{xcolor} % You already have this!
\usepackage{xcolor}

\usepackage[T1]{fontenc}
\DeclareMathOperator{\sinc}{sinc}
\AtBeginDocument{\RenewCommandCopy\qty\SI}
\usepackage[hidelinks]{hyperref}

\usepackage{xcolor}

\begin{document}
\title{Ultra-low-light computer vision using trained photon correlations}
\author{Mandar~M.~Sohoni}
\email{Equal contribution}
\affiliation{School of Applied and Engineering Physics, Cornell University, Ithaca, NY 14853, USA}
\author{J\'{e}r\'{e}mie~Laydevant}
\email{Equal contribution}
\affiliation{School of Applied and Engineering Physics, Cornell University, Ithaca, NY 14853, USA}
\author{Mathieu~Ouellet}
\affiliation{School of Applied and Engineering Physics, Cornell University, Ithaca, NY 14853, USA}
\author{Shi-Yuan~Ma}
\affiliation{School of Applied and Engineering Physics, Cornell University, Ithaca, NY 14853, USA}
\email{Present address: Research Laboratory of Electronics, Massachusetts Institute of Technology,
Cambridge, MA 02139, USA}
\author{Ryotatsu~Yanagimoto}
\affiliation{School of Applied and Engineering Physics, Cornell University, Ithaca, NY 14853, USA}
\affiliation{NTT Physics and Informatics Laboratories, NTT Research Inc., Sunnyvale, CA 94085, USA}
\author{Benjamin~A.~Ash}
\affiliation{School of Applied and Engineering Physics, Cornell University, Ithaca, NY 14853, USA}
\author{Tatsuhiro~Onodera}
\affiliation{School of Applied and Engineering Physics, Cornell University, Ithaca, NY 14853, USA}
\author{Tianyu~Wang}
\affiliation{Department of Electrical and Computer Engineering, Boston University, Boston, MA 02215, USA}
\author{Logan~G.~Wright}
\affiliation{Department of Applied Physics, Yale University, New Haven, CT 06511, USA}
\author{Peter~L.~McMahon}
\email{Correspondence to be addressed to: mms477@cornell.edu, jl3668@cornell.edu, pmcmahon@cornell.edu}
\affiliation{School of Applied and Engineering Physics, Cornell University, Ithaca, NY 14853, USA}
\affiliation{Kavli Institute at Cornell for Nanoscale Science, Cornell University, Ithaca, NY 14853, USA}

\begin{abstract}

Illumination using correlated photon sources has been established as an approach to allowing high-fidelity images to be reconstructed from noisy camera frames by taking advantage of the knowledge that signal photons are spatially correlated whereas detector clicks due to noise are uncorrelated. However, in computer-vision tasks, the goal is often not ultimately to reconstruct an image, but to make inferences about a scene---such as what object is present. Here we show how correlated-photon illumination can be used to gain an advantage in a hybrid optical--electronic computer-vision pipeline for object recognition. We demonstrate \textit{correlation-aware training} (CAT): end-to-end optimization of a trainable correlated-photon illumination source and a Transformer backend in a way that the Transformer can learn to benefit from the correlations, using a small number ($\leq$ 100) of shots. We show a classification accuracy enhancement of up to 15 percentage points over conventional, uncorrelated-illumination-based computer vision in ultra-low-light and noisy imaging conditions, as well as an improvement over using untrained correlated-photon illumination. Our work illustrates how specializing to a computer-vision task---object recognition---and training the pattern of photon correlations in conjunction with a digital backend allows us to push the limits of accuracy in highly photon-budget-constrained scenarios beyond existing methods focused on image reconstruction.

\end{abstract}

\maketitle

\vspace{-0.5cm}
\section{Introduction}
\label{sec:intro}

Optical imaging has widely been used to probe and then control the dynamics of physical systems. In cases where the goal is to make inferences about entities in an image, for example classifying cell organelles in images of cells, the extraction of task-specific information from high-dimensional images is increasingly relying on computer vision. A typical computer vision pipeline can be thought of as follows. A camera provides a high-dimensional digital image of a scene and a digital-neural-network extracts information from the image to make inferences about the scene. To enhance the efficacy of information extraction from high-dimensional image data, the design of the image-acquisition hardware (the camera and optics in front of it) has become inextricably linked with the image-processing software through methods such as end-to-end optimization \cite{wetzstein2020inference}. Conceptually, all of these methods operate as an encoder-decoder architecture: a physical-domain encoder extracts scene information and converts it into measurable signals, leaving a digital-domain decoder to interpret the data for specific inference tasks.

Concurrently, the field of photonic quantum sensing has explored the fundamental limits of information carried by photons, developing advanced illumination and measurement schemes to approach these limits \cite{pirandola2018advances, magana2019quantum, moreau2019imaging, defienne2024advances, tsao2025enhancing}. Many of these enhancements arise from leveraging photon correlations in the state of light. For instance, squeezed states have improved microscope performance \cite{casacio2021quantum}, and \textit{N}00\textit{N} states have benefited interferometers \cite{nagata2007beating, israel2014supersensitive, slussarenko2017unconditional}, building on the initial proposal that such correlations could mitigate the effects of shot noise \cite{caves1981quantum}. Light sources with intrinsic spatial correlations between photons, such as those produced by spontaneous parametric down-conversion (SPDC), have enabled the development of novel imaging techniques. These techniques have achieved sub-shot-noise imaging \cite{brida2010experimental, taylor2013biological, zhang2024quantum}, background-light suppression \cite{lloyd2008enhanced, dylov2011nonlinear, lopaeva2013experimental, gregory2020imaging}, ghost imaging \cite{pittman1995optical, strekalov1995observation, bennink2002two}, pattern recognition \cite{ortolano2023quantum}, and enhanced resolution \cite{giovannetti2009sub, toninelli2019resolution, he2023quantum}. By accounting for the illumination's intrinsic photon statistics, these methods capitalize on a perspective often overlooked by their uncorrelated counterparts: higher-order photon correlations can carry information distinct from the mean field\footnote{Here mean field refers to the average intensity of the illumination pattern, i.e., a vector consisting of the mean photon number for each camera pixel.}. This information is particularly valuable in low-light settings, as these correlations demonstrate a natural robustness \cite{grochowski2026distributed} to uncorrelated noise such as that from background light or camera readout. While these techniques have shown a lot of promise, harnessing the information embedded within these higher-order correlations presents a practical challenge. 

\begin{figure}[H]
    \centering
    \includegraphics[width = \textwidth
    ]{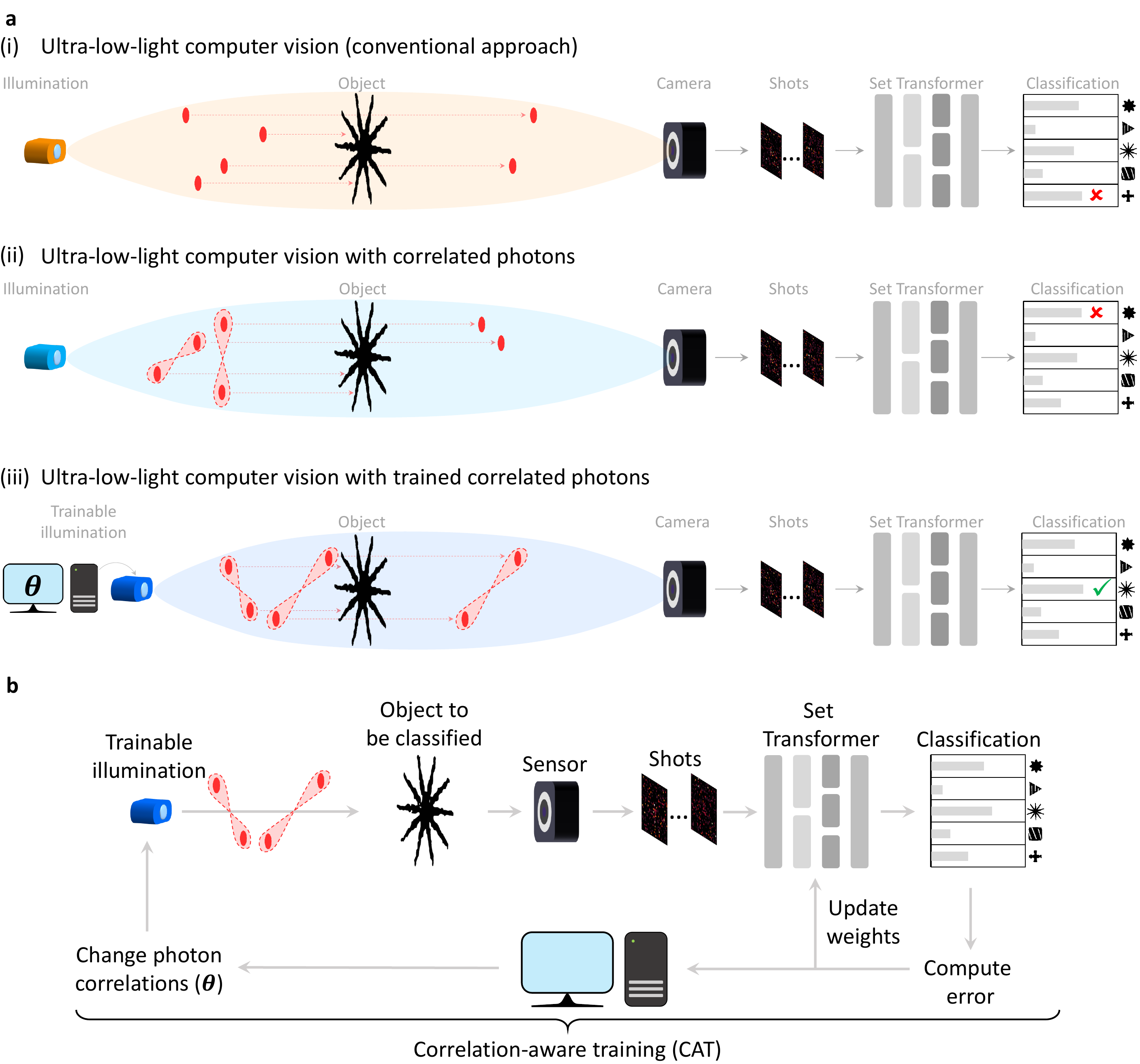}
    \caption{\textbf{Ultra-low-light computer vision with uncorrelated and correlated illumination. a,} A comparison of conventional and our proposed computer vision pipelines. \textit{(i)}, A conventional computer vision pipeline. A light source emits uncorrelated photons towards an object and the photons that are not blocked by the object are recorded on a single-photon-sensitive camera. The shots (frames) from the camera are sent to a Set Transformer backend to classify which object was being imaged. \textit{(ii)}, A computer vision pipeline with a correlated-photon illumination source. Instead of uncorrelated photons as in (i), the light source emits spatially correlated photons. \textit{(iii)}, Our proposed computer vision pipeline with a trainable-correlated-photon source. The light source emits trainable correlated photons that, when trained, allow the object to classified with fewer shots and/or photons. Appendix \ref{Pedagogical_example} introduces an example of a classification task where photon correlations are useful. \textbf{b,} Correlation-aware training (CAT). A schematic of the end-to-end optimization loop used to train photon correlations in a computer vision pipeline. A Set Transformer uses shots from the camera to classify the object being image, and backpropagates the error through a digital model of the correlated light source to update the photon correlations and the weights of the Set Transformer.}
    \label{fig:schematic}
\end{figure}

Estimating the photon correlations or a projection of these correlations into a lower-dimensional space with high fidelity typically requires averaging over a prohibitive number of shots \cite{defienne2024advances}. This necessity for extensive data acquisition can hinder the advantages of correlated illumination, particularly in applications where a low-photon budget or low-latency operation is an inherent constraint.

State-of-the-art digital image classification has been revolutionized by neural networks designed to detect statistical correlations within image pixels. Vision Transformers, for example, use self-attention mechanisms to learn long-range relationships, such as determining that a patch of pixels representing a ``dog's ear'' is statistically related to another patch representing a ``dog's tail''. However, the efficacy of these networks is fundamentally limited by the quality of the input data; in noisy conditions, such as when one is operating with ultra-low-light illumination, the statistical correlations these algorithms rely on can become obscured and difficult to detect without acquiring a large number of shots. We hypothesize that merging correlated illumination techniques with an advanced Transformer backend can enhance ultra-low-light computer vision. This expectation is twofold: (1), the Transformer's self-attention mechanisms—conceptually linked to Gram and covariance matrices \cite{el2021xcit}—are theoretically well-suited to efficiently extract information from photon correlations; and (2), the deterministic relationship between photon pairs should provide a uniquely robust signal for the network to lock onto, surpassing purely statistical correlations. Figures \ref{fig:schematic}a (i) and (ii) outline this concept by contrasting a computer vision pipeline using an uncorrelated-photon source with one using a correlated-photon source. To elucidate how these deterministic photon correlations might help, consider a spatially correlated bi-photon source. The detection of one photon constrains the expected location of its twin, and deviations from this correlated detection pattern can provide additional information about the object across spatially separated, i.e., `non-local'\footnote{non-local refers to pixels positioned rather far away from each other} pixels. Appendix \ref{Pedagogical_example} walks through a pedagogical example in a two-mode photonic circuit that delves deeper into why these photon correlations help when dealing with low-light and noisy imaging conditions. 

The enhancement in performance of correlated-photon sources can further be increased by tailoring the photon correlations to the classes of objects being sensed, as depicted in Figure \ref{fig:schematic}a (iii). Without optimization, the illumination pattern may not be as sensitive to the salient features that differentiate between various classes of images, thereby limiting the potential performance enhancement. Controlling the spatial correlations between photons has been explored in the context of overcoming scattering and wavefront distortion \cite{defienne2018adaptive, lib2020real, cameron2024adaptive}, designing states \cite{boucher2021engineering}, and hiding information in the correlations between photons \cite{nirala2023information, verniere2024hiding}. However, to optimize these correlations for object classification with as few photons and shots as possible, one would need to perform end-to-end optimization \cite{wetzstein2020inference} of the illumination system together with a backend designed to extract photon correlations efficiently.  

Our work shows that designing photon correlations within the illumination in conjunction with a Set Transformer \cite{lee2019settransformer, kim2024attentionquantumcomplexity} backend can enhance classification accuracy in ultra-low-light and noisy imaging conditions. As depicted in Figure \ref{fig:schematic}b, we propose an end-to-end-optimization framework termed correlation-aware training (CAT). CAT learns a tailored illumination pattern for a specific image classification task by training the photon number statistics of a spontaneous parametric down-conversion (SPDC) light source. By co-designing the illumination source and the digital backend whose attention mechanism is well suited for pixel-correlation detection, our framework creates photon-pair patterns tailored to the objects being sensed, as depicted in Figure \ref{fig:schematic}b. This approach allows us to dramatically reduce the number of illumination photons per shot ($\leq 200$) and shots required ($\leq 100$) while simultaneously keeping a high classification accuracy. We experimentally show that, in ultra-low-light conditions, our approach improves classification accuracy by up to 10 percentage points relative to untrained correlated illumination and by up to 18 percentage points relative to conventional uncorrelated illumination. For the classification tasks studied here, our results suggest a 2-6$\times$ decrease in the required photon-budget when using trained correlations as compared to conventional computer vision pipelines. By treating the correlated photon source as a trainable element in a computer vision pipeline, in many cases one can improve a sensor's performance in conditions where photon flux is critically limited and background noise is unavoidable.

\section{Results}\label{Results}
\subsection{An illumination source with trainable photon correlations}\label{illum_source}
\begin{figure}[H]
    \centering
    \includegraphics[width = \textwidth
    ]{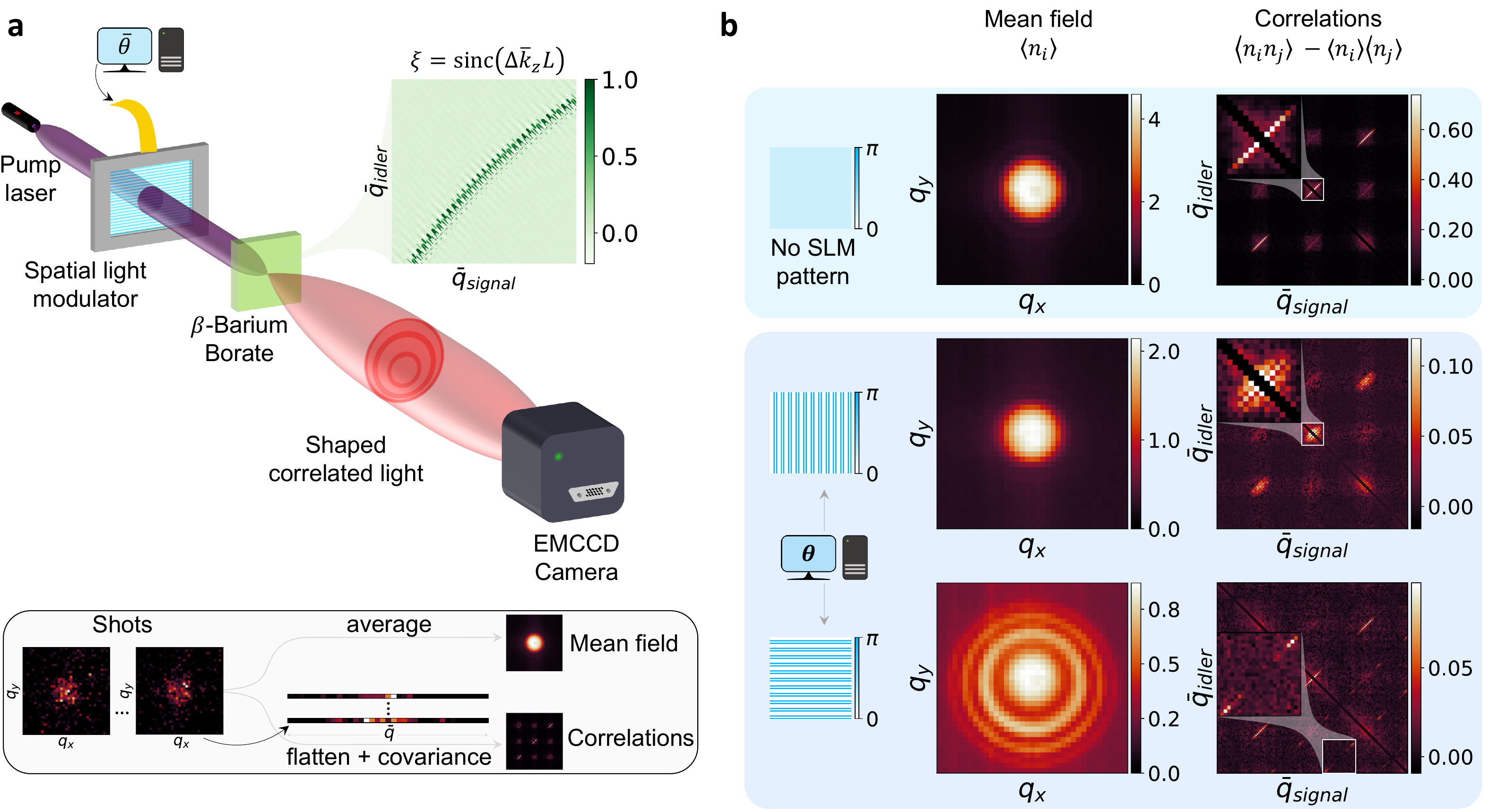}
    \caption{\textbf{Characterization of the trainable-correlated-photon source. a,} A schematic of the experimental setup to create programmable correlations in SPDC photons. An SLM shapes the angular spectrum of a \SI{355}{nm} laser which then produces signal and idler SPDC photons at \SI{710}{nm} in a $\beta$-BBO crystal. Momentum conservation ensures that the SPDC photons are correlated. By changing the angular spectrum of the pump, one can shape the correlations of these photons. The inset figure originating from the $\beta$-BBO crystal is a 2-D representation of the phasematching function. \textbf{b,} Trainable photon correlations. The columns show the phase pattern uploaded to the SLM to tune the angular spectrum of the pump, followed by the experimentally measured mean fields and correlations of the down-converted light averaged over $40$,$000$ shots. $q_x$ and $q_{y}$ denote the transverse wave vector components of the SPDC biphotons (camera pixels since the camera was placed at the fourier plane of the $\beta$-BBO crystal). A diagonal mask was applied to the correlation images to filter out the variance. Appendix \ref{PM_fitting} contains more information about how the phasematching function was fit, how the correlations are computed and the agreement with simulations. We'd like to note here that the covariance matrices were constructed only to demonstrate control of the photon correlations.}
    \label{fig:light_source}
\end{figure}

Photon pairs generated through SPDC were chosen to be the correlated illumination source. The experimental setup used a \SI{355}{nm} laser to pump a $\beta$-Barium borate ($\beta$-BBO) crystal. This produced correlated signal and idler SPDC photons at \SI{710}{nm} using type I phasematching ($e \rightarrow o + o$). A spatial light modulator (SLM) was placed in the beam path of the pump which allowed for control of the pump beam's angular spectrum. The far-field of the SPDC photons was imaged onto an electron-multiplying-charge-coupled-device (EMCCD) camera, and these images were used to characterize the light source. Fig. \ref{fig:light_source}a shows a conceptual diagram of the trainable, correlated illumination source. A detailed description and diagram of the setup can be found in Appendix \ref{illum_sources}.

The spatial correlations between the signal and idler SPDC photons depend on both the phasematching function of the nonlinear crystal used for their generation, and the angular spectrum of the pump laser. By controlling the angular spectrum of the pump light, one can control the spatial correlations of the SPDC photons \cite{boucher2021engineering}. There are multiple ways to model these correlations in the literature \cite{cameron2024shaping, walborn2010spatial}. Here, however, we seek a model that remains valid beyond the biphoton regime, particularly at gains large enough to produce strongly squeezed states. To do this, we use the quantized version of the three-wave mixing problem \cite{wasilewski2006pulsed, quesada2022beyond} in the context of the spatial domain. Our model is built upon work done for parametric amplifiers and/or thermal sources \cite{law2004analysis, brambilla2004simultaneous, gatti1999quantum, gatti2004ghost, shapiro2008computational} while allowing for arbitrary pump angular spectra. 

In the Heisenberg picture, any Gaussian transformation of an optical field (in this case, 2-mode squeezing), can be written as
\begin{gather}
    a^{\text{out}}_{i} = \sum_{j}C_{ij}a^{\text{in}}_{j} + S_{ij}\big(a^{\text{in}}_{j}\big)^{\dagger}
    \label{eq1}
\end{gather}
where $C$ and $S$ are the Green's functions that map the input field operators ${a^{\text{in}}_{i}}$ to the output field operators ${a^{\text{out}}_{i}}$. In the context of SPDC generation (weak squeezing) and when considering the field at the Fourier plane of the $\beta$-BBO crystal, we find that $C \approx \mathbb{I}$ and $S_{ij} \propto \nu_{\text{pump}}(i + j)\xi(i, j)$. Here, $i$ and $j$ denote a discrete basis of transverse-plane momentum vectors\footnote{The basis is continuous in the transverse plane, however, since our camera has pixels of finite size, and for ease of modelling, we assume a discrete basis. The index $i$ is a transverse-plane momentum vector denoted by ($q_{x}, q_{y}$).} (these can be thought of camera pixels since the camera is imaging the Fourier plane of the $\beta$-BBO crystal), $\nu_{\text{pump}}$ denotes the angular spectrum of the pump beam and $\xi$ denotes the phasematching function of the $\beta$-BBO crystal. The inset in Fig. \ref{fig:light_source}a shows a two-dimensional representation of the phasematching function $\xi$ that was fit for the $\beta$-BBO crystal used in our setup. More details about the SPDC model and how the phasematching function was fit can be found in Appendices \ref{SPDC_Hamiltonian} and \ref{PM_fitting}. Using equation \ref{eq1}, one can write down the mean field and photon covariance ($2^{\text{nd}}$ order correlations) as
\begin{gather}
    \expval{n_{i}} = \expval{\big(a^{\text{out}}_{i}\big)^{\dagger}a^{\text{out}}_{i}}{0} = \sum_{j}\abs{S_{ij}}^{2} = \big(SS^{\dagger}\big)_{ii}
    \label{eq2}
\end{gather}
\begin{gather}
    \expval{n_{i}n_{k}} - \expval{n_{i}}\expval{n_{k}} = \Big(\big(S^{*}S^{\top}\odot CC^{\dagger}\big) + \big(S^{*}C^{\dagger}\odot CS^{\top}\big)\Big)_{ik}
    \label{eq3}
\end{gather}
where $\odot$ denotes element-wise multiplication, $*$ denotes a complex conjugate, $\top$ denotes a transpose, and $\dagger$ denotes a hermitian conjugate. These Green's functions can also be used to obtain the biphoton wavefunction at the Fourier plane to simulate the acquisition of shots (see Appendices \ref{SPDC_correlations} and \ref{SPDC_jpd}). Figure \ref{fig:light_source}b shows the experimentally measured mean field (equation \ref{eq2}) and $2^{\text{nd}}$ order photon correlations (equation \ref{eq3}) for different patterns displayed on the SLM, i.e., different pump angular spectra. The photon correlations in experiment were computed by first flattening each shot (frame) and then computing the covariance over these flattened representations of the shots using $40$,$000$ shots. We'd like to note here that the covariance matrices were constructed only as a demonstration of control over the photon correlations. No covariance matrices were explicitly constructed in subsequent experiments. In Appendix \ref{PM_fitting} we describe how a digital model of the experiment was created and show good agreement between simulation and experiment. This model of the SPDC biphotons is used to train an end-to-end optimized image sensor as described in section \ref{sec:training-section}.

\subsection{End-to-end optimization}
\label{sec:training-section}

The output of the experimental setup described earlier (see Appendix \ref{illum_sources}) is a collection of high-dimensional and noisy camera frames (shots), where the information about the absorptive objects to be sensed is sparsely encoded. As mentioned earlier, we chose a Set Transformer as our digital backend \cite{lee2019settransformer}. This choice is motivated by several factors. First, its architecture is designed to efficiently detect pixel relationships within a single shot, bypassing the need to compute an explicit covariance matrix, which would be computationally intensive and require a prohibitive number of shots \cite{defienne2024advances}. The Set Transformer allows us to train on larger sets to effectively capture the underlying correlations, while retaining the flexibility to perform inference on arbitrary (smaller or larger) sets at test time, trading accuracy for throughput as needed. Second, the order in which the shots are collected is inherently permutation-invariant. The Set Transformer, having full attention blocks and no causal attention, correctly treats the collected shots as unordered, meaning each shot is equivalent.

Our primary innovation is the end-to-end optimization of the entire computer vision pipeline, from the illumination source to the digital backend. We employ a Physics-Aware Training (PAT) \cite{wright2022deep} approach, CAT, using the experiment for the forward pass while using the digital model for error backpropagation. A key challenge is backpropagation through the camera, since photon detection is a stochastic sampling operation. While we model the emission of a pair of photons via categorical sampling, standard differentiable relaxations \cite{jang2017categorical, maddison2017concrete} are computationally prohibitive at our system's scale because of the combinatorial explosion of the probability vector's size (scaling as $N^4$ for pairs in a frame of size $N$ pixels $\times$ $N$ pixels). Instead, we use the straight-through estimator \cite{bengio2013estimating} (STE) to bypass the high-dimensional intermediate probability distribution entirely during backpropagation. While the forward pass is simulated as a sampling from the $N^4$-dimensional probability distribution, the backward pass propagates gradients from the set of experimentally sampled frames ($B \times S\times N \times N$) directly to the $B \times N \times N$ mean field\footnote{Modulo an average operation over the set dimension during the backward pass that reduces the gradient dimension from $B \times S\times N \times N$ to $B \times N \times N$.}. Here $B$ refers to the batch size during training and $S$ refers to the number of shots used. CAT backpropagates only through the mean field, treating the gradient of the stochastic sampling step as the identity. Nevertheless, it increases the number of transmitted correlated photon pairs because the mean field is derived from the full $N^4$-dimensional distribution and thus retains information about the underlying photon correlations (see equation \ref{eq2}). Appendix \ref{increase_in_correlations} shows an example, in simulation, where the probability of detecting correlated photon pairs increases when the system is trained with CAT. 

\begin{figure}[H]
    \centering
    \includegraphics[width = \textwidth
    ]{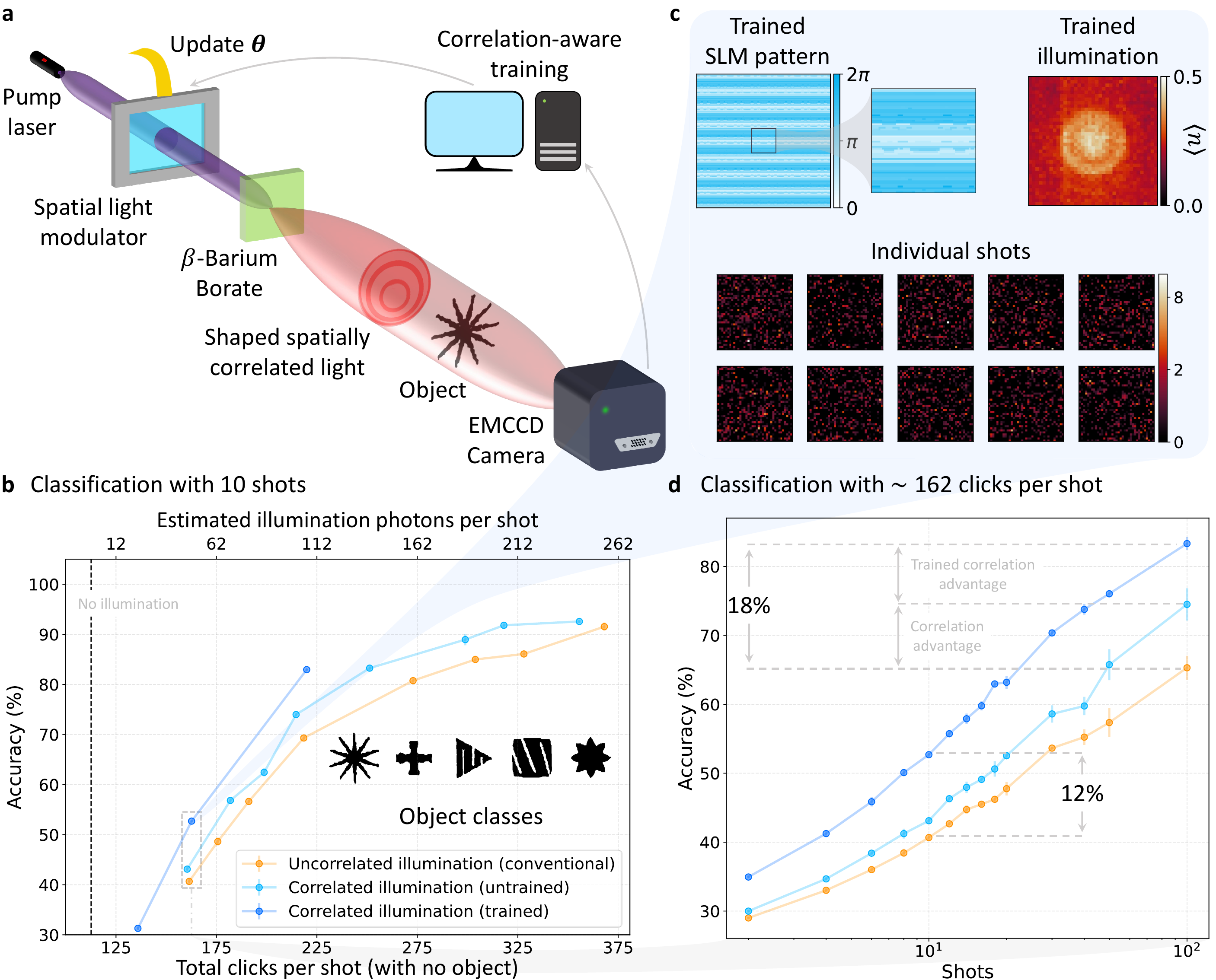}
    \caption{\textbf{Photon-budget-efficient computer vision enabled by trained photon correlations. a,} A schematic of the end-to-end optimization framework used to train this system. The objects to be classified were a subset of the MPEG-7 dataset (see Appendix \ref{MPEG7_task}). The cross-entropy loss computed using experimentally measured shots was used to compute gradients of the SLM parameters using a digital model of the illumination source. \textbf{b,} Classification accuracy of trained correlated illumination (SPDC light generated by angular spectrum shaping of the pump beam; dark blue), untrained correlated illumination (SPDC light generated by a plane wave pump beam; light blue), and untrained uncorrelated illumination (coherent light; light orange) as a function of the illumination power when 10 shots were used for inference. The advantage in classification accuracy is most apparent at lower powers and just enough shots for the Transformer to be able to extract correlation information. Appendix \ref{MPEG7_task} discusses what objects were used in the dataset. Appendix \ref{mpeg7_experiment_results_extended} has a detailed description of the classification accuracy curves across a various range of illumination powers and shots. \textbf{c,} The trained SLM pattern, illumination pattern (averaged over 2000 shots) and example of a series of individual shots after end-to-end optimization was performed for $\sim162$ total clicks per shot (with no object present). \textbf{d,} Classification accuracy of trained correlated illumination, untrained correlated illumination, and untrained uncorrelated illumination as a function of the illumination power when 10 shots are used for inference. As the number of shots increases, the benefit of trained correlations becomes apparent, with an enhancement of $\sim$ 10\% over untrained correlations with just 10 shots.}
    \label{fig:spdc_training}
\end{figure}

Although the physical and digital components can theoretically be trained simultaneously in a standard end-to-end fashion, we found that an alternating optimization schedule yields superior training stability and convergence. We effectively treat the training as a bi-level optimization problem: we update the physical parameters (SLM) for a single epoch, then freeze them to optimize the digital post-processing model (Transformer) for five epochs. This 1:5 cycle is iterated throughout the training process. We hypothesize that this alternating schedule is necessary due to the quantized nature of the physical parameters. A gradient update on the hardware can induce abrupt changes in the system's output distribution. This iterative process allows the digital backend sufficient time to adapt to and learn the new correlation features generated by the updated illumination pattern, ensuring stable and effective convergence for the entire system. Figure \ref{fig:spdc_training}a outlines a high-level schematic of the end-to-end optimization that we implemented. More details can be found in Appendix \ref{training}.

\subsection{Photon correlations enable more photon-budget-efficient computer vision}\label{corr_vs_uncorr}

We performed an image classification task on objects from a subset of the MPEG-7 dataset \cite{yang2012affinity} (see Appendix \ref{MPEG7_task} for more details). These absorptive objects were etched on to a thin chromium film deposited on a fused-silica substrate in the shapes of the selected objects from the MPEG-7 dataset (see Appendix \ref{MPEG7_task}). They were placed at the Fourier plane of the $\beta$-BBO crystal. The EMCCD camera was placed at the image plane of a $4$-$f$ imaging setup following the objects (see Appendix section \ref{mpeg7_experiment_setup}). The image classification experiments were done with three illumination conditions: (1) untrained, uncorrelated light (conventional) using a coherent source at \SI{710}{nm}, (2) untrained, correlated light using a SPDC source at \SI{710}{nm}, and (3) trained, correlated light using a pump-shaped SPDC source at \SI{710}{nm}. The untrained illuminations were chosen so that each object was completely covered (see Appendix \ref{mpeg7_EMCCD_calibration}). We then varied both the illumination power, quantified as the mean number of detection events per shot in the absence of an object, and the number of shots used for classification, in order to assess the effects of shot noise and background noise, including camera readout noise and stray photons.

Figure \ref{fig:spdc_training}b shows the classification accuracy as a function of the average illumination power per shot when 10 shots were used per classification for all three illumination sources. The illumination power was calibrated as the average number of clicks per shot (see Appendices \ref{cam_gain} and \ref{camera_frame_postprocessing} for more details) for a given pump power with no object present. The calibration was restricted to the central 23$\times$23 pixels, which corresponded to the region covered by the absorptive objects. Figure \ref{fig:spdc_training}c shows the trained SLM pattern, the trained mean field, and examples of individual shots used during inference for one of the trained correlated illumination points ($\sim 162$ clicks). On the other hand, Figure \ref{fig:spdc_training}d shows the classification accuracy on the test dataset as a function of shots for $\sim 162$ clicks per shot. We find a benefit to training the correlations even when classifying with as few as 2 shots ($\sim 5$\%) -- we attribute this to the fact that the photons are correlated within a shot and not across different shots. We also find that the training correlations reduces the required number of shots by up to $\sim 6\times$ for a given illumination level (see Figure \ref{fig:spdc_training}d and Appendix Figure \ref{fig:S38}). The advantage from training these correlations continues to grow as the number of shots used for classification increases. For this particular task, the largest benefit from having trained, and even untrained correlations, appears to be in the low illumination power regime with just enough shots for the Set Transformer to be able to access correlation information. Once a high enough illumination or number of shots is used, all illumination sources achieve similar accuracies of $> 90$\%. Appendix Figure \ref{fig:S36} shows the mean-field illumination patterns for all the three light sources at two different powers ($\sim162$ and $\sim220$ average clicks per shot). We see that the optimized illuminations differ from the unoptimized ones and even change at different power levels. At $\sim 162$ clicks per shot, the training procedure increases the brightness in the center as opposed to the periphery, whereas for $\sim 220$ clicks per shot, the optimal pattern is brighter in the periphery. Appendix \ref{mpeg7_experiment_results_extended} has more details regarding the classification accuracy curves across a various range of powers and shots.

\begin{figure}[H]
    \centering
    \includegraphics[width = \textwidth
    ]{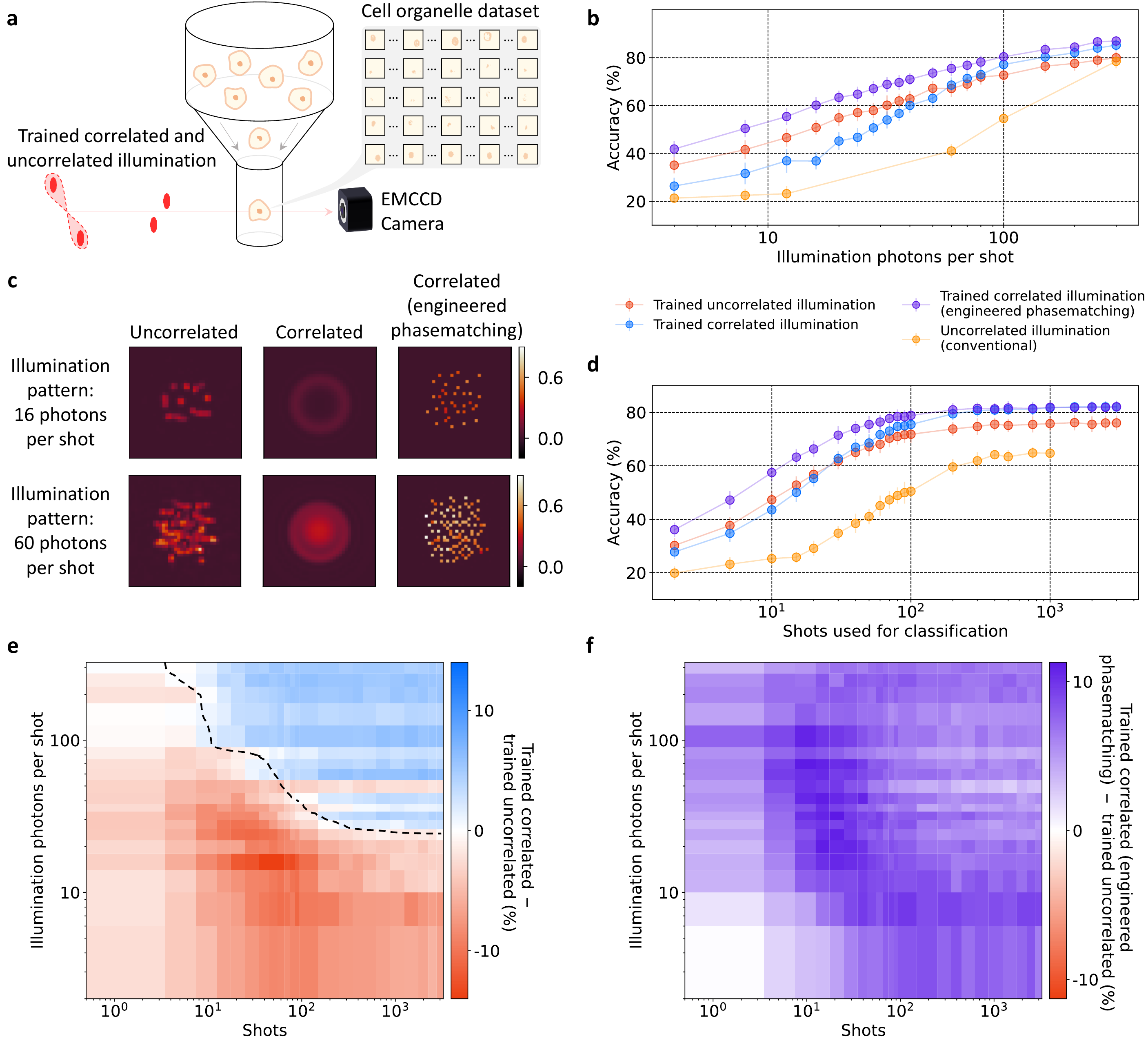}
    \caption{\textbf{Simulation of cell organelle classification using trained uncorrelated and correlated light. a,} An overview of the cell organelle dataset that consists of 5 classes. In simulation, we trained three kinds of illumination to classify cell organelles: (1) an ideal correlated bi-photon source (assuming engineered phase matching and pump shaping, purple), (2) our experimental correlated bi-photon source (blue), and (3) uncorrelated illumination (a laser and an SLM, dark orange). We also compared it to the conventional computer vision pipeline with untrained, uncorrelated illumination (light orange) \textbf{b,} Classification accuracy of the different illuminations as a function of the number of illumination photons per shot with 50 shots used per inference. \textbf{c,} The trained illumination patterns for the three different illumination sources mention in panel \textbf{a} at different photon-budgets. \textbf{d,} Classification accuracy of the different illuminations as a function of the number of shots used during inference at 60 illumination photons per shot. \textbf{e,} The difference between the classification accuracies for trained correlated and trained uncorrelated illumination as a function of the number of illumination photons per shot and the number of shots used for classification. The dotted line (hand-drawn) divides the regions in parameter space where correlated illumination has a higher classification accuracy than uncorrelated illumination. \textbf{f,} The difference between the classification accuracies for trained correlated (engineered phasematching) and trained uncorrelated illumination as a function of the number of illumination photons per shot and the number of shots used for classification. All the simulations were trained using 100 shots at different photon fluxes and used a background-noise probability distribution derived from the EMCCD camera used in experiments. More details can be found in Appendix \ref{cell_training}.}
    \label{fig:cell_classification}
\end{figure}

A natural question that follows is how the performance of the uncorrelated illumination changes when optimized. This is similar to the techniques used in structured illumination \cite{guerra1995super, gustafsson2000surpassing, rego2014practical}, but in this case the illumination is optimized for classification instead of image reconstruction. To answer this, we performed simulations where both the correlated and uncorrelated illumination patterns were trained to classify cell organelles (the cell organelle images were acquired in a flow-cytometry device \cite{schraivogel2022high}). Figure \ref{fig:cell_classification}a outlines a high-level schematic of the proposed cell organelle classification setup with trained correlated and uncorrelated illumination. We considered two kinds of trained correlated (SPDC) light\footnote{We will refer to (1) as the correlated bi-photon source and (2) as the ideal correlated bi-photon source.}: (1) a simulation of the experiment with its digital model (blue), and (2) an `ideal' correlated bi-photon source (purple) that has an arbitrary green's function, which, in principle is experimentally feasible by engineering the phase-matching function \cite{rozenberg2022inverse}. The trained uncorrelated light source (dark orange) was modeled as a coherent Gaussian beam phase-modulated by an SLM, with the cell organelles placed in the Fourier plane of the modulator.

Figure \ref{fig:cell_classification}b shows classification accuracy curves on the test dataset as a function of the illumination photons per shot when 50 shots were used for inference. The ideal correlated bi-photon source consistently achieved the highest performance across our tested conditions, lending support to our hypothesis that trained correlations benefit ultra-low-light computer vision pipelines. The uncorrelated illumination performs better than the correlated bi-photon source when fewer photons are allowed, albeit with a low classification accuracy. Once there are enough photons or shots to be able to extract correlation information, the correlated bi-photon source quickly overtakes and achieves accuracies comparable to the ideal correlated bi-photon source. Conclusions should be drawn here with a little nuance. We attribute the success of the uncorrelated light source over the correlated light source in the few-photon and low-shot regime to its ability to drastically shape the mean field and focus all of the light to a few pixels. 

Figure \ref{fig:cell_classification}c show the trained illumination patterns for two different photon fluxes. This kind of mean-field shaping is not possible with our correlated bi-photon source, but is with the ideal correlated bi-photon source. We hypothesize that, in the low photon-flux regime, as the uncorrelated illumination focuses the light to a few pixels, it appears to `artificially' create correlations for the Set Transformer to use to differentiate between the cell organelles. As the photon/shot budget increases, to increase classification accuracy, the light spreads out more while still keeping a speckle pattern, but this appears to be insufficient to outperform the deterministic correlations from the correlated bi-photon source.

Figure \ref{fig:cell_classification}d shows classification accuracy curves on the test dataset as a function of the shots used for inference when the photon budget was 60 illumination photons per shot. We find that for this ratio of illumination photons to background photons, the maximum accuracy achieved by the correlated sources ($>80\%$) is higher than the uncorrelated ones ($<80\%$). We find that the ideal correlated bi-photon source has up to a $10\times$ reduction in the number of shots required to achieve any classification accuracy larger than 60\% when compared to the conventional classification pipeline. Figure \ref{fig:cell_classification}e shows the difference in classification accuracies between the trained correlated bi-photon and the trained uncorrelated illumination sources as a function of the illumination photons per shot and the number of shots used for inference. The hand-draw dotted line depicts the crossover point where the deterministic photon correlations from the correlated bi-photon source start to surpass the artificially created ones from the uncorrelated source. Figure \ref{fig:cell_classification}f shows the difference in classification accuracies between the trained ideal correlated bi-photon and the trained uncorrelated illumination sources as a function of the illumination photons per shot and the number of shots used for inference. All the simulations were performed with a background-noise probability distribution derived from the EMCCD camera used in experiments and with no photon loss. Appendix \ref{loss_effect} has details about the effect of photon loss on the ideal correlated bi-photon source. More details about these simulations can be found in Appendix \ref{cell_training}.

\section{Discussion}\label{sec:discussion}
\subsection{Summary of results}
We have demonstrated a computer vision scheme that leverages trainable, correlated illumination to enhance classification accuracy in ultra-low-light and noisy environments. By co-optimizing a correlated light source (SPDC biphotons) and a Transformer-based backend, our end-to-end pipeline learns to generate correlation patterns tailored to specific image classification tasks. For the experimental task considered here with signal-to-background ratios (SBRs) lower than $1$, we show (see Figure \ref{fig:spdc_training} and Appendix Figures \ref{fig:S31} - \ref{fig:S38}) that training correlations can reduce the photon budget up to $6\times$ when compared to conventional computer vision pipelines to reach a classification accuracy of 90\%. In simulation, for the cell organelle classification task we show a photon budget reduction of up to $10\times$, albeit with no photon loss considered.  

Our results show that the advantage of training these correlations is most pronounced in ultra-low-light conditions with uncorrelated, background noise. As the SBR improves with higher photon flux, the performance of other illumination sources converges with that of ours for the classification tasks studied here. The central thesis of this work is that photon correlations within an illumination source are a powerful, trainable resource. Our proposed computer vision pipeline can both generate and efficiently interpret these correlations. The light source encodes class distinguishing information not only into the mean field, but also higher-order correlations that are resilient to noise, and the Set Transformer is particularly adept at identifying these non-local relationships. We are of the opinion that this work serves as a proof-of-concept for a broader class of sensors that treat the illumination itself as a trainable component of the inference task.

\subsection{Limitations}
It is important, however, to acknowledge the several practical drawbacks of our current experimental prototype and methods. While all appear to not be fundamental, addressing these issues would improve the practicality and performance of our approach. First, any advantage offered by photon correlations is inherently susceptible to photon loss (see Appendix \ref{loss_effect}). Lost photons make extracting correlation information harder, which in turn reduces the performance enhancement. However, in the setting we consider, the primary source of loss is at detection and not the imaging pipeline preceding it \cite{moreau2019imaging}. The ongoing development of better detectors, such as improved quantitative CMOS cameras \cite{roberts2024comparison} and SNSPD arrays \cite{resta2023gigahertz}, will directly mitigate this issue. As detectors approach near-unity efficiency, the impact of loss on the integrity of photon correlations will diminish, making the benefits of our approach more accessible.

Second, our choice of a correlated photon source, SPDC biphotons, while excellent for generating strong correlations and a proof-of-concept experiment, does not allow for substantial mean-field shaping due to its spatial incoherence \cite{cameron2024shaping}. This restricts the kinds of illumination patterns that can be produced. Fortunately, there have been recent demonstrations of other trainable light sources that might offer greater control over both mean-field patterns and photon correlations simultaneously. For instance, shaping the input of a nonlinear fiber and filtering the output allows one to create below-shot-noise correlations while also shaping the mean field \cite{sloan2025programmable}. Programmable phase matching functions in slab-waveguides \cite{yanagimoto2025programmable} might also allow the generation of tailored squeezed states for computer vision. Incorporating such sources could potentially extend the benefits of our scheme to lower SNR levels as shown in Figure \ref{fig:cell_classification}. 

Third, it is unclear if CAT will work for $3^{\text{rd}}$-order correlations and higher. The dimensionality of the space in which higher order correlations lie scales exponentially with the order. As a result, it is unlikely that backpropagation of gradients only through the mean field will be able to train such sorts of correlations efficiently for non-gaussian states. Extending this framework will necessitate moving beyond simple straight-through estimators toward methods capable of handling entangled joint distributions. Recent advances in machine learning, such as implicit multivariate reparameterization \cite{figurnov2018implicit} or path-wise gradients \cite{jankowiak2019pathwise} might provide mathematical pathways to backpropagate through highly correlated variables without discarding off-diagonal covariance information.

\subsection{Related work}
The past several decades have seen the development of computer vision and image sensing systems that have pushed towards photon-starved and noise-dominated regimes \cite{wernick1986image, ota2018ghost, goyal2021photon, ortolano2023quantum, minati2026quantum}. Early demonstrations of photon-limited pattern recognition established foundational benchmarks in ideal dark-room conditions, relying on high SBRs to extract features \cite{saaf1995photon, zhu2020photon}. As sensor technologies matured, modern inference frameworks successfully reduced the requisite photon budget down to tens or hundreds of photons per inference \cite{chen2016vision, ma2026machine}. However, pushing performance into environments where the noise floor is larger the signal---regimes where the SBR drops below unity---requires different approaches to information encoding. To contextualize our contribution, Appendix~Figure~\ref{fig:comparison_QI} presents a comparison of low-light computer vision and pattern recognition literature, mapping the total illumination photon budget per inference against the operating SBR. While foundational quantum illumination experiments have operated at very low SBRs (often below $10^{-6}$), these demonstrations are fundamentally about single-mode binary classification tasks (i.e., detecting a target's presence or absence) and require a large photon budget to outperform uncorrelated-photon techniques \cite{zhang2015entanglement}. We attempt to bridge this gap, by translating the noise resilience of engineered photon correlations into the domain of spatially multimode computer vision at ultra-low photon budgets. By training spatial biphoton correlations, our framework isolates target features from the noise floor, maintaining high-accuracy inference with less than a 100 shots even at an SBR of 0.45.

\subsection{Outlook}
The quantum statistical properties of light are usually treated as imposing an immutable limit on imaging performance, such as setting a photon-budget floor due to shot noise. Instead, we have advocated for treating them as trainable degrees of freedom, e.g., having illumination with programmable photon statistics optimized in an end-to-end fashion with a neural-network backend. Our work is a step towards fully quantum computational sensing and imaging \cite{sarovar2023quantum, khan2025quantum}, as we explain in the rest of this section.

Our experiment only took advantage of the spatial intensity correlations of the biphotons from our Type-I SPDC illumination source\footnote{Even when the biphotons' full bipartite entanglement is not strictly utilized due to the lack of phase measurements, quantum discord ensures the biphoton source retains a quantum statistical advantage over purely classical light, such as coherent states \cite{weedbrook2016discord}.}. However, such sources are known to generate bipartite entanglement \cite{couteau2018spontaneous}. An exciting avenue for future exploration is whether and how it could be possible to gain further advantage by exploiting the fact that the photons are not merely classically correlated but also have quantum correlations \cite{ono2013entanglement, defienne2021polarization}. As illumination sources and single-photon detectors mature, they will enable computer-vision systems that can take advantage of quantum-optical states---such as two-mode squeezed states---with tailored many-photon ($3+$) correlations.

A natural extension of the scheme we have demonstrated, in which the illumination source is trained, would be to combine a trainable quantum illumination source with a trainable optical system (called an \textit{encoder}) that is inserted before the camera. Rather than relying solely on a digital backend to process the detected photon statistics, for which the fundamental limits of feature extraction under sampling noise have been established \cite{hu2023tackling}, the light that has interacted with the object could be pre-processed before conversion to digital electronic signals using an optical encoder \cite{wetzstein2020inference,choi2025free,ma2026machine}. The optical encoder could even take the form of a quantum-optical neural network \cite{steinbrecher2019quantum, sarovar2023quantum}. By performing end-to-end optimization across the illumination source, the optical encoder, and the digital backend, the optical encoder could help to extract task-relevant information from the optical state prior to its measurement by the camera. This approach, where the illumination, optical encoder, and digital backend are jointly trained, could realize a quantum computational vision pipeline capable of operating at the physical limits of noise and photon scarcity.

The approach we have presented in this paper, and the possible extensions to it that we have just described, could be useful in conditions where signal power and/or integration time are small, and the optical loss from illumination to detection is relatively low. In biomedical imaging, for instance, live-cell microscopy and deep-tissue imaging face photon-budget limits to prevent phototoxicity and photobleaching \cite{taylor2013biological}. By extracting features from photon correlations rather than just intensity, there is the prospect to increase the accuracy of classification of cellular structures (for example) in photon-budget-limited imaging scenarios. 

\section*{Data and code availability}\label{code_links}

All simulation data, experimental data, and analysis code needed to reproduce the main results presented in this paper are available at \url{https://doi.org/10.5281/zenodo.19556664}.

\section*{Author contributions}
M.M.S., T.W., L.G.W., and P.L.M. conceived the project. M.M.S., T.W., and L.G.W. designed the experimental setup. M.M.S built the experimental setup. S-Y.M. assisted with EMCCD camera setup and calibration. S-Y.M. and J.L. proposed the training framework. M.M.S. and J.L. performed the end-to-end optimization experiments. M.M.S., J.L., and M.O. performed the numerical simulations. R.Y. and T.O. aided in the modeling of the SPDC light source. B.A.A. and R.Y. fabricated the objects used in classification experiments. All authors contributed to writing the manuscript. P.L.M. supervised the project.

\section*{Acknowledgements}
The authors wish to thank NTT Research for their financial and technical support. Portions of this work were supported by the National Science Foundation (award CCF-1918549) and a David and Lucile Packard Foundation Fellowship. P.L.M. acknowledges membership of the CIFAR Quantum Information Science Program as an Azrieli Global Scholar. This work was performed in part at the Cornell NanoScale Facility, a member of the National Nanotechnology Coordinated Infrastructure (NNCI), which is supported by the National Science Foundation (Grant NNCI-2025233). We acknowledge helpful discussions with Chris Alpha, Maxwell Anderson, Valeria Cimini, Nata\v{s}a \DJ uri\'{c}, Saeed Ahmed Khan, Vladimir Kremenetski, Benjamin Malia, Sridhar Prabhu, Federico Presutti, Purnendu Sen, Linnea Smith, Martin Stein, Wayne Wang, Fan Wu, Yang Xu, Yongqi Zhang, Yiqi Zhao, and Ruomin Zhu.

\section*{Competing interests}
The authors declare no competing interests.

%\bibliography{references}


\counterwithin{figure}{section} 
\renewcommand{\thefigure}{\thesection\arabic{figure}}
\renewcommand{\thetable}{\thesection\arabic{table}}
\newpage
\appendix
\section{The benefit of correlations during sensing}\label{Pedagogical_example}
\begin{figure}[H]
    \centering
    \includegraphics[width = \textwidth]{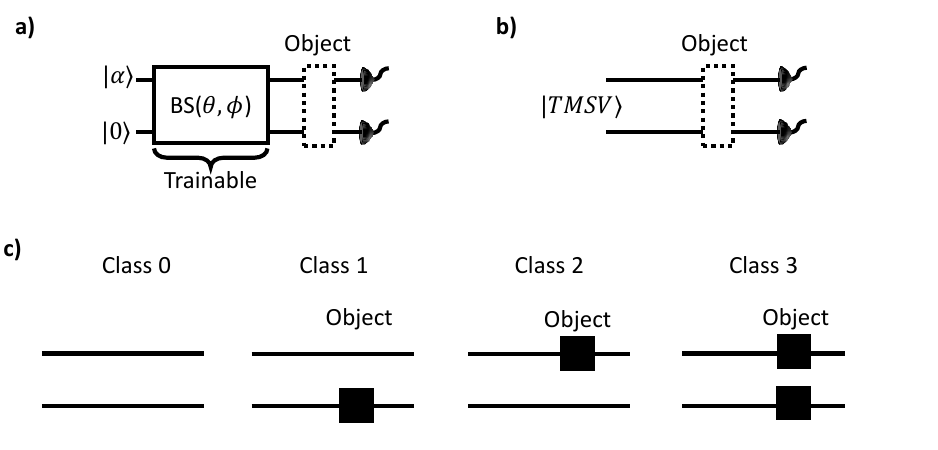}
    \caption{\textbf{Toy example demonstrating the benefit of correlations. a,} A coherent state input to a two-mode photonic circuit with a trainable beamsplitter.\textbf{b, } A two-mode-squeezed vacuum input to a two-mode photonic circuit. \textbf{c, } The different classes to be differentiated between: (1) no object present, (2) the object blocks all photons in mode 1, (3) the object blocks all photons in mode 0, (4) the object blocks all photons in mode 0 and mode 1.}
    \label{fig:S1}
\end{figure}

To demonstrate that correlations in the illumination source can help when classifying objects, we will analyze the following toy example. Consider a two-mode photonic circuit with a detector on each mode as shown in Figure \ref{fig:S1}. There are 2 illumination conditions: (1) A coherent state followed by a trainable beamsplitter (uncorrelated light), (2) A two-mode squeezed state (correlated light). Each of these illuminations will try to distinguish between three classes of objects: (1) no object present, (2) the object blocks all photons in mode 1, (3) the object blocks all photons in mode 0, (4) the object blocks all photons in mode 0 and mode 1. For simplicity, we consider the detectors to be threshold detectors, i.e., they will click if they detect one or more photons. Finally, we assume a background click can happen on each detector with probability $\epsilon$.
Since the detectors are threshold detectors, the most general probability mass function of the possible photon events before interaction with the object can be represented as $p_{0}$ (for event $\ket{00}$), $p_{1}$ (for event $\ket{10}$), $p_{2}$ (for event $\ket{01}$) and $p_{3}$ (for event $\ket{11}$), where $p_{0} + p_{1} + p_{2} + p_{3} = 1$ and $\ket{1}$ denotes one or more photons. The probabilities are determined by the beam splitter for the coherent case. Each object class occurs with a probability of $\frac{1}{4}$. Now, there are four possible measurements that can be made -- $\ket{00}$, $\ket{10}$, $\ket{01}$ and $\ket{11}$. Tables \ref{tab:Meas00} $\rightarrow$ \ref{tab:Meas11} show the probabilities of measuring $\ket{00}$ $\rightarrow$ $\ket{11}$ from the possible illuminations and object classes.

\begin{table}[H]
    \centering
    % Changed column specifiers for a better fit
    \begin{NiceTabular}{|p{5cm}|c|c|c|c|}
        \toprule
        % First header row: empty cell, then a block spanning 4 columns
        \diagbox{\textbf{Illumination}}{\textbf{Class}} & 0 & 1 & 2 & 3 \\
        \midrule
        $\ket{00}$ & $\frac{1}{4}p_{0}(1 - \epsilon)^{2}$ & $\frac{1}{4}p_{0}(1 - \epsilon)^{2}$ & $\frac{1}{4}p_{0}(1 - \epsilon)^{2}$ & $\frac{1}{4}p_{0}(1 - \epsilon)^{2}$\\
        \midrule
        $\ket{10}$ & 0 & 0 & $\frac{1}{4}p_{1}(1 - \epsilon)^{2}$ & $\frac{1}{4}p_{1}(1 - \epsilon)^{2}$\\
        \midrule
        $\ket{01}$ & 0 & $\frac{1}{4}p_{2}(1 - \epsilon)^{2}$ & 0 & $\frac{1}{4}p_{2}(1 - \epsilon)^{2}$\\
        \midrule
        $\ket{11}$ & 0 & 0 & 0 & $\frac{1}{4}p_{3}(1 - \epsilon)^{2}$\\
        \bottomrule
    \end{NiceTabular}
    \caption{Probabilities of measuring $\ket{00}$ from different illuminations and object classes}
    \label{tab:Meas00}
\end{table}

\begin{table}[H]
    \centering
    % Changed column specifiers for a better fit
    \begin{NiceTabular}{|p{5cm}|c|c|c|c|}
        \toprule
        % First header row: empty cell, then a block spanning 4 columns
        \diagbox{\textbf{Illumination}}{\textbf{Class}} & 0 & 1 & 2 & 3\\
        \midrule
        $\ket{00}$ & $\frac{1}{4}p_{0}\epsilon(1 - \epsilon)$ & $\frac{1}{4}p_{0}\epsilon(1 - \epsilon)$ & $\frac{1}{4}p_{0}\epsilon(1 - \epsilon)$ & $\frac{1}{4}p_{0}\epsilon(1 - \epsilon)$\\
        \midrule
        $\ket{10}$ & $\frac{1}{4}p_{1}(1 - \epsilon)$ & $\frac{1}{4}p_{1}(1 - \epsilon)$ & $\frac{1}{4}p_{1}\epsilon(1 - \epsilon)$ & $\frac{1}{4}p_{1}\epsilon(1 - \epsilon)$\\
        \midrule
        $\ket{01}$ & 0 & $\frac{1}{4}p_{2}\epsilon(1 - \epsilon)$ & 0 & $\frac{1}{4}p_{2}\epsilon(1 - \epsilon)$\\
        \midrule
        $\ket{11}$ & 0 & $\frac{1}{4}p_{3}(1 - \epsilon)$ & 0 & $\frac{1}{4}p_{3}\epsilon(1 - \epsilon)$\\
        \bottomrule
    \end{NiceTabular}
    \caption{Probabilities of measuring $\ket{10}$ from different illuminations and object classes}
    \label{tab:Meas10}
\end{table}

\begin{table}[H]
    \centering
    % Changed column specifiers for a better fit
    \begin{NiceTabular}{|p{5cm}|c|c|c|c|}
        \toprule
        % First header row: empty cell, then a block spanning 4 columns
        \diagbox{\textbf{Illumination}}{\textbf{Class}} & 0 & 1 & 2 & 3\\
        \midrule
        $\ket{00}$ & $\frac{1}{4}p_{0}\epsilon(1 - \epsilon)$ & $\frac{1}{4}p_{0}\epsilon(1 - \epsilon)$ & $\frac{1}{4}p_{0}\epsilon(1 - \epsilon)$ & $\frac{1}{4}p_{0}\epsilon(1 - \epsilon)$\\
        \midrule
        $\ket{10}$ & 0 & 0 &$\frac{1}{4}p_{1}\epsilon(1 - \epsilon)$ & $\frac{1}{4}p_{1}\epsilon(1 - \epsilon)$\\
        \midrule
        $\ket{01}$ & $\frac{1}{4}p_{2}(1 - \epsilon)$ & $\frac{1}{4}p_{2}\epsilon(1 - \epsilon)$ & $\frac{1}{4}p_{2}(1 - \epsilon)$ & $\frac{1}{4}p_{2}\epsilon(1 - \epsilon)$\\
        \midrule
        $\ket{11}$ & 0 & 0 &$\frac{1}{4}p_{3}(1 - \epsilon)$ & $\frac{1}{4}p_{3}\epsilon(1 - \epsilon)$\\
        \bottomrule
    \end{NiceTabular}
    \caption{Probabilities of measuring $\ket{01}$ from different illuminations and object classes}
    \label{tab:Meas01}
\end{table}

\begin{figure}[H]
    \centering
    \includegraphics[width = 0.75\textwidth]{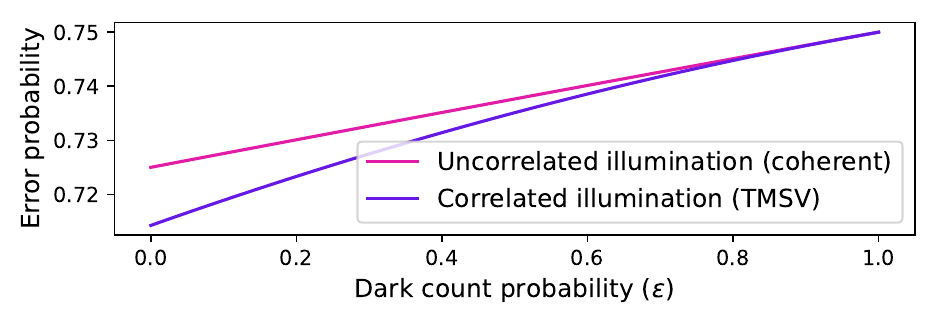}
    \caption{\textbf{Single-shot error probabilities for correlated and uncorrelated illumination.} The single-shot error probability for both cases when $0.1$ illumination photons were used for classification. For the coherent case, the beamsplitter was optimized for each value of $\epsilon$ using a grid search.}
    \label{fig:S2}
\end{figure}

\begin{table}[H]
    \centering
    % Changed column specifiers for a better fit
    \begin{NiceTabular}{|p{5cm}|c|c|c|c|}
        \toprule
        % First header row: empty cell, then a block spanning 4 columns
        \diagbox{\textbf{Illumination}}{\textbf{Class}} & 0 & 1 & 2 & 3 \\
        \midrule
        $\ket{00}$ & $\frac{1}{4}p_{0}\epsilon^{2}$ & $\frac{1}{4}p_{0}\epsilon^{2}$ & $\frac{1}{4}p_{0}\epsilon^{2}$ & $\frac{1}{4}p_{0}\epsilon^{2}$\\
        \midrule
        $\ket{10}$ & $\frac{1}{4}p_{1}\epsilon$ & $\frac{1}{4}p_{1}\epsilon$ & $\frac{1}{4}p_{1}\epsilon^{2}$ & $\frac{1}{4}p_{1}\epsilon^{2}$\\
        \midrule
        $\ket{01}$ & $\frac{1}{4}p_{2}\epsilon$ & $\frac{1}{4}p_{2}\epsilon^{2}$ & $\frac{1}{4}p_{2}\epsilon$ & $\frac{1}{4}p_{2}\epsilon^{2}$\\
        \midrule
        $\ket{11}$ & $\frac{1}{4}p_{3}$ & $\frac{1}{4}p_{3}\epsilon$ & $\frac{1}{4}p_{3}\epsilon$ & $\frac{1}{4}p_{3}\epsilon^{2}$\\
        \bottomrule
    \end{NiceTabular}
    \caption{Probabilities of measuring $\ket{11}$ from different illuminations and object classes}
    \label{tab:Meas11}
\end{table}
Given a particular measurement, the decision about which class to choose is decided by class that has the highest probability contribution. For example, if one measures $\ket{00}$, from table \ref{tab:Meas00} it's clear that the decision should be class 3 because its contribution to the probability of measuring $\ket{00}$ is $\frac{1}{4}(1 - \epsilon)^{2}$ which is higher than that of the other classes (the maximum posterior rule can be applied here since all the prior class probabilities are equal). Once optimized for each $\epsilon$ we can compare the single shot error probabilities between the uncorrelated and correlated illuminations. Figure \ref{fig:S2} shows the single-shot error probability for both cases when $0.1$ illumination photons were used for classification and we clearly see that the correlated case outperforms the uncorrelated case for all values of $\epsilon$.

We'd like to note here that if another classification task was chosen, such as only differentiating between class 1 and class 2, uncorrelated light would perform the same as correlated light.

\section{Experimental setup and methods}
\subsection{Correlated and uncorrelated sources}\label{illum_sources}
\begin{figure}[H]
    \centering
    \includegraphics[width = 0.75\textwidth
    ]{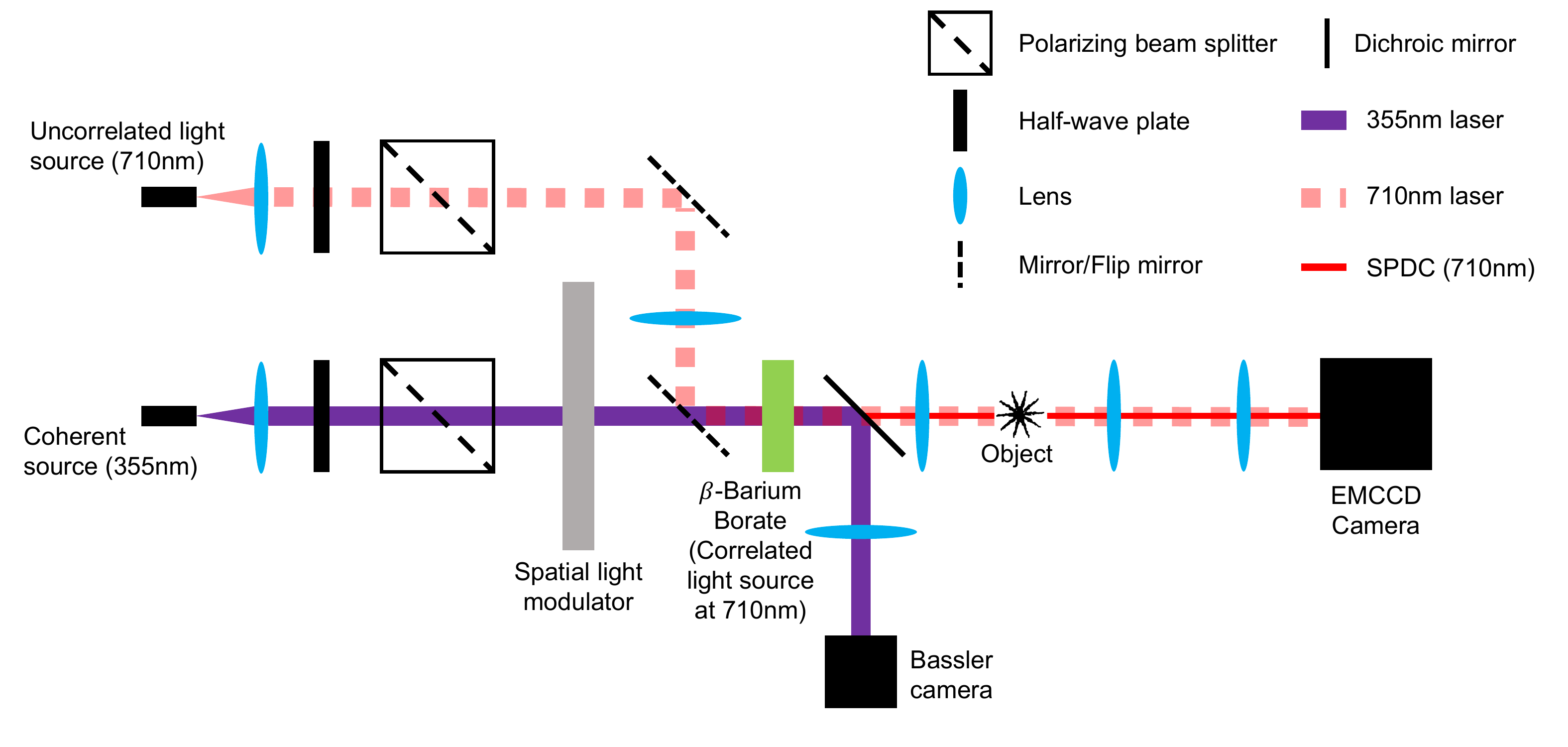}
    \caption{\textbf{Schematic of the experimental setup.} The uncorrelated illumination was a coherent source at \SI{710}{nm} produced using a laser diode. The correlated illumination source was generated using spontaneous parametric down conversion (SPDC) of \SI{355}{nm} pump photons.}
    \label{fig:S4}
\end{figure}

\begin{figure}[H]
    \centering
    \includegraphics[width = 0.75\textwidth
    ]{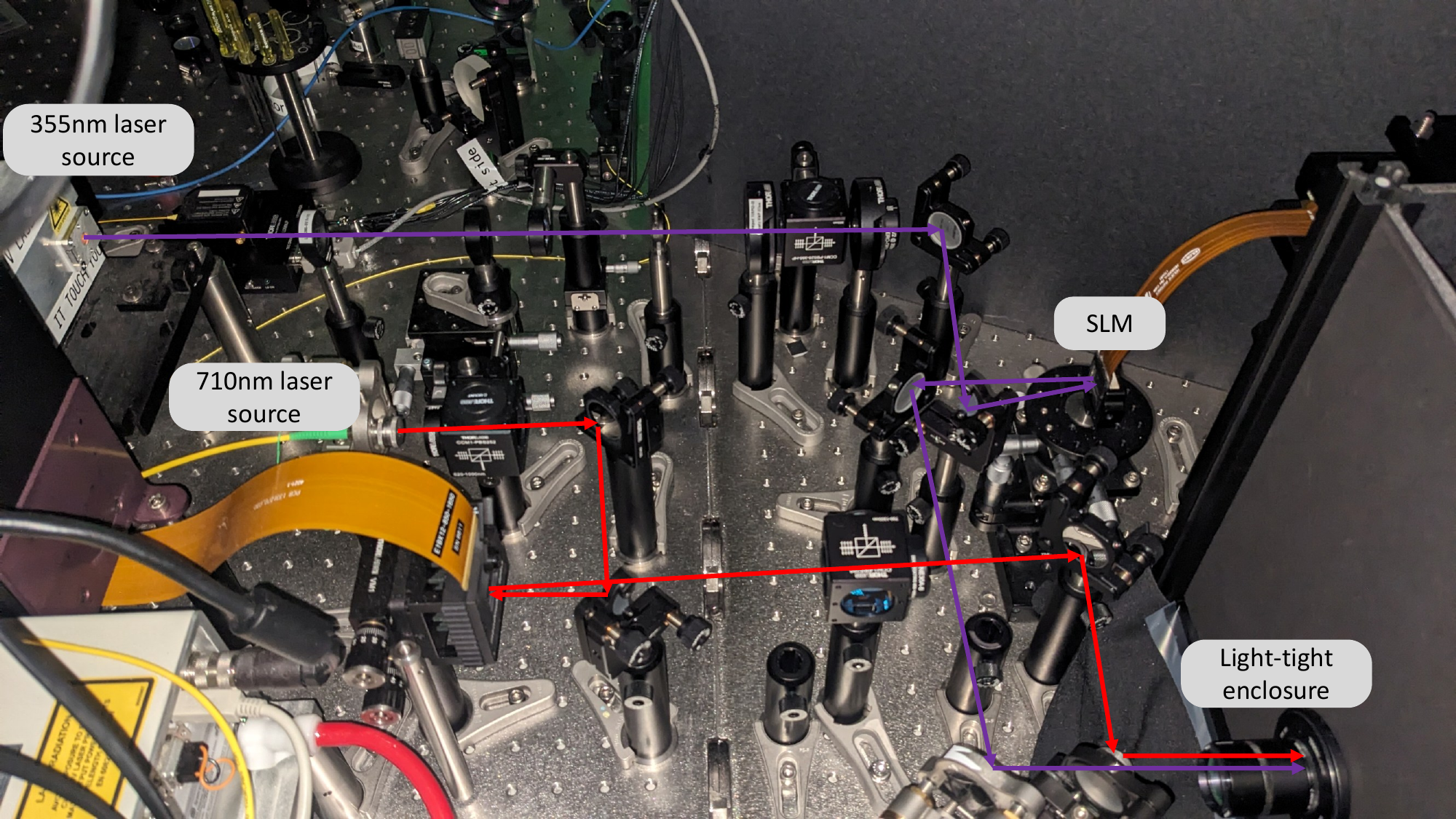}
    \caption{\textbf{Picture of the experimental setup (part 1).} A picture of the experimental setup used to shape the angular spectrum of the \SI{355}{nm} pump and the beam path of the \SI{710}{nm} laser used for the conventional computer vision scheme.}
    \label{fig:S40}
\end{figure}

\begin{figure}[H]
    \centering
    \includegraphics[width = 0.75\textwidth
    ]{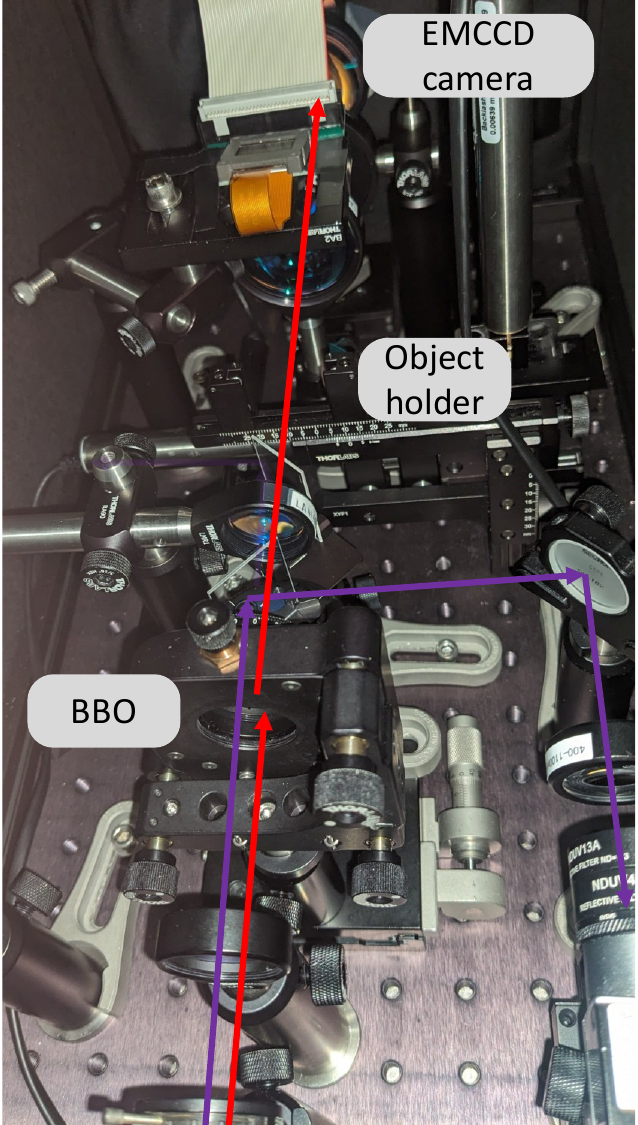}
    \caption{\textbf{Picture of the experimental setup (part 2).} A picture of the experimental setup inside the light-tight enclosure from supplementary figure \ref{fig:S40}.}
    \label{fig:S41}
\end{figure}

Both the correlated and uncorrelated illumination sources used were at \SI{710}{nm}. The correlated illumination source was generated using spontaneous parametric down conversion (SPDC) of \SI{355}{nm} pump photons (produced using an Xcyte CY-355-020-M1 laser) in a $\beta$-Barium borate ($\beta$-BBO, \SI{5}{mm} thickness) crystal phasematched for type I SPDC. The uncorrelated illumination was produced using a laser diode (Thorlabs LP705-SF15). Supplementary figure \ref{fig:S4} shows a schematic of the experimental setup. Supplementary figures \ref{fig:S40} and \ref{fig:S41} show pictures of the experimental setup. Detailed diagrams with part numbers can be found in Appendix \ref{mpeg7_experiment_setup}.

\subsection{Modeling the SPDC hamiltonian and green's functions}\label{SPDC_Hamiltonian}
We will largely be following the derivation done in reference \cite{walborn2010spatial}, and framing it in the language of the green's functions operators \cite{wasilewski2006pulsed, quesada2022beyond}. To begin, we write down the hamiltonian, $H$, of an arbitrary EM field in terms of its electric and magnetic components
\begin{gather}\label{eq:EM_hamilotnian}
    H = \frac{1}{2}\int d^{3}r\big[\bar{D}(\bar{r}, t)\cdot\bar{E}(\bar{r}, t) + \bar{H}(\bar{r}, t)\cdot\bar{B}(\bar{r}, t)\big]
\end{gather}
Now, the displacement field, $\bar{D}(\bar{r}, t)$, is defined in terms of the electric field, $\bar{E}(\bar{r}, t)$, and electric polarization, $\bar{P}(\bar{r}, t)$, as $\bar{D}(\bar{r}, t) = \epsilon_{0}\bar{E}(\bar{r}, t) + \bar{P}(\bar{r}, t)$. Inside a nonlinear crystal like $\beta$-BBO, the dominant nonlinear electric polarization, $\bar{P}^{\text{NL}}(\bar{r}, t)$, contribution comes from it's second-order term given by
\begin{gather}\label{eq:nonlinear_P}
    P^{\text{NL}}_{i}(\bar{r}, t) = \epsilon_{0}\int\int dt'dt''\chi_{ijk}(t', t'')E_{j}(\bar{r}, t - t')E_{k}(\bar{r}, t - t'')
\end{gather}
where $\chi_{ijk}(t', t'')$ is the second-order susceptibility tensor. This allows us to write equation \ref{eq:EM_hamilotnian} as 
\begin{gather}\label{eq:heis_hamiltonian}
    H = H_{0} + H_{I}
\end{gather}
where $H_{0}$ is the EM field hamiltonian from the linear part of the electric polarization and 
\begin{gather}\label{eq:interaction_hamiltonian}
    H_{I} = \int\int\int d^{3}rdt'dt''\chi_{ijk}(t', t'')E_{i}(\bar{r}, t)E_{j}(\bar{r}, t - t')E_{k}(\bar{r}, t - t'')
\end{gather}
is the EM field hamiltonian from the nonlinear part of the electric polarization. In our case $H_{I}$ can be treated as a perturbation since the second-order susceptibility is much smaller than the first-order susceptibility. To quantize $H_{I}$, we first express the electric field in a basis of plane waves, i.e., $\bar{E}(\bar{r}, t) = \bar{E}^{+}(\bar{r}, t) + \bar{E}^{-}(\bar{r}, t)$, and then use the operator form of the plane waves. Essentially,
\begin{gather}
    \hat{E}^{+}(\bar{r}, t) = \frac{1}{\sqrt{V}}\sum_{\bar{k}, \sigma} \epsilon_{\bar{k}, \sigma}\hat{a}_{\bar{k}, \sigma}e^{i(\bar{k}.\bar{r} - \omega t)}\bar{e}_{\bar{k}, \sigma} = \bigg(\hat{E}^{-}(\bar{r}, t)\bigg)^{\dagger}
\end{gather}
where $\bar{e}_{\bar{k}, \sigma}$ is the 2D polarization vector for spatial mode $\bar{k}$ with polarization $\sigma$, $\hat{a}_{\bar{k}, \sigma}$ is the photon annihilation operator for a photon in spatial mode $\bar{k}$ with polarization $\sigma$, $\epsilon_{\bar{k}, \sigma} = \sqrt{\frac{\hbar\omega(\bar{k}, \omega)}{2\epsilon_{0}n^{2}(\bar{k}, \sigma)}}$, $V$ is the quantization volume and $n(\bar{k}, \sigma)$ is the refractive index for spatial mode $\bar{k}$ and polarization $\sigma$. Thus,
\begin{gather}
    \hat{H}_{I} = \frac{1}{2V^{\frac{3}{2}}}\sum_{\bar{k}_{p}, \sigma_{p}}\sum_{\bar{k}_{i}, \sigma_{i}}\sum_{\bar{k}_{s}, \sigma_{s}}\bigg[(\bar{e}_{\bar{k}_{s}, \sigma_{s}})_{\alpha}^{*}(\bar{e}_{\bar{k}_{i}, \sigma_{i}})_{\beta}^{*}(\bar{e}_{\bar{k}_{p}, \sigma_{p}})_{\gamma}\hat{a}^{\dagger}_{\bar{k}_{s}, \sigma_{s}}\hat{a}^{\dagger}_{\bar{k}_{i}, \sigma_{i}}\hat{a}_{\bar{k}_{p}, \sigma_{p}}\chi_{\alpha, \beta, \gamma}
    \notag \\
    e^{i(\omega_{s} + \omega_{i} - \omega_{p})t}\int d^{3}re^{-i(\bar{k}_{s} + \bar{k}_{i} - \bar{k}_{p}).\bar{r}}\bigg] + \text{h.c.}
\end{gather}
where $s, i, p$ denote signal, idler and pump (the three fields involved in this interaction), $\chi_{\alpha, \beta, \gamma} \equiv 3\chi_{\alpha, \beta, \gamma}(\omega_{p} = \omega_{i} + \omega_{s})$ and
\begin{gather}
    \chi_{\alpha, \beta, \gamma}(\omega = \omega' + \omega'') = \int\int dt'dt''\chi_{\alpha, \beta, \gamma}(t', t'')e^{-i(\omega't' + \omega''t'')}
\end{gather}
We now make a few approximations:
\begin{enumerate}
    \item The pump is weak and undepleted. This allows us to treat the pump classically (replace $\hat{a}^{\dagger}_{\bar{k}_{p}}$ with the angular spectrum of the pump), and consider only the first order term of the time-evolution operator, i.e., $U = e^{\bigg(-\frac{i}{\hbar}\int_{0}^{t}d\tau\hat{H}_{I}(\tau)\bigg)} \approx 1 - \frac{i}{\hbar}\int_{0}^{t}d\tau\hat{H}_{I}(\tau)$.
    \item The pump beam propagates along the z-axis and the crystal is large in the transverse directions. This allows us to neglect the phase-matching considerations of the transverse momentum components.
    \item The pump beam only contains one polarization. In our case it is along the extraordinary polarization of our $\beta$-BBO crystal.
    \item We use narrow-band filters (at \SI{710}{nm}) and that the interaction time is long enough so that the only significant contribution at measurement is from $\omega_{i} + \omega_{s} = \omega_{p}$ and $\omega_{i} = \omega_{s}$.
\end{enumerate}
With this, we can write an approximation of $U$ as
\begin{gather}\label{eq:Unitary}
    \notag U = \mathbb{I} + g\sum_{\bar{k}_{s}, \bar{k}_{i}}\nu_{\text{pump}}(\bar{q}_{p})\delta(\bar{q}_{s} + \bar{q}_{i} - \bar{q}_{p})\sinc{\bigg(\frac{(\bar{k}_{s} + \bar{k}_{i} - \bar{k}_{p})_{z}l_{z}}{2}\bigg)\hat{a}^{\dagger}_{\bar{q}_{s}}\hat{a}^{\dagger}_{\bar{q}_{i}}}\\
    = \mathbb{I} + g\sum_{\bar{k}_{s}, \bar{k}_{i}}\nu_{\text{pump}}(\bar{q}_{s} + \bar{q}_{i})\sinc{\bigg(\frac{(\bar{k}_{s} + \bar{k}_{i} - \bar{k}_{p})_{z}l_{z}}{2}\bigg)\hat{a}^{\dagger}_{\bar{q}_{s}}\hat{a}^{\dagger}_{\bar{q}_{i}}}
\end{gather}
where $\bar{q}$ denotes a transverse plane wavevector, $\nu_{\text{pump}}$ is the angular spectrum of the pump inside the crystal, $g$ is the coupling strength and $l_{z}$ is the length of the crystal in the propagation direction. We drop the phase term $e^{i(\frac{(\bar{k}_{s} + \bar{k}_{i} - \bar{k}_{p})_{z}l_{z}}{2})}$ since all the object and image planes are at the fourier plane of the BBO. To simplify the $(\bar{k}_{s} + \bar{k}_{i} - \bar{k}_{p})_{z}l_{z}$ term in equation \ref{eq:Unitary} we'll need to look at the dispersion relations of the three waves. Now, in an anisotropic medium with permittivity tensor $\overset{\scriptscriptstyle\leftrightarrow}{\epsilon}$, Maxwell's equations tell us that
\begin{gather}\label{eq:maxwell_E_field}
    \nabla(\nabla.\bar{E}) - \nabla^{2}\bar{E} = -\mu_{0}\overset{\scriptscriptstyle\leftrightarrow}{\epsilon}\frac{\partial^{2}\bar{E}}{\partial t^{2}}
\end{gather}
Taking a fourier transform of equation \ref{eq:maxwell_E_field}, we get
\begin{gather}\label{eq:maxwell_E_field_ft}
    \bar{k}(\bar{k}.\bar{E}) - \bar{E}(\bar{k}.\bar{k}) = -\mu_{0}\omega^{2}\overset{\scriptscriptstyle\leftrightarrow}{\epsilon}\bar{E}
\end{gather}
In a frame where the crystal axis is the $z$ axis, the permittivity tensor would be diagonal, however, for our purposes we will assume it makes an angle $\theta$ with the $z$ axis. In this frame, one just has to apply a rotational transform to the tensor, which results in
\begin{gather}\label{eq:perm}
    \overset{\scriptscriptstyle\leftrightarrow}{\epsilon} = 
    \begin{pmatrix}
        \epsilon_{o}\cos^{2}\theta + \epsilon_{e}\sin^{2}\theta & 0 & (\epsilon_{e} - \epsilon_{o})\sin\theta\cos\theta \\
        0 & \epsilon_{o} & 0 \\
        (\epsilon_{e} - \epsilon_{o})\sin\theta\cos\theta & 0 & \epsilon_{o}\cos^{2}\theta + \epsilon_{e}\sin^{2}\theta
    \end{pmatrix}
\end{gather}
where $\epsilon_{o}$($\epsilon_{e}$) is the ordinary(extraordinary) permittivity. Using equation \ref{eq:perm} in equation \ref{eq:maxwell_E_field_ft}, we get three linear equations that can be written as $M\bar{E} = 0$ where 
\begin{gather}
    M = 
    \begin{pmatrix} 
        \omega^2\mu_{0}\overset{\scriptscriptstyle\leftrightarrow}{\epsilon}_{xx} - q_y^2 - k_z^2 & q_x q_y & q_x k_z + \omega^2\mu_0 \overset{\scriptscriptstyle\leftrightarrow}{\epsilon}_{xz} \\ 
        q_y q_x & \omega^2\mu_0\overset{\scriptscriptstyle\leftrightarrow}{\epsilon}_{yy} - q_x^2 - k_z^2 & q_y k_z \\ 
        k_z q_x + \omega^2\mu_0\overset{\scriptscriptstyle\leftrightarrow}{\epsilon}_{zx} & k_z q_y & \omega^2\mu_0\overset{\scriptscriptstyle\leftrightarrow}{\epsilon}_{zz} - q_x^2 - q_y^2 
    \end{pmatrix}
\end{gather}
To ensure there exists a non-trivial solution, we need $|M| = 0$. This gives us
\begin{align}\label{eq:determinant}
    \notag|M| = &\left(\omega^2 \mu_0 (\epsilon_o \cos^2\theta + \epsilon_e \sin^2\theta) - q_y^2 - k_z^2\right)
    \times \left[ (\omega^2 \mu_0 \epsilon_o - q_x^2 - k_z^2)(\omega^2 \mu_0 (\epsilon_o \sin^2\theta + \epsilon_e \cos^2\theta) - q_x^2 - q_y^2) - (q_y k_z)^2 \right] \\ \notag
    &  - (q_x q_y)\left[ (q_x q_y)(\omega^2 \mu_0 (\epsilon_o \sin^2\theta + \epsilon_e\cos^2\theta) - q_x^2 - q_y^2) - (q_y k_z)(q_x k_z + \omega^2 \mu_0 (\epsilon_e - \epsilon_o) \sin\theta \cos\theta) \right] \\ \notag
    & + \left(q_x k_z + \omega^2 \mu_0 (\epsilon_e - \epsilon_o) \sin\theta \cos\theta\right) \times \left[ (q_x q_y)(q_y k_z) - (\omega^2 \mu_0 \epsilon_o - q_x^2 - k_z^2)(q_x k_z + \omega^2 \mu_0 (\epsilon_e - \epsilon_o) \sin\theta \cos\theta) \right] \\ 
    = & \quad 0
\end{align}
From equation \ref{eq:determinant} we get the dispersion relations for an ordinary wave
\begin{gather}
    \notag\frac{q_{x}^{2} + q_{y}^{2} + k_{z}^{2}}{\epsilon_{o}} - \mu_{0}\omega^{2} = 0
    \label{eq:ordinary}
\end{gather}
and an extraordinary wave
\begin{gather}
    \frac{(k_{z}\sin{\theta} - q_{x}\cos{\theta})^{2}}{\epsilon_{e}} + \frac{q_{y}^{2}}{\epsilon_{e}} + \frac{( k_{z}\cos{\theta} + q_{x}\sin{\theta})^{2}}{\epsilon_{o}} - \mu_{0}\omega^{2} = 0
    \label{eq:extraordinary}
\end{gather}
From equation \ref{eq:ordinary}, in the paraxial limit, we see that
\begin{gather}
    k_{z} = \sqrt{\bigg(\frac{n_{o}\omega}{c}\bigg)^{2} - \abs{\bar{q}}^{2}} \approx \frac{n_{o}\omega}{c} - \frac{c}{2n_{o}\omega}\abs{\bar{q}}^{2}
    \label{eq:ordi_disp}
\end{gather}
where $n_{o}$($n_{e}$) is the ordinary(extraordinary) refractive index ($n_{i}^{2}\epsilon_{0} = \epsilon_{i}$). From equation \ref{eq:extraordinary}, we see that the roots of the quadratic equation are
\begin{align}
    \notag k_{z} &= \frac{\sin(2\theta)q_{x}\Big(\frac{1}{n_{e}^{2}} - \frac{1}{n_{o}^{2}}\Big) \pm \sqrt{\sin^{2}(2\theta)q_{x}^{2}\Big(\frac{1}{n_{e}^{2}} - \frac{1}{n_{o}^{2}}\Big)^{2} - 4\Big(\frac{\sin^{2}(\theta)}{n_{e}^{2}} + \frac{\cos^{2}(\theta)}{n_{o}^{2}}\Big)\Big(\frac{q_{y}^{2} + q_{x}^{2}\cos^{2}(\theta)}{n_{e}^{2}} + \frac{q_{x}^{2}\sin^{2}(\theta)}{n_{o}^{2}} - \frac{\omega^{2}}{c^{2}}\Big)}}{2\Big(\frac{\sin^{2}(\theta)}{n_{e}^{2}} + \frac{\cos^{2}(\theta)}{n_{o}^{2}}\Big)}\\
    &= q_{x}\frac{(n_{o}^{2} - n_{e}^{2})\sin(\theta)\cos(\theta)}{n_{o}^{2}\sin^{2}(\theta) + n_{e}^{2}\cos^{2}(\theta)} \pm \frac{\omega n_{o}n_{e}}{c\sqrt{n_{o}^{2}\sin^{2}(\theta) + n_{e}^{2}\cos^{2}(\theta)}}\sqrt{1 - q_{x}^{2}\frac{c^{2}}{\omega^{2}(n_{o}^{2}\sin^{2}(\theta) + n_{e}^{2}\cos^{2}(\theta))} - q_{y}^{2}\frac{c^{2}}{2\omega^{2}n_{e}^{2}}}
\end{align}
Defining $\alpha = \frac{(n_{o}^{2} - n_{e}^{2})\sin(\theta)\cos(\theta)}{n_{o}^{2}\sin^{2}(\theta) + n_{e}^{2}\cos^{2}(\theta)}$, $\beta = \frac{n_{o}n_{e}}{n_{o}^{2}\sin^{2}(\theta) + n_{e}^{2}\cos^{2}(\theta)}$, $\gamma = \frac{n_{o}}{\sqrt{n_{o}^{2}\sin^{2}(\theta) + n_{e}^{2}\cos^{2}(\theta)}}$, $\eta = \frac{n_{o}n_{e}}{\sqrt{n_{o}^{2}\sin^{2}(\theta) + n_{e}^{2}\cos^{2}(\theta)}}$ and using the paraxial approximation, we see that
\begin{gather}\label{eq:extraordi_disp}
    k_z \approx \alpha q_{x} \pm \Big(\frac{\eta\omega}{c} - \frac{\beta^{2}c}{2\eta\omega}q_{x}^{2} - \frac{\gamma^{2}c}{2\eta\omega}q_{y}^{2}\Big)
\end{gather}
In our case since we consider Type I SPDC, we have $e \rightarrow o + o$ conversion, giving us
\begin{gather}\label{eq:signal_dispersion}
    (\bar{k}_{s})_{z} = \frac{n_{o}(\omega_{s})\omega_{s}}{c} - \frac{c}{2n_{o}(\omega_{s})\omega_{s}}\abs{\bar{q}_{s}}^{2}
\end{gather}
\begin{gather}\label{eq:idler_dispersion}
    (\bar{k}_{i})_{z} = \frac{n_{o}(\omega_{i})\omega_{i}}{c} - \frac{c}{2n_{o}(\omega_{i})\omega_{i}}\abs{\bar{q}_{i}}^{2}
\end{gather}
\begin{gather}
    (\bar{k}_{p})_{z} = \alpha q_{px} + \frac{\eta\omega_{p}}{c} - \frac{\beta^{2}cq_{px}^{2}}{2\eta\omega_{p}} - \frac{\gamma^{2}cq_{py}^{2}}{2\eta\omega_{p}}
\end{gather}
\begin{figure}[H]
    \centering
    \includegraphics[width = 0.75\textwidth
    ]{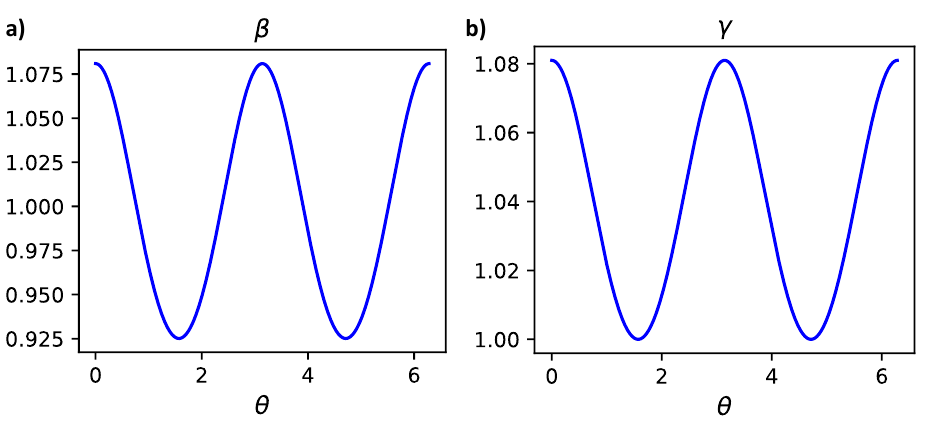}
    \caption{\textbf{Phasematching function terms. a,} $\beta$ as a function of $\theta$. \textbf{b,} $\gamma$ as a function of $\theta$. All the curves were generated using the pump wavelength (\SI{355}{nm}) and the sellmeier equations.}
    \label{fig:S5}
\end{figure}
Another set of assumptions we can make is that of approximating both $\beta^{2}$ and $\gamma^{2}$ as 1 (see Figure \ref{fig:S5}). We note that this approximation doesn't need to be made, however we find that for the purposes of training our system, this small mismatch makes no difference. Thus,
\begin{gather}\label{eq:pump_dispersion}
    (\bar{k}_{p})_{z} \approx \frac{\eta\omega_{p}}{c} + \alpha q_{px} - \frac{cq_{px}^{2}}{2\eta\omega_{p}} - \frac{cq_{py}^{2}}{2\eta\omega_{p}}
\end{gather}  
With equations \ref{eq:signal_dispersion}, \ref{eq:idler_dispersion} and \ref{eq:pump_dispersion}, and in the degenerate SPDC limit ($\omega_{s} = \omega_{i} = \frac{\omega_{p}}{2} = \omega$) we can rewrite the $\Delta k_{z}$ term as
\begin{gather}
    (\bar{k}_{s} + \bar{k}_{i} - \bar{k}_{p})_{z} = \frac{2\omega\big(n_{o}(\omega) - \eta(2\omega)\big)}{c} - \alpha(q_{sx} + q_{ix}) + \frac{c}{2\omega}\bigg(\frac{(\abs{\bar{q}_{s} + \bar{q}_{i}}^{2}}{2\eta(2\omega)} - \frac{\abs{\bar{q}_{s}}^{2} + \abs{\bar{q}_{i}}^{2}}{n_{o}(\omega)}\bigg)
\end{gather}
Defining $\mu_{00} = \frac{2\omega\big(n_{o}(\omega) - \eta(2\omega)\big)l_{z}}{c}$, $\delta = \frac{c}{4n_{o}(\omega)\omega}$ and approximating $\eta(2\omega) \approx n_{o}(\omega)$ (true for critical phasematching), we get
\begin{gather}\label{eq:phasematching}
    \sinc{\bigg(\frac{(\bar{k}_{s} + \bar{k}_{i} - \bar{k}_{p})_{z}l_{z}}{2}\bigg)} = \xi(\bar{q}_{s}, \bar{q}_{i}) \approx \sinc{\bigg(\mu_{00} - \alpha l_{z}(q_{sx} + q_{ix}) + \delta l_{z}\abs{\bar{q}_{s} - \bar{q}_{i}}^{2}\bigg)}
\end{gather}
From equation \ref{eq:phasematching}, we can see that there are only three terms to be fit from experimental data to be able to model our correlated light source reasonably accurately. Going back to equation \ref{eq:Unitary}, we can write
\begin{gather}\label{eq:Heisenberg}
    U = \mathbb{I} + \sum_{\bar{q}_{s}, \bar{q}_{i}}\bigg(\epsilon_{\bar{q}_{s}, \bar{q}_{i}} a^{\dagger}_{\bar{q}_{s}}a^{\dagger}_{\bar{q}_{i}} - \epsilon^{*}_{\bar{q}_{s}, \bar{q}_{i}}a_{\bar{q}_{s}}a_{\bar{q}_{i}}\bigg)
\end{gather}
where $\epsilon_{\bar{q}_{s}, \bar{q}_{i}} = g\nu_{\text{pump}}(\bar{q}_{s} + \bar{q}_{i})\xi(\bar{q}_{s}, \bar{q}_{i})$. Now we look at how the field operators evolve (Heisenberg picture). Only keeping terms to first order in $|\epsilon|$ and replacing indices\footnote{When imaging the fourier plane of the $\beta$-BBO crystal, the indices $\bar{q}_{s}, \bar{q}_{i}$ (i, j) are just camera pixels. From here on, we will interchangeably use $\bar{q}_{s}, \bar{q}_{i}$ and i, j to refer to these.} $\bar{q}_{s}, \bar{q}_{i}$ with $p, q$, we get
\begin{gather}
    U^{\dagger}a_{k}U = \bigg(\mathbb{I} + \sum_{p, q}\epsilon^{*}_{p, q} a_{q}a_{p} - \epsilon_{p, q} a^{\dagger}_{q}a^{\dagger}_{p}\bigg)a_{k}\bigg(\mathbb{I} + \sum_{p, q}\epsilon_{p, q} a^{\dagger}_{p}a^{\dagger}_{q} - \epsilon^{*}_{p, q}a_{p}a_{q}\bigg) 
    \notag\\
    = a_{k} + \sum_{p, q}\bigg(\epsilon_{p,q}\Big(a_{k}a^{\dagger}_{p}a^{\dagger}_{q} - a^{\dagger}_{p}a^{\dagger}_{q}a_{k}\Big) + \epsilon^{*}_{p, q}\Big(-a_{k}a_{p}a_{q} + a_{p}a_{q}a_{k}\Big)\bigg)
\end{gather}
Noting that $	a_{k}a^{\dagger}_{p}a^{\dagger}_{q} - a^{\dagger}_{p}a^{\dagger}_{q}a_{k} = \delta_{pk}a_{q}^{\dagger} + \delta_{qk}a_{p}^{\dagger}$ and $a_{p}a_{q}a_{k} - a_{k}a_{p}a_{q} = 0$, we see
\begin{gather}\label{eq:Unitary_heisenberg}
    U^{\dagger}a_{k}U = a_{k} + \sum_{i}\epsilon_{p,k}a_{p}^{\dagger} + \sum_{q}\epsilon_{k,q}a_{q}^{\dagger} = \sum_{p}a_{p}\delta_{pk} + 2\epsilon_{p,k}a_{p}^{\dagger}
\end{gather}
Comparing equation \ref{eq:Unitary_heisenberg} to equation \ref{eq1}, we see that
\begin{gather}
    C_{\bar{q}_{s}\bar{q}_{i}} = \delta_{\bar{q}_{s}\bar{q}_{i}}
\end{gather}
\begin{gather}\label{eq:spdc_S}
    S_{\bar{q}_{s}\bar{q}_{i}} \propto \nu_{\text{pump}}(\bar{q}_{s} + \bar{q}_{i})\sinc{(\mu_{00} - \alpha l_{z}(q_{sx} + q_{ix}) + \delta l_{z}\abs{\bar{q}_{s} - \bar{q}_{i}}^{2})}
\end{gather}

\subsection{SPDC mean field and photon correlations}\label{SPDC_correlations}
The mean field of the SPDC wavefunction at the fourier plane of the $\beta$-BBO crystal (which is also imaged onto the camera) is given by (also see equation \ref{eq2} in the main text)
\begin{gather}
    \ev{n_{k}} = \ev{\big(a^{\text{out}}_{k}\big)^{\dagger}a^{\text{out}}_{k}}{0}
\end{gather}
From equation \ref{eq1}, we can write
\begin{gather}
    \big(a^{\text{out}}_{k}\big)^{\dagger}a^{\text{out}}_{k} = \sum_{j, m}C_{km}^{*}C_{kj}\big(a^{\text{in}}_{m}\big)^{\dagger}a^{\text{in}}_{j} + C_{km}^{*}S_{kj}\big(a^{\text{in}}_{m}\big)^{\dagger}\big(a^{\text{in}}_{j}\big)^{\dagger} + S_{km}^{*}C_{kj}a^{\text{in}}_{m}a^{\text{in}}_{j} + S_{km}^{*}S_{kj}a^{\text{in}}_{m}\big(a^{\text{in}}_{j}\big)^{\dagger}
\end{gather}
It's clear that the only the last term survives, thus
\begin{gather}\label{eq:mean_field}
    \ev{n_{k}} = \ev{\big(a^{\text{cam}}_{k}\big)^{\dagger}a^{\text{cam}}_{k}}{0} = \sum_{j}|S_{kj}|^{2} = (SS^{\dagger})_{kk}
\end{gather}
For the correlations, the algebra is a little more complicated, but we know that
\begin{gather}\label{eq:correlations}
    \big(a^{\text{out}}_{k}\big)^{\dagger}a^{\text{out}}_{k} \big(a^{\text{out}}_{l}\big)^{\dagger}a^{\text{out}}_{l} 
    \notag\\
    = \sum_{j, m, p, q}\bigg(C_{kj}^{*}\big(a^{\text{in}}_{j}\big)^{\dagger} + S_{kj}^{*}a^{\text{in}}_{j}\bigg)\bigg(C_{km}a^{\text{in}}_{m} + S_{km}\big(a^{\text{in}}_{m}\big)^{\dagger}\bigg)
    \bigg(C_{lp}^{*}\big(a^{\text{in}}_{p}\big)^{\dagger} + S_{lp}^{*}a^{\text{in}}_{p}\bigg)\bigg(C_{lq}a^{\text{in}}_{q} + S_{lq}\big(a^{\text{in}}_{q}\big)^{\dagger}\bigg)
\end{gather}
The terms that survive the sandwich between $\ket{0}$ need to be of the form $a...a^{\dagger}$ with an equal number of rasing and lowering operators. Thus,
\begin{gather}\label{eq:correlations_ev}
    \expval{n_{k}n_{l}} - \expval{n_{k}}\expval{n_{l}} = \bigg(\Big(S^{*}S^{\top}\odot CC^{\dagger}\Big) +\Big(S^{*}C^{\dagger}\odot CS^{\top}\Big)\bigg)_{kl}
\end{gather}
For SPDC, the first term in equation \ref{eq:correlations_ev} (or equation \ref{eq3}) contributes to the variance, and the second term is the two-body correlations. 

\subsection{The biphoton wavefunction and join probability distribution}\label{SPDC_jpd}
From equation \ref{eq:Unitary}, we can write down the unnormalized wavefunction at the fourier plane of the $\beta$-BBO crystal (what the camera is imaging) as
\begin{gather}\label{eq:SPDC_wavefunction}
    \ket{\Psi} = \ket{0} + \sum_{i, j}S_{ij}a^{\dagger}_{i}a^{\dagger}_{j}\ket{0}
\end{gather}
For our simulation purposes, we just want the probabilities of each photon pair, so we just normalize the bi-photon term. Keeping in mind that $S$ is a symmetric matrix in our case, we get
\begin{gather}
    \ev{\sum_{i, j, p, q}S^{*}_{pq}S_{ij}a_{q}a_{p}a^{\dagger}_{i}a^{\dagger}_{j}}{0} = 2\text{Tr}[S^{\dagger}S] = 1
\end{gather}
Now defining $J = \frac{S}{\sqrt{2\text{Tr}[S^{\dagger}S]}}$, we can find the probabilities for a biphoton event as
\begin{gather}
    P(\ket{1}_{i}\ket{1}_{j}) = \abs{J_{ij} + J_{ji}}^{2} = \frac{2\abs{S_{ij}}^{2}}{\text{Tr}[S^{\dagger}S]}
\end{gather}
\begin{gather}
    P(\ket{2}_{i}) = 2\abs{J_{ii}}^{2} = \frac{\abs{S_{ii}}^{2}}{\text{Tr}[S^{\dagger}S]}
\end{gather}

\subsection{Fitting the phasematching function to experimental measurements}\label{PM_fitting}
To fit the phasematching function we recorded illumination patterns (mean fields and correlations) produced by different SLM patterns and used equation \ref{eq:mean_field} to minimize the difference between the simulated and predicted illumination patterns. We will outline how the simulation was performed in this section, and all code and collected data is available on GitHub and Zenodo (see \ref{code_links}).

\begin{figure}[H]
    \centering
    \includegraphics[width=\linewidth]{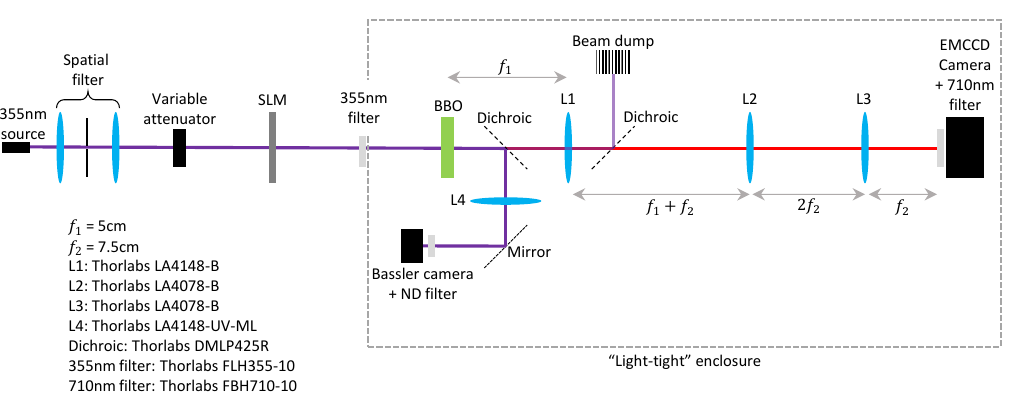}
    \caption{\textbf{Detailed schematic of the experimental setup.}}
    \label{fig:S8}
\end{figure}

Figure \ref{fig:S8} shows a schematic of the setup used to record the various illuminations. The EMCCD camera views the fourier plane of the BBO crystal (where the SPDC photons are generated), i.e., a distance of \SI{5}{cm} away from lens L1. The dimension of each frame recorded on the camera was $37 \times 37$. The angular spectrum of the \SI{355}{nm} pump was shaped using a spatial light modulator (SLM; Holoeye PLUTO-2.1-UV-099B SLM). The pump beam covered roughly $328 \times 328$ pixels on the SLM and these were used to apply phases to the pump beam's wavefront. The angular spectrum of the pump inside the crystal for each pattern was estimated by projecting the correlations of the down-converted onto a basis spanned by $q_{sx} + q_{ix}$ and $q_{sy} + q_{iy}$ (we note here that a bassler camera was used to record the angular spectrum of the pump in free space). As we saw from equation \ref{eq:spdc_S}, there are only a few parameters needed to fit the green's function, namely $\mu_{00}$, $\alpha$ and $\delta$, and a constant of proportionality. We fit these parameters by simulating equation \ref{eq:mean_field} in pytorch and minimizing the L2 norm between the simulated and experimentally measured mean fields. 

\begin{figure}[H]
    \centering
    \includegraphics[width=0.23\linewidth]{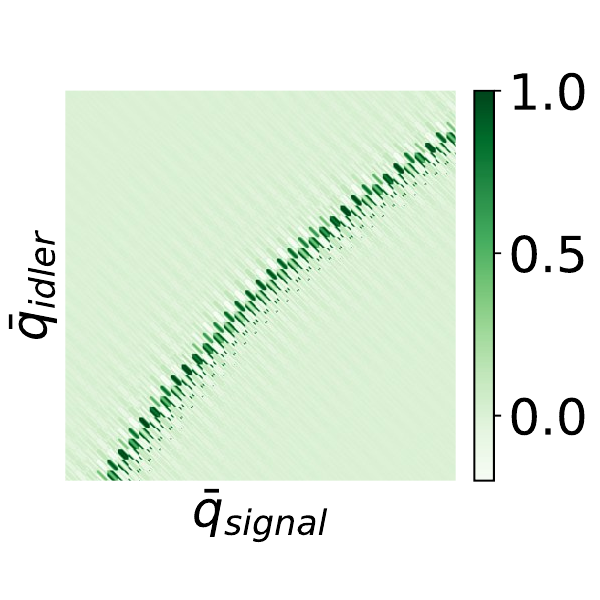}
    \caption{\textbf{Phasematching function.}}
    \label{fig:S9}
\end{figure}

\begin{figure}[H]
    \centering
    \includegraphics[width=\linewidth]{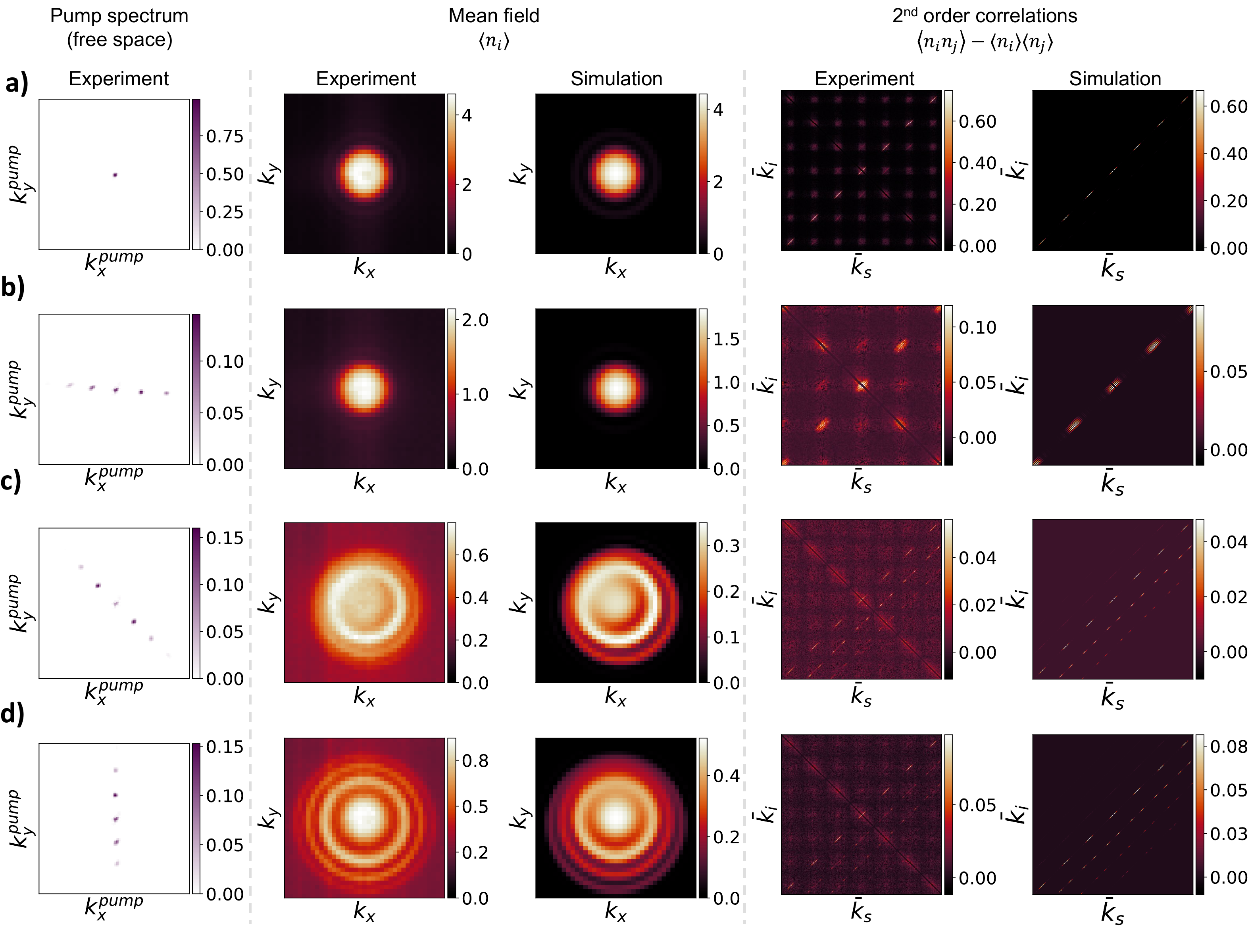}
    \caption{\textbf{A digital model of the correlated light source.} }
    \label{fig:S10}
\end{figure}

Figure \ref{fig:S9} shows a 2-dimensional representation of the phase matching function (equation \ref{eq:phasematching}) fit for our particular experiment (also see the inset in Figure \ref{fig:light_source}). The x and y axes in figure \ref{fig:S9} and all other photon correlation plots are the camera frame pixels unraveled. The dimension of this 2-dimensional representation is $37^{2} \times 37^{2}$. Figure \ref{fig:S10} shows the agreement between simulation and experiment of SPDC illumination patterns produced by different pump spectra.

\subsection{Camera gain calibration}\label{cam_gain}
The camera used in experiments was an electron-multiplying charge coupled device (EMCCD) camera. The specific model was Princeton Instrument's ProEM-HS: 1024B eXcelon (1024x1024 CCD). The camera was operated in 8x8 pixel binning mode. The analog gain of the camera was set to high, and the manufacturer certificate of performance indicated that at high analog gain the electrons/count were about $1.85$ when operating at \SI{10}{MHz}. The EM gain of the camera was set at it's maximum value of 1000. To calibrate the EM gain of the camera, we followed the procedure described in ref \cite{avella2016absolute}. The main assumption is that any non-camera-read noise counts come from random spurious charges that are equivalent to less than one photon of excitation. Under this assumption, we can use an exponential decay to model the camera counts.

\begin{figure}[H]
    \centering
    \includegraphics[width=\linewidth]{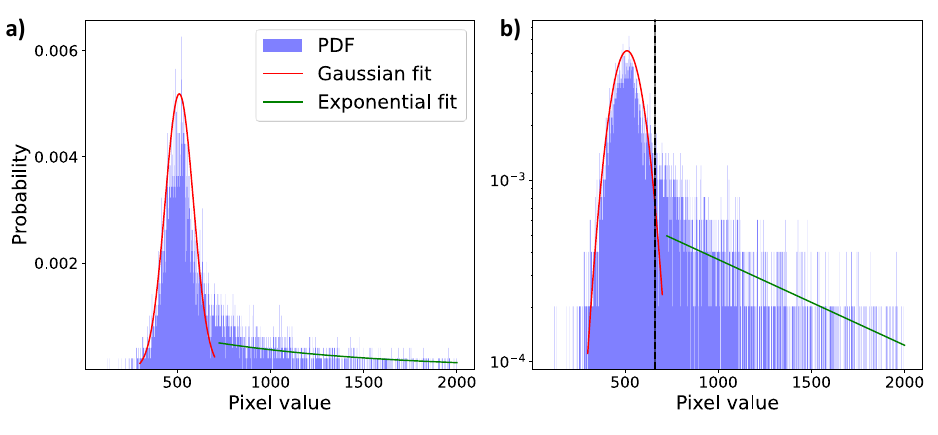}
    \caption{\textbf{Histogram of pixel values for a single pixel across 5000 background frames. a,} The histogram (represented as a probability density) of pixel values for a randomly chosen pixel across 5000 frames. The red line depicts the gaussian fit and the green line depicts the exponential fit for the gain. \textbf{b, } A log-scale representation of the histogram plotted in \textbf{a} for better visualization of the exponential fit. The black dotted line depicts the mean value of the pixel across the 5000 frames, and the gray line depicts the value of $\mu + 2\sigma$ from the gaussian fit.}
    \label{fig:S6}
\end{figure}

We took 5000 frames (the exposure time was set to be \SI{1}{ms}) with no illumination and fit a gaussian curve and an exponential tail to the histogram of counts for each pixel. The gaussian curve which models the camera read noise was fit using
\begin{gather}
    Ae^{-\frac{(x - \mu)^{2}}{\sigma^{2}}}
\end{gather}
where $A$, $\mu$ and $\sigma$ are constants. The exponential tail was fit using
\begin{gather}
    \frac{1}{g}e^{-\frac{x}{g}}
\end{gather}
where $g$ is the EM gain of the camera. Figure \ref{fig:S6} shows the fit to the a histogram of pixel counts for a randomly chosen pixel with no illumination and the shutter open. Doing this for all pixels, we found that the fits to the exponential tail gave us a gain value of $\sim 940$. Since this was close to the manufacturer's specifications, we used an EM gain value of 1000 for all post-processing. 

\subsection{Post-processing of raw camera frames}\label{camera_frame_postprocessing}
\begin{figure}[H]
    \centering
    \includegraphics[width=\linewidth]{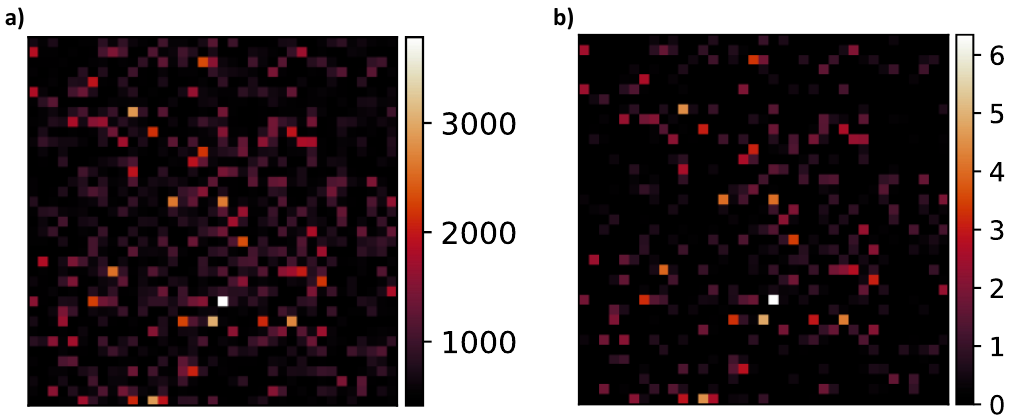}
    \caption{\textbf{Conversion of raw camera frame pixel values to estimated photon counts. a,} A raw camera frame recorded at an EM gain of 1000 with an exposure time of \SI{1}{ms}. \textbf{b, } The processed raw frame turned into estimated photon counts using equation \ref{eq:count_from_raw}.}
    \label{fig:S7}
\end{figure}

To estimate the number of photons incident on a pixel from the 16-bit value that the camera outputs, we apply a simple transformation to each frame. First, a mean background level is estimated by collecting a large number (in our experiments this number was 5000) of frames with all light sources blocked and the shutter open (we'd like to point out that a background level was taken before every data collection run or training run). The average frame, $\mathbf{x}_{\text{bg}}$ is computed. This is then subtracted from each frame taken during data collection (see the dashed-black line in \ref{fig:S6}b) and any negative values are set to 0. After subtraction, we divide by the gain and the manufacturer reported quantum efficiency ($\sim 90.5$\% at \SI{710}{nm}), and multiply with the electrons/count. Thus, the processed frame, $\mathbf{x}$ is
\begin{gather}\label{eq:count_from_raw}
    \mathbf{x} = \frac{(\mathbf{x}_{\text{raw}} - \mathbf{x}_{\text{bg}})e_{\text{D}}}{\eta g}
\end{gather}
where $e_{\text{D}}$ is the electrons/count, $g$ is the EM gain (see section \ref{cam_gain}) and $\eta$ is the quantum efficiency. Figure \ref{fig:S7} shows an example of a raw frame it's processed frame. This was the only processing done before the data was fed to the set transformers. 

\section{End-to-end optimization}\label{training}
In this section we describe the training framework, correlation-aware training (CAT), that we used for training on the experiment (Figure \ref{fig:spdc_training}) and in simulation (Figure \ref{fig:cell_classification}). We used the same code for both except than in experiment we use the actual physical system for the forward pass (from the laser light source down to the detection at the camera plane) while in simulation we use the digital twin for the simulated stochastic forward pass.

Our framework rely on three main components:
\begin{itemize}
    \item The physical system itself
    \item A differentiable digital twin of the physical system
    \item A digital post-processing model (a Transformer).
\end{itemize}

By chaining each of those components, we built a end-to-end differentiable model, similarly to a standard digital neural network. This allows us to optimize a desire loss function at the very end of the model (classification accuracy in our case) with standard gradient descent, viewing the entire system as a differentiable hybrid physical-digital neural network.

\begin{figure}[H]
    \centering
    \hspace*{-0.9cm}
    \includegraphics[width = 1.2\textwidth]{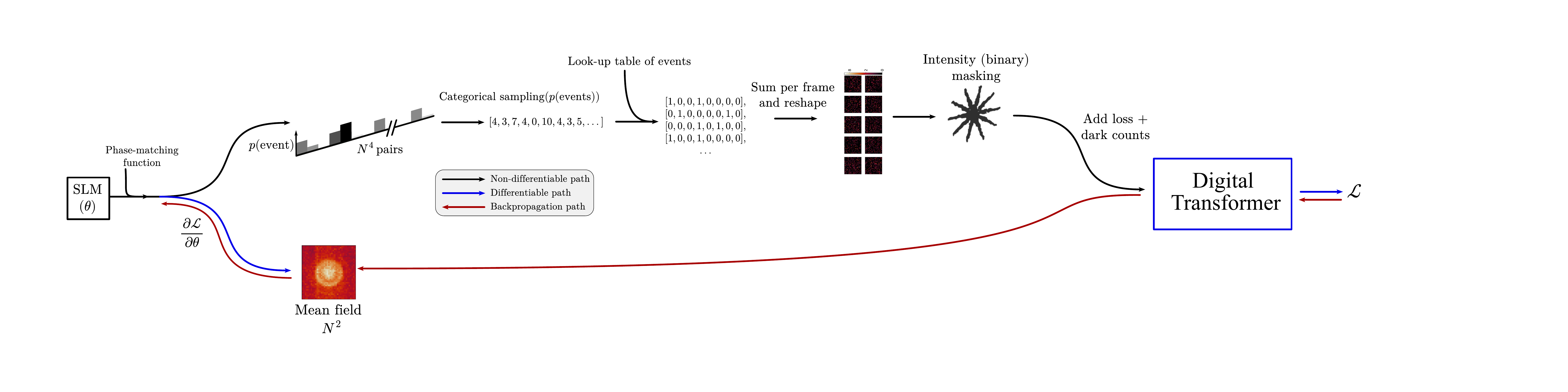}
    \caption{\textbf{Training pipeline.} We chain our differentiable sampler with a digital model (Transformer), that allows us to trained the entire model end-to-end with backpropagation.}
    \label{fig:S29}
\end{figure}

We detail here the implementations details of the last two blocks, the first block having been described in previous sections.

\subsection{Differentiable sampling}
Training the physical system with gradient descent requires to be able to compute the gradient of the final cost function with respect to the physical parameters. However, it's not easy to get those gradients directly: we can just run backpropagation on a given physical system (especially on ours) and zero-order perturbative methods hardly scale to the number of free parameters we have access to in our experiment.

We therefore adopted a Physics-Aware-Training (PAT) \cite{wright2022deep} approach. In PAT, the physical setup is used only for the forward pass, while a differentiable “digital twin” is used to propagate gradients backward to the trainable parameters. Unlike the common choice of a neural-network surrogate trained on input–output pairs, our twin is a first-principles physical simulation of the experiment, calibrated on experimental data. In particular, we fit the phase-matching function of the SPDC source to reproduce the measured joint emission pattern (\ref{PM_fitting}).

Our differentiable sampler is a multi-steps function that maps a given SLM patterns (ie the trainable parameter) to a camera frame, that is, a collection of pair of photons emission events. 

The first step of the sampler is a physical "simulation". We compute, given the trainable SLM pattern two quantities: 
\begin{enumerate}
    \item The mean field of the photons emissions - we use this 2D tensor, constructed in a differentiable manner from the SLM pattern, as the differentiable path for the backward pass.
    \item The probability distribution over all possible events (ie emission of a pair of photons) in the $N \cdot N$ frame (results in $N^4$ possible pairs of photons. Obviously not all pairs are physically realizable, but this is taken into account through the phase matching function, which (almost, down to machine precision) zero out the probability of impossible events).
\end{enumerate}

While propagating only through the mean field does not ensure we will train for correlations at all we showed in \ref{increase_in_correlations} that it indeed lead to increased and more "useful" correlations while we use this training scheme. Because we use sampled frames for the forward pass, the digital post-processing model is actually trained to classify the data based on correlated frames so during the backpropagation step, the gradients of the post-processing model take into account for the correlations in the different frames within a set of frames.

Another crucial aspect of our scheme is to allow for the sampling step to be computational tractable. While computing the $N^4$ dimensional probability vector is tolerable if done once per forward pass, as in practice we train on a mini-batch (size $B$) of data, and each data point correspond to set of frames (size $S\times N \times N$), this would result in an input tensor of size $B \times S \times N^4$ for the categorical sampling function, which increases the RAM used (for typical values used for training: $B=16$, $S=100$, $N=37$, this results in a tensor of $\approx 12$Gb, which will freeze an experimental computer with small RAM). Fortunately, because we mask the emitted photons \textit{after} the sampling step, we can sample at \textit{once} all the frames  required for a mini-batch of data (ie $B \times S \times N_{\text{events}}$, where $N_{\text{events}}$ is the number of pairs of photons we expect on a given frame. At this step, we only sample \textit{indices} of events stored in a look-up table (of size $N^{4} \times N^{2}$, stored as a Boolean tensor to save memory as it stores only a pair of emitted photon on the $N\times N$ grid). To convert those indices to actual sampled frames, we just retrieve the events in the look-up table at the sampled indices. Then, we sum over the $N_{\text{events}}$ dimension and reshape so the final tensor has shape $B \times S \times N \times N$. The actual code used is detailed in the forward function of Alg. \ref{alg:ste_sampling}.

One should note that this forward sampling is only necessary for when we train in simulation. For experiments where we trained the actual physical system with PAT, we don't need to do this sampling step, we just need to use the experimentally sampled frames as the forward frames to feed to the transformer, and use the backward method we used for backpropagation.

\begin{algorithm}[H]
\caption{PyTorch pseudo-code for the Straight-Through Estimator (STE) used to bypass the high-dimensional probability bottleneck.}
\label{alg:ste_sampling}
\begin{lstlisting}[language=Python, frame=none, basicstyle=\small\ttfamily, numbers=left]
class DifferentiablePhotonSampling(torch.autograd.Function):
    """
    Custom autograd function to enable backpropagation through 
    discrete photon sampling via the Straight-Through Estimator.
    """

    def forward(ctx, mean_field, events_lookup_table, data, set_size, n_events_per_frame):
        # Detach from graph to save memory during the expensive N^4 computation
        with torch.no_grad():
            # 1. Compute the high-dimensional probability distribution
            # Shape: [1, N^4]
            probs = compute_physical_probs(mean_field)

            # 2. Compute the number of samples per mini-batch of data: B * S * n_events_per_frame
            num_samples = data.size(0) * set_size * n_events_per_frame

            # 3. Perform discrete categorical sampling
            # Indices represent specific photon pair events
            # We amortize the cost of creating the huge probability vector by sampling the total number of events required for the current mini-batch at once
            indices = torch.multinomial(probs, num_samples, replacement=True)

            # 4. Construct frames by aggregating events
            # We map indices to physical pixel locations and sum them
            # Result Shape: [Batch, Set_Size, Pixels, Pixels]
            sampled_frames = events_lookup_table(indices).sum(dim=2)

        return sampled_frames

    @staticmethod
    def backward(ctx, grad_output):
        """
        Backward pass: The Straight-Through Estimator.
        We treat the sampling + physical transform as the identity function.
        """
        # grad_output shape: [Batch, Set_Size, Pixels, Pixels]
        # we want grad_mean_field shape to be: [Pixels, Pixels]
        
        # We propagate gradients directly from the sampled frames to the 
        # mean field, bypassing the N^4 probability vector entirely.
        
        # Average gradients over Batch and Set dimensions to match mean_field
        grad_mean_field = grad_output.mean(dim=(0, 1))

        return grad_mean_field, None, None, None, None
\end{lstlisting}
\end{algorithm}

\subsection{Transformer as a digital backend}

As mentioned in the main text, we used a Transformer as the digital backend used to process the noisy frames, based on the set-Transformer model developed in \cite{lee2019settransformer}. We chose the set-Transformer for the following reason. The signal is encoded in radially symmetric pairs of pixels. The presence of one pixel is informative of the presence of its paired counterpart. This induces long-range correlations between many structured pixel pairs. While other U-Net-like architectures can capture such patterns to some extent, they rely on local propagation of information and are therefore poorly matched to this setting. In contrast, transformers capture such long-range interactions naturally through attention, without requiring repeated local propagation across the image. The Transformer we used is described as follows (this architecture was used for all experiments and simulations):
\begin{itemize}
    \item Each frame is flattened to a $N\times N$ vector and use a "token" by the Transformer. The "sequence" dimension being the set dimension.
    \item Full self-attention: this allows the model to be permutation invariant about the frames contained in the input set and the model to learn desired properties about the set of frames.
    \item Embedding size of 512 and 2 attention layers: we perform a manual architecture search, and this set of hyperparameters seems to be ideal on our data. Constraining the model to 2 layers constrains the model to only learn correlation up to second order - more layers would allow for higher order - but since SPDC biphotons only have second order correlations, we did not use more. Using more layers actually made the model to overfit too much and was not as good as the model with 2 layers on the est data.
    \item A last attention-based pooling layer: the Transformer process a set of frames, that is a tensor of size $S \times N^2$. But as we do classification, we need to collapse the sequence dimension to 1. And because we use full self-attention (and not causal attention), we can't use the last token as the input of the final classifier. Instead we use a attention-based pooling where a single query token is learnt throughout the training process, and used to pool the relevant information contained in all the tokens after the last attention layer.
    \item A 1 hidden layer (128 neurons) Multi-Layer Perceptron classifier attached after the attention-based pooling layer that project the $1, N^2$ vector to the final loggits used for classification (size $1, N_{\text{classes}})$.
\end{itemize}

While training simultaneously the physical parameters and the digital backend is in theory ideal, we empirically found that this simultaneous training was unstable. We conjectured that is was because the physical parameters were updating too fast compared to the digital parameters and the Transformer couldn't catch up with the frames collected with the updated physical parameters. We decided to implement two fixes:
\begin{itemize}
    \item No longer update the physical parameters and the digital backend at the same time but instead we train the physical parameters for one epoch while keeping the digital parameters frozen. Then, for the next 5 epochs we do the opposite: we only optimize the digital parameters while keeping the physical parameters frozen. And we repeat this loop for many epochs.
    \item We used two different optimizers for the physical parameters (we use SGD here) and the digital parameters (Adam). There is not formal reason for that other than it worked better.
\end{itemize}

\subsection{Data loading}

We detail here the two strategies we used for loading the training and test data. The first strategy is used for when we only post-process experimental frames collected on the experiment with untrained illuminations. The second strategy is used when we train with the experiment in the loop (and when we do end-to-end training in simulation too). 

The data collected on the experiment with untrained illumination has the following size: ($N_{classes} \times N_{\text{images per class}} \times N_{\text{frames per image}}\times N^2$ where $N^2$ ) (where $N_{\text{images per class}}$ depends on whether we deal with the training or testing set. In this work, experimentally we have $N_{\text{images per class}}=3$ for training images and $N_{\text{images per class}}=2$ for test images). Because most of the times the set size is smaller than the number of frames per image per class, we have to devise a sampling strategy that allows for using the relatively scarce data in order to train the best model as possible. We wrote a custom sampler class in PyTorch so that for each training epoch we sample - without replacement - frames for each data point to form sets having the desired set size. Per mini batch we can have multiple sets per data point, but none share any frames because of the sampling without replacement. We do the same at test time which results in what we could get an actual experiment: sets of random frames. But because the data sampling step introduces randomness in the test set being evaluated, we average the test accuracy over many evaluations on different test sets sampled from the same underlying experimental data.

For the experiments where we train the illumination jointly with the transformer, we also wrote a custom PyTorch sampler. However in that case, we don't sample frames because the forward pass (simulated or experimental) produce the frames, we sample with replacement the objects to be sensed. Even if one object is sample twice, that will results in different sets due to the randomness of the physical sampling step (simulated or experimentally realized). 

In the code used for training on the experiment, what we really sample is the actual (x,y) position of the object to be sensed and a number of frames to collect for this specific object in that specific mini batch (ie. $N_{\text{sets}} \times \text{set size}$. The reason for that is because the motorized stage we used to place the object to be sensed on the optical axis is slow, so we want to move as few as possible the objects per mini batch. The strategy we used allows to scan only once all of the objects per mini batch, sample $N_{\text{sets}} \times \text{set size}$ for the current $N_{\text{sets}}$ for a specific object in that mini batch, and before digital post-processing we recast the sampled frames in actual sets. This allowed to speed up a lot the experimental training. 

In the simulations for Fig. \ref{fig:cell_classification}, we added data augmentation (random x-y translation) at training time (standard) but also at test time (not standard) to simulate an actual scenario where the cells wouldn't be centered on the scene but rather would be located at random locations in the field of view. Because of the random augmentation, we repeat the testing 50 times (i.e. on virtually 50 close but different test dataset, each with a different random seed) and return the average test accuracy on those 50 instances to average out the randomness coming from the test-time augmentation used in this setting.

\subsection{CAT increases correlations}\label{increase_in_correlations}
To show that CAT, with its straight-through estimator and backpropagation through the mean-field, still trains the correlations between the photons, we consider the following simulation. CAT was used to optimize the sensing pipeline to be able to distinguish between 5 object classes see Appendix \ref{cell_training}. Two illumination sources were considered: (1) untrained correlated SPDC bi-photons, and (2) trained correlated SPDC bi-photons from a source that had arbitrarily programmable green's functions (again, see Appendix \ref{cell_training}). Figure \ref{fig:S37} shows the results of how CAT can increase the number of correlated photons transmitted through the different classes of objects.

\begin{figure}[H]
    \centering
    \includegraphics[width = \textwidth]{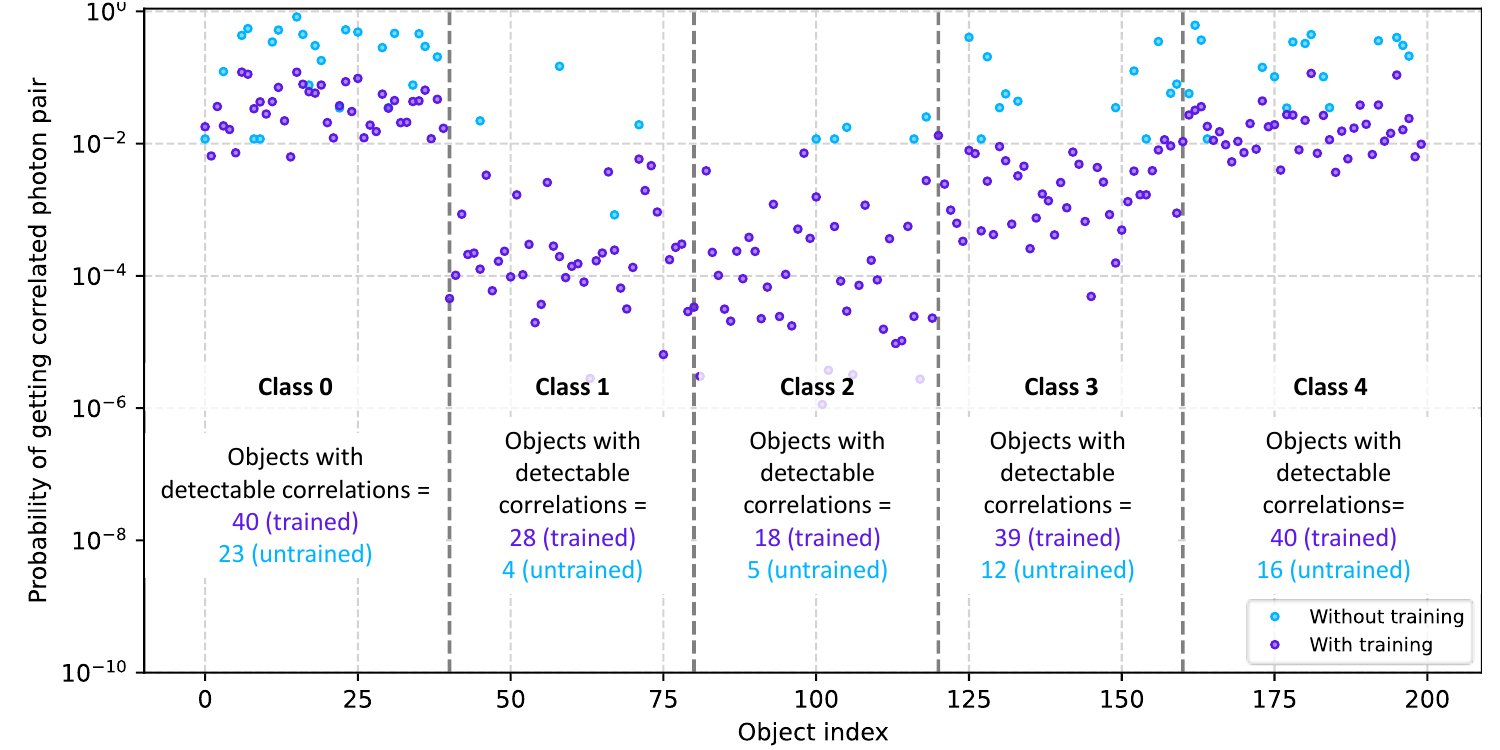}
    \caption{\textbf{CAT increases detectable correlations.} Simulation results of how CAT can increase the number of correlated photons transmitted through different objects.}
    \label{fig:S37}
\end{figure}

The points in light blue show the probabilities of getting at least one correlated photon pair at the camera with untrained illumination for each object in the test dataset, i.e., objects that were not used during training. The points in purple show the same probabilities after the illumination was trained CAT. We see that the probability of getting a correlated photon pair are significantly increased per object despite not using the explicit probability distribution during backpropagation. These simulations suggest that our proposed training method, which effectively treats the gradient of the sampling operation as the identity function, proves effective in practice.

\section{Photon-budget-efficient image sensing}

\subsection{MPEG7 dataset and object fabrication}\label{MPEG7_task}
\begin{figure}[H]
    \centering
    \includegraphics[width = 0.75\textwidth]{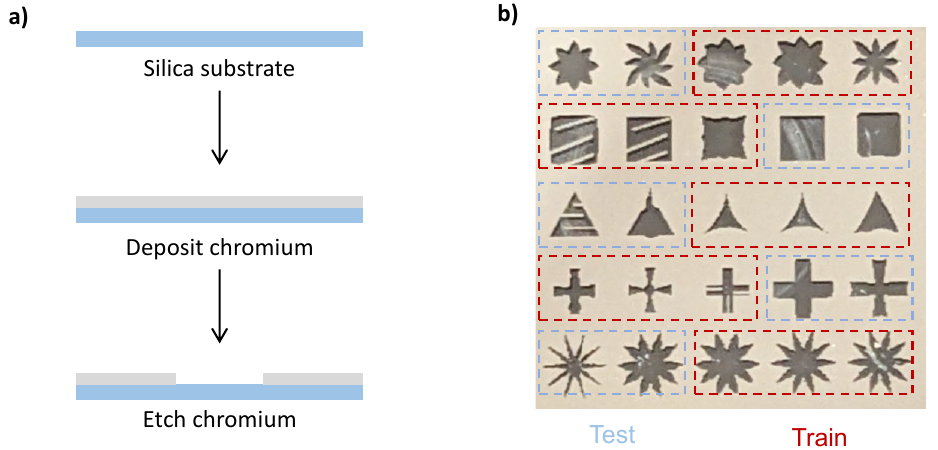}
    \caption{\textbf{Fabircated MPEG7 objects.} A subset of the MPEG7 dataset chosen for experiments. The objects were etched into a chromium film on top of a fused silica substrate.}
    \label{fig:S3}
\end{figure}

For the experimental demonstration we chose a subset of the MPEG7 dataset \cite{yang2012affinity} to classify. 5 classes were chosen, and 5 objects per class were used, resulting in a total of 25 objects. 15 were used in the training set, and 10 were used in the testing set. The objects were etched into a chromium film on top of a fused silica substrate. Figure \ref{fig:S3}a shows the fabrication process and Figure \ref{fig:S3}b shows the fabricated objects. The ones encircled by blue were the test set, and the ones in red were the train set. Each object was about \SI{2.5}{mm}$\times$\SI{2.5}{mm}.

\subsection{MPEG7 classification experimental setups}\label{mpeg7_experiment_setup}
\begin{figure}[H]
    \centering
    \includegraphics[width = \textwidth]{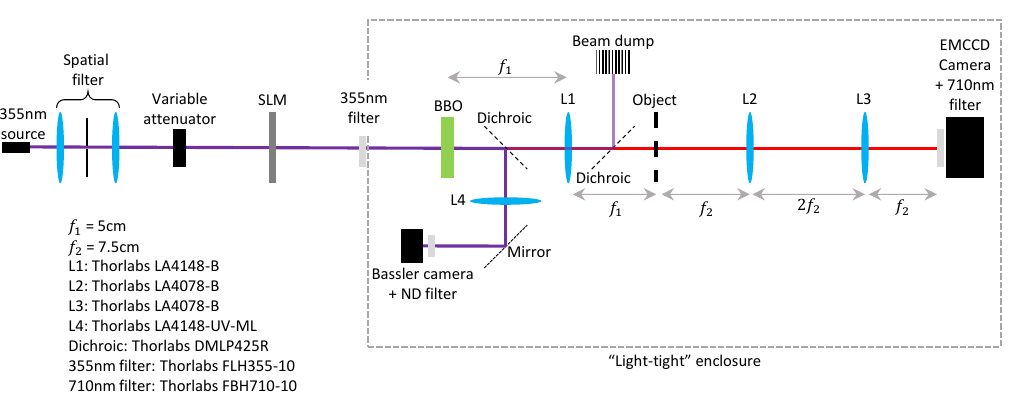}
    \caption{\textbf{Schematic for correlated-photon computer vision.} A $4-f$ system images the objects onto an EMCCD camera. The objects are placed at the fourier plane of the $\beta$-BBO crystal.}
    \label{fig:S11}
\end{figure}

\begin{figure}[H]
    \centering
    \includegraphics[width = \textwidth]{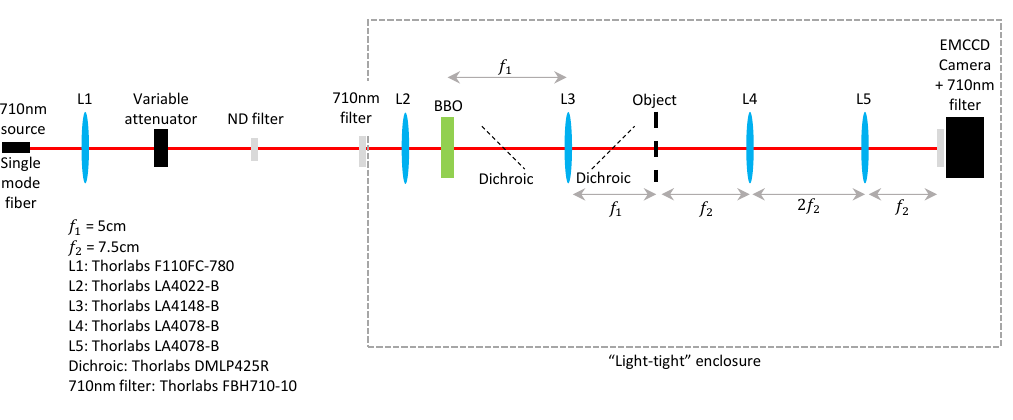}
    \caption{\textbf{Schematic for uncorrelated-photon computer vision.} A $4-f$ system images the objects onto an EMCCD camera. The illumination is a laser diode source connected to an single-mode fiber whose spot size is adjusted by a telescope.}
    \label{fig:S12}
\end{figure}

As mentioned earlier, both the correlated and uncorrelated illumination sources used were at \SI{710}{nm}. Figure \ref{fig:S11} shows a schematic of the setup used to classify the MPEG7 objects using correlated light. An \SI{355}{nm} laser source was first sent through a spatial filter, followed by a variable attenuator (a half-wave plate + a polarizing beam splitter) before interactins with an SLM. The SLM shaped the angular spectrum of the pump beam (\SI{355}{nm} source) and then the beam entered a ``light-tight'' enclosure through a 355nm spectral filter. The pump beam produced SPDC photons at \SI{710}{nm} in a $\beta$-BBO crystal (the distance between the SLM and the $\beta$-BBO crystal was about $\sim$\SI{20}{cm}). The fabricated MPEG7 objects were placed at the fourier plane of the $\beta$-BBO, i.e., at the focal plane of Lens L1. A $4-f$ system imaged the objects onto the EMCCD camera. The mount holding the mounted objects was controlled by two linear actuators (Newport CONEX-TRA25CC) for automated data collection.

Figure \ref{fig:S12} shows a schematic of the setup used to classify the MPEG7 objects using uncorrelated light. An \SI{710}{nm} single-mode fiber coupled laser source was collimated using a lens (L1) into free space, followed by a variable attenuator (a half-wave plate + a polarizing beam splitter) before being further attenuated by a series of ND filters. The beam entered a ``light-tight'' enclosure through a 710nm spectral filter. The beam's size was adjusted using a telescope (lenses L2 and L3) before illuminating the objects. A $4-f$ system imaged the objects onto the EMCCD camera. 

\subsection{EMCCD images of the MPEG7 objects}\label{mpeg7_EMCCD_calibration}
\begin{figure}[H]
    \centering
    \includegraphics[width = \textwidth]{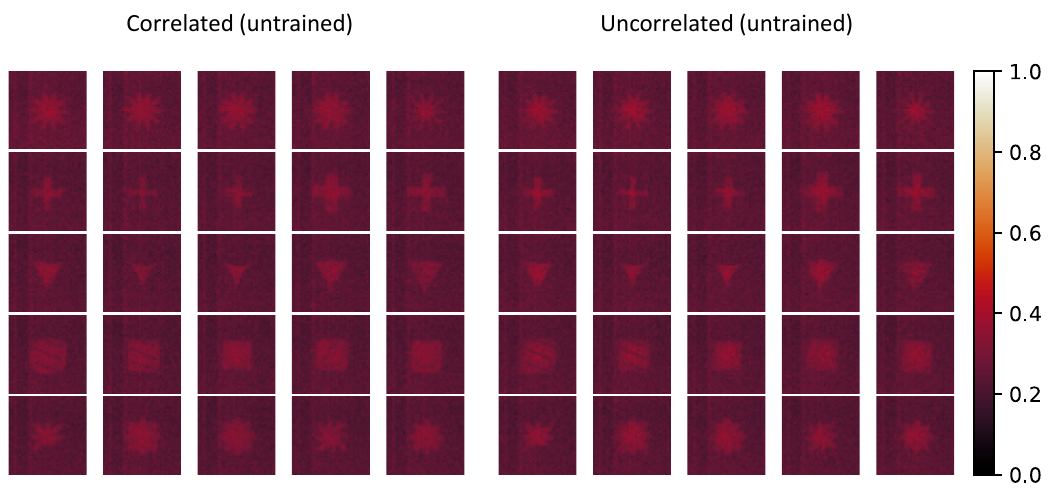}
    \caption{\textbf{EMCCD images of the MPEG7 objects.} Images of the 25 MPEG7 objects for both the untrained correlated and uncorrelated illuminations for the first corresponding data point in figure \ref{fig:spdc_training}.}
    \label{fig:S13}
\end{figure}

\begin{figure}[H]
    \centering
    \includegraphics[width = \textwidth]{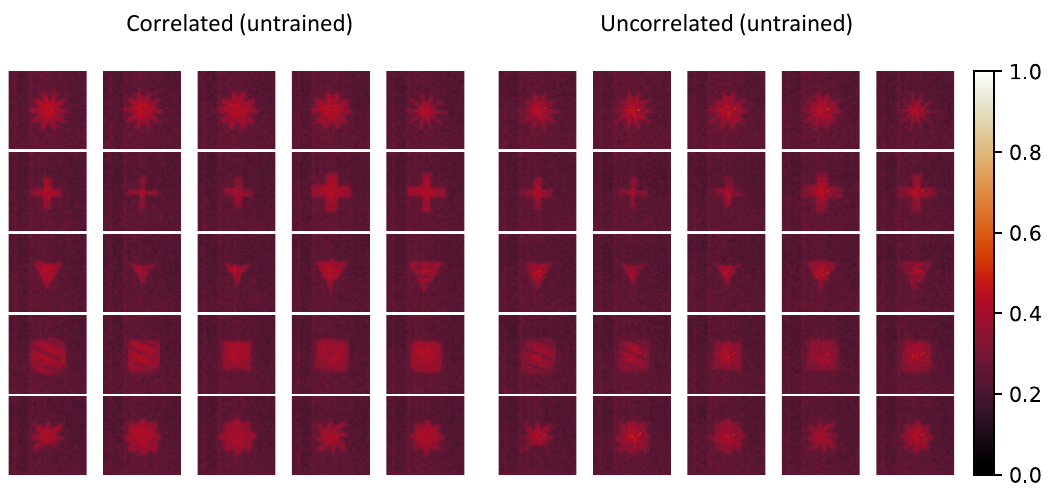}
    \caption{\textbf{EMCCD images of the MPEG7 objects.} Images of the 25 MPEG7 objects for both the untrained correlated and uncorrelated illuminations for the second corresponding data point in figure \ref{fig:spdc_training}.}
    \label{fig:S14}
\end{figure}

\begin{figure}[H]
    \centering
    \includegraphics[width = \textwidth]{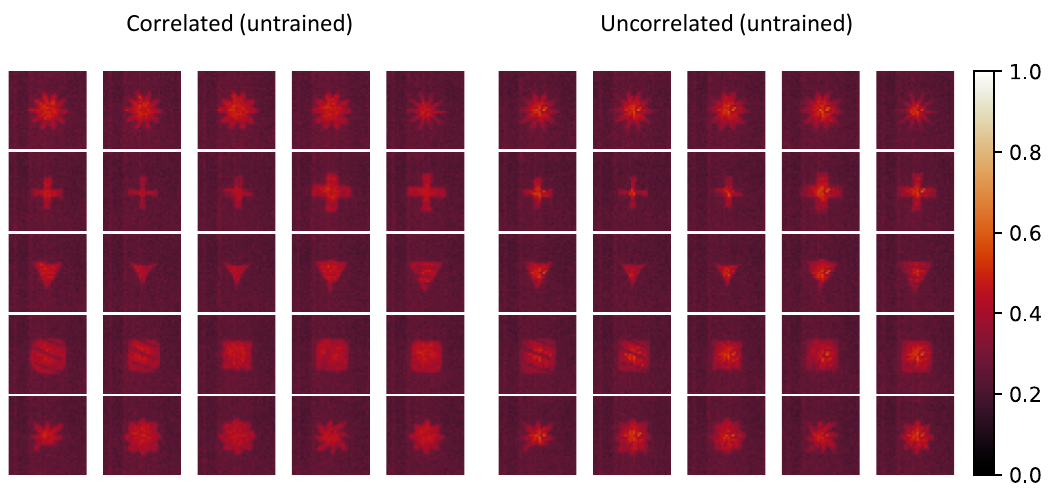}
    \caption{\textbf{EMCCD images of the MPEG7 objects.} Images of the 25 MPEG7 objects for both the untrained correlated and uncorrelated illuminations for the third corresponding data point in figure \ref{fig:spdc_training}.}
    \label{fig:S15}
\end{figure}

\begin{figure}[H]
    \centering
    \includegraphics[width = \textwidth]{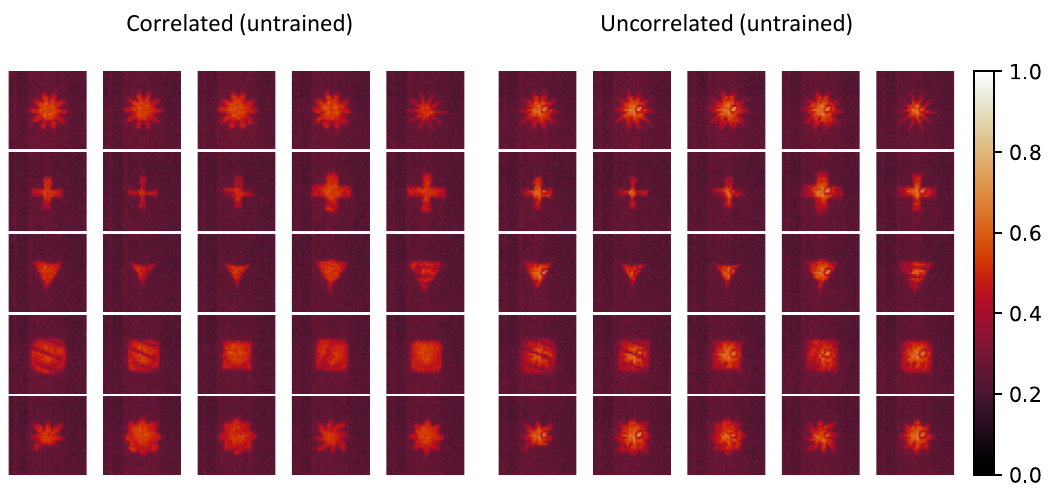}
    \caption{\textbf{EMCCD images of the MPEG7 objects.} Images of the 25 MPEG7 objects for both the untrained correlated and uncorrelated illuminations for the fourth corresponding data point in figure \ref{fig:spdc_training}.}
    \label{fig:S16}
\end{figure}

\begin{figure}[H]
    \centering
    \includegraphics[width = \textwidth]{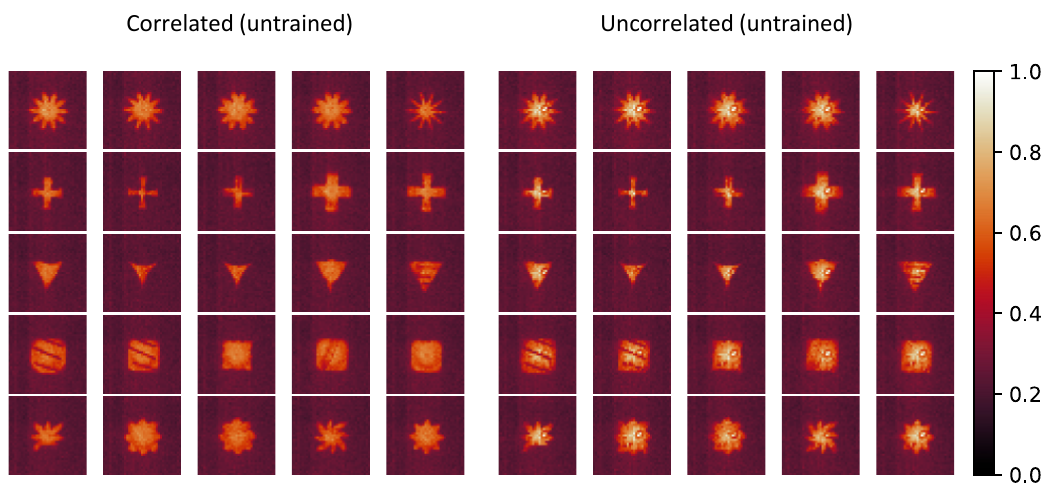}
    \caption{\textbf{EMCCD images of the MPEG7 objects.} Images of the 25 MPEG7 objects for both the untrained correlated and uncorrelated illuminations for the fifth corresponding data point in figure \ref{fig:spdc_training}.}
    \label{fig:S17}
\end{figure}

\begin{figure}[H]
    \centering
    \includegraphics[width = \textwidth]{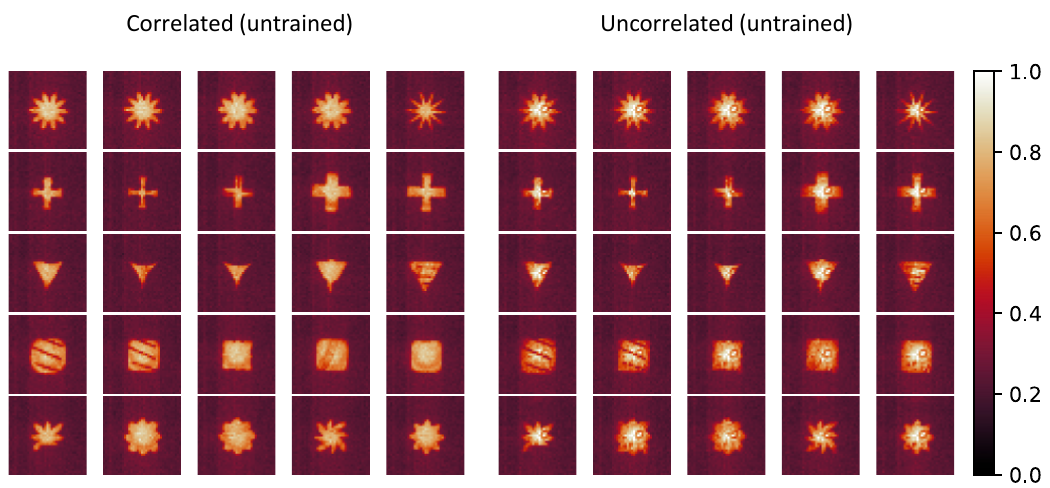}
    \caption{\textbf{EMCCD images of the MPEG7 objects.} Images of the 25 MPEG7 objects for both the untrained correlated and uncorrelated illuminations for the sixth corresponding data point in figure \ref{fig:spdc_training}.}
    \label{fig:S18}
\end{figure}

\begin{figure}[H]
    \centering
    \includegraphics[width = \textwidth]{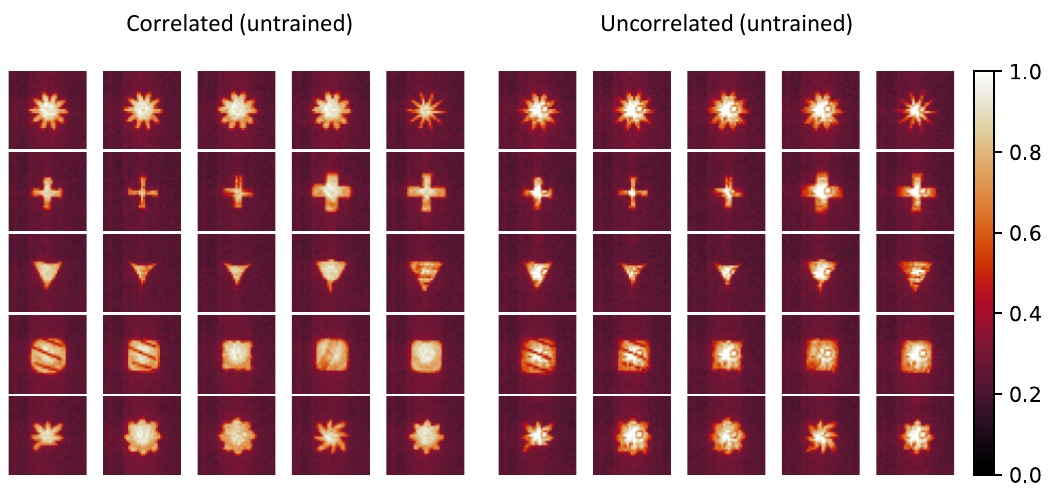}
    \caption{\textbf{EMCCD images of the MPEG7 objects.} Images of the 25 MPEG7 objects for both the untrained correlated and uncorrelated illuminations for the seventh corresponding data point in figure \ref{fig:spdc_training}.}
    \label{fig:S19}
\end{figure}

\begin{figure}[H]
    \centering
    \includegraphics[width = \textwidth]{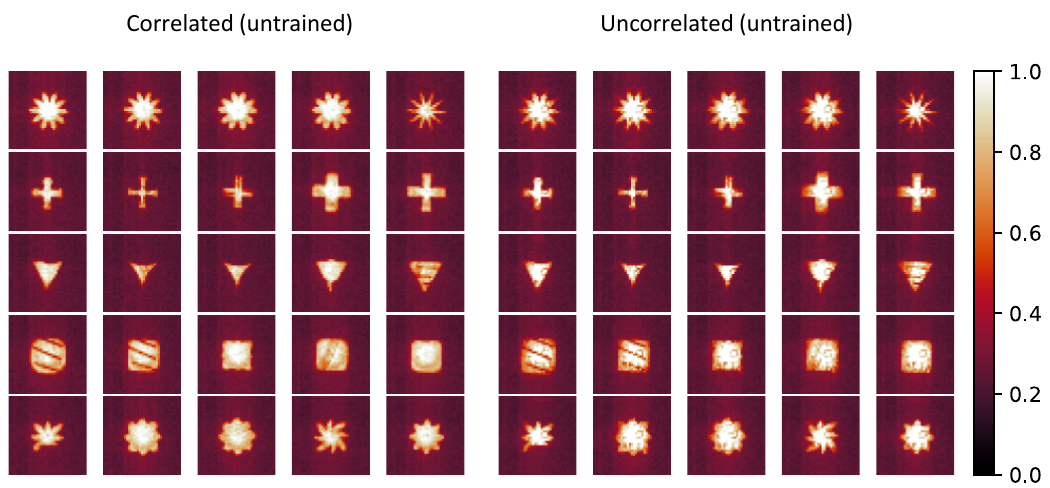}
    \caption{\textbf{EMCCD images of the MPEG7 objects.} Images of the 25 MPEG7 objects for both the untrained correlated and uncorrelated illuminations for the eighth corresponding data point in figure \ref{fig:spdc_training}.}
    \label{fig:S20}
\end{figure}

Figures \ref{fig:S13} through \ref{fig:S20} show the 25 MPEG7 objects for both the 8 untrained correlated and uncorrelated illumination points in figure \ref{fig:spdc_training} (the figures are in increasing order of illumination power). Each image shows the average of 2000 frames each taken with a \SI{1}{ms} exposure time. Each frame was converted to estimated photon counts as described in Appendix \ref{camera_frame_postprocessing}. Each shot had a size of 37 pixels $\times$ 37 pixels (8$\times$8 binning was used during camera readout; the original region of interest size on the camera was 296 $\times$ 296). These images are purely for demonstrative purposes, all inferences with the set Transformer were performed with the processed frames without any averaging.

\begin{figure}[H]
    \centering
    \includegraphics[width = 0.75\textwidth]{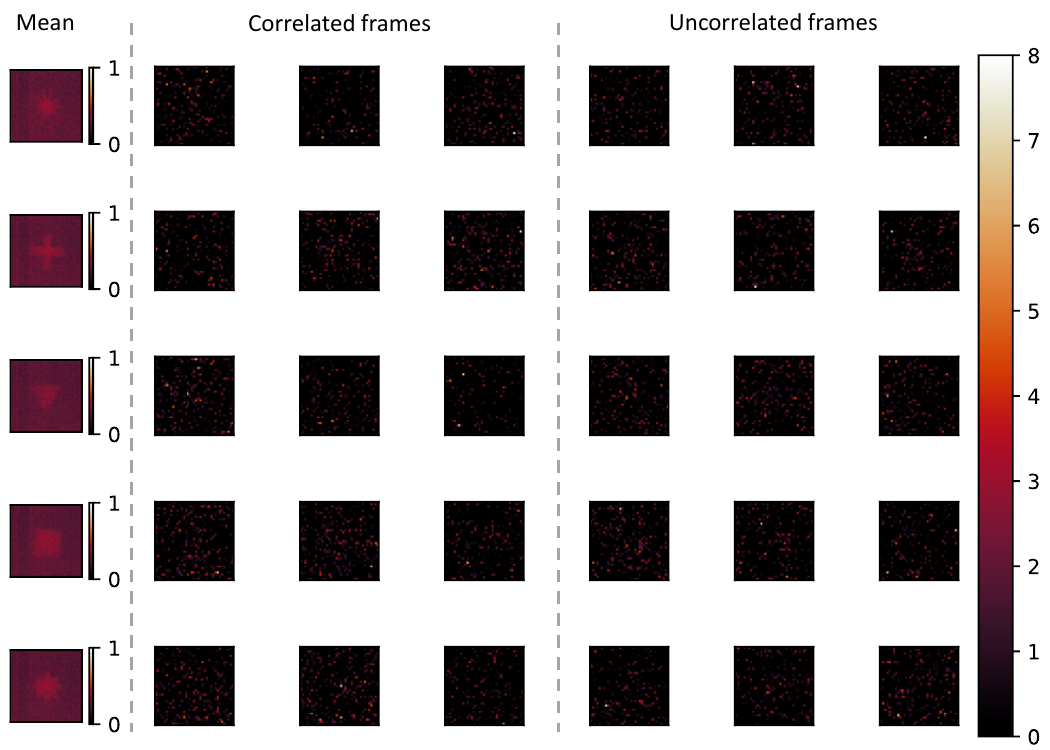}
    \caption{\textbf{Examples of single frames of the MPEG7 objects.} Examples of single frames from each class that were fed as part of the input to the set Transformer. The data is from the first corresponding point for the untrained correlated and uncorrelated illumination points in figure \ref{fig:spdc_training}.}
    \label{fig:S21}
\end{figure}

\begin{figure}[H]
    \centering
    \includegraphics[width = 0.75\textwidth]{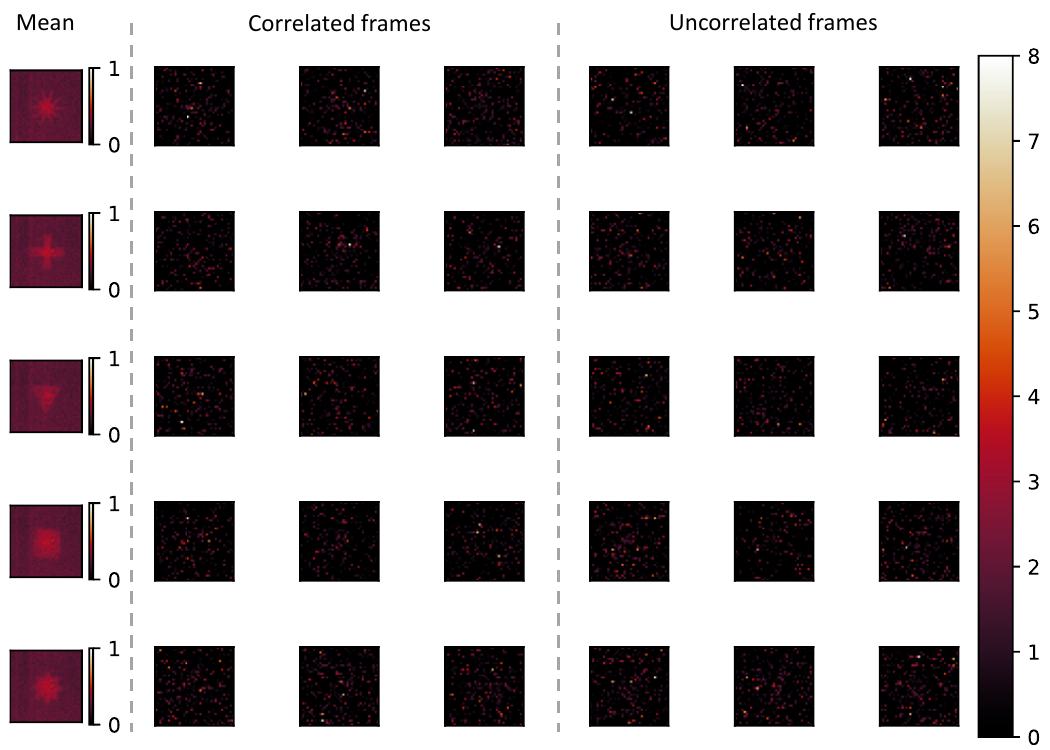}
    \caption{\textbf{Examples of single frames of the MPEG7 objects.} Examples of single frames from each class that were fed as part of the input to the set Transformer. The data is from the second corresponding point for the untrained correlated and uncorrelated illumination points in figure \ref{fig:spdc_training}.}
    \label{fig:S22}
\end{figure}

\begin{figure}[H]
    \centering
    \includegraphics[width = 0.75\textwidth]{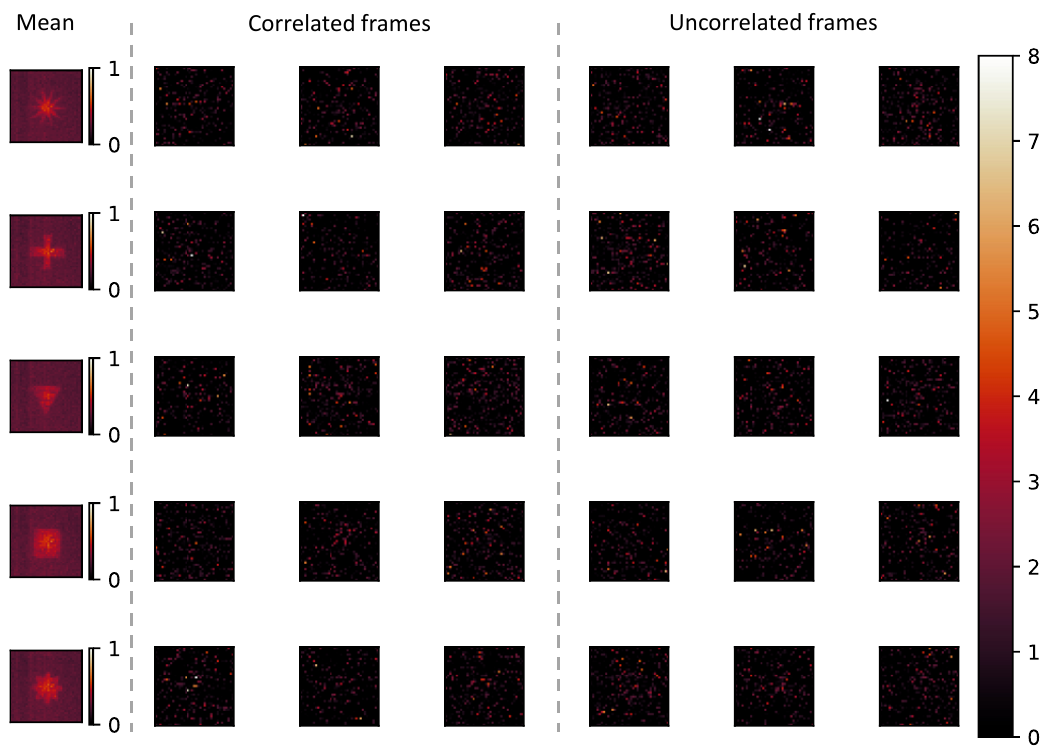}
    \caption{\textbf{Examples of single frames of the MPEG7 objects.} Examples of single frames from each class that were fed as part of the input to the set Transformer. The data is from the third corresponding point for the untrained correlated and uncorrelated illumination points in figure \ref{fig:spdc_training}.}
    \label{fig:S23}
\end{figure}

\begin{figure}[H]
    \centering
    \includegraphics[width = 0.75\textwidth]{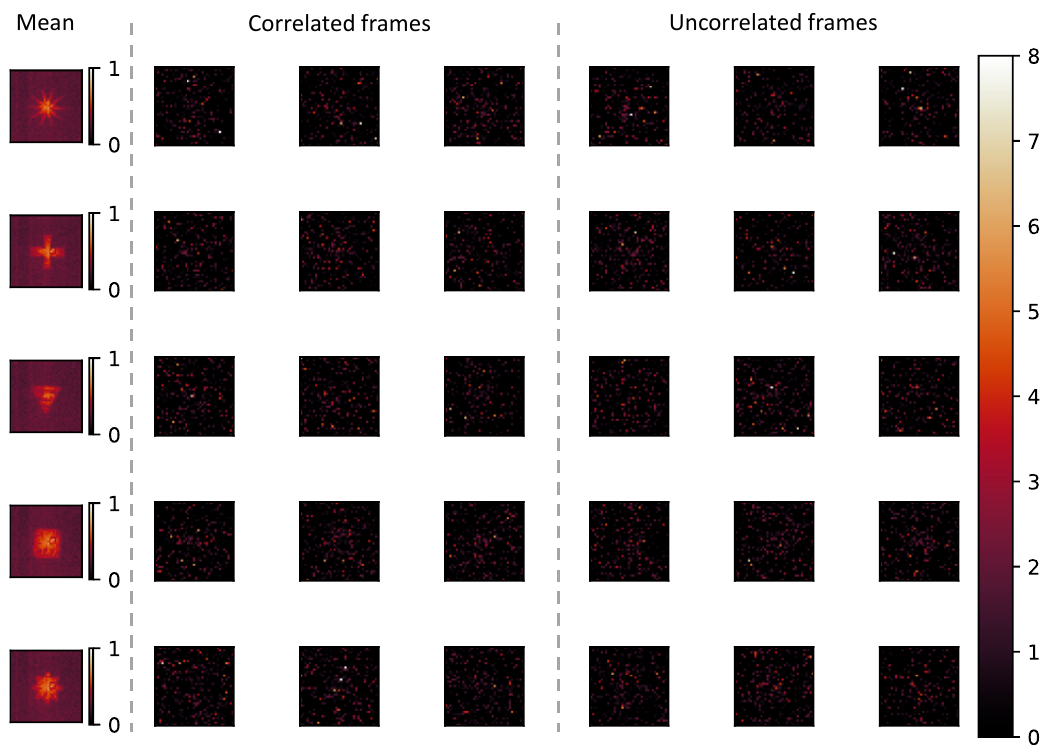}
    \caption{\textbf{Examples of single frames of the MPEG7 objects.} Examples of single frames from each class that were fed as part of the input to the set Transformer. The data is from the fourth corresponding point for the untrained correlated and uncorrelated illumination points in figure \ref{fig:spdc_training}.}
    \label{fig:S24}
\end{figure}

\begin{figure}[H]
    \centering
    \includegraphics[width = 0.75\textwidth]{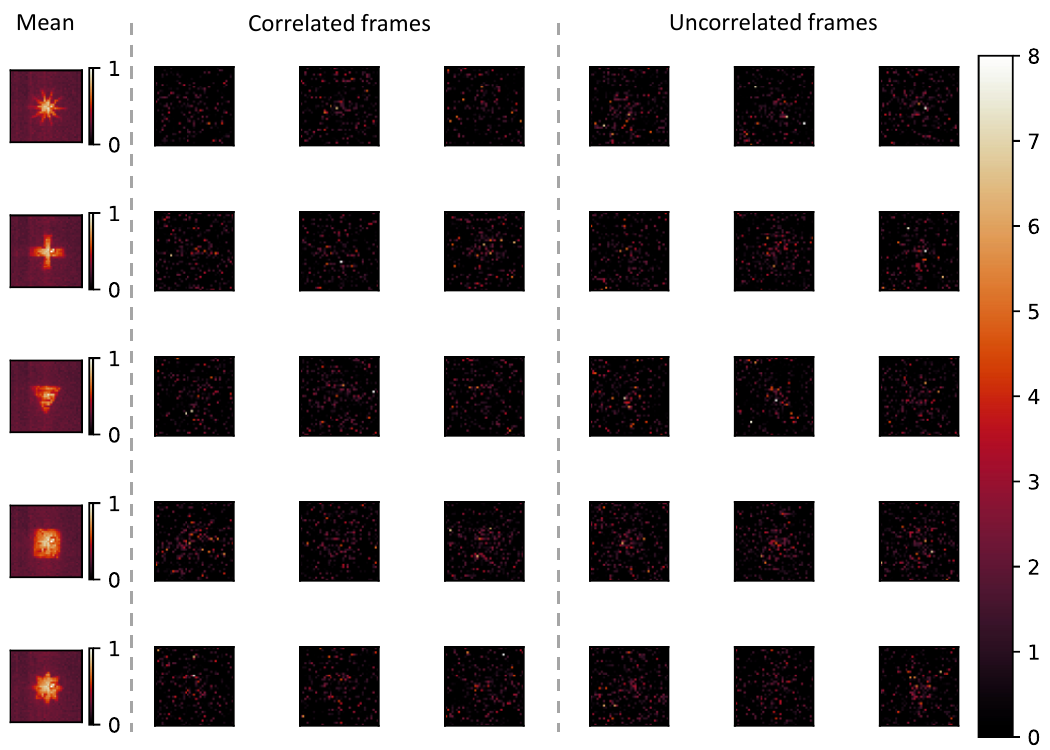}
    \caption{\textbf{Examples of single frames of the MPEG7 objects.} Examples of single frames from each class that were fed as part of the input to the set Transformer. The data is from the fifth corresponding point for the untrained correlated and uncorrelated illumination points in figure \ref{fig:spdc_training}.}
    \label{fig:S25}
\end{figure}

\begin{figure}[H]
    \centering
    \includegraphics[width = 0.75\textwidth]{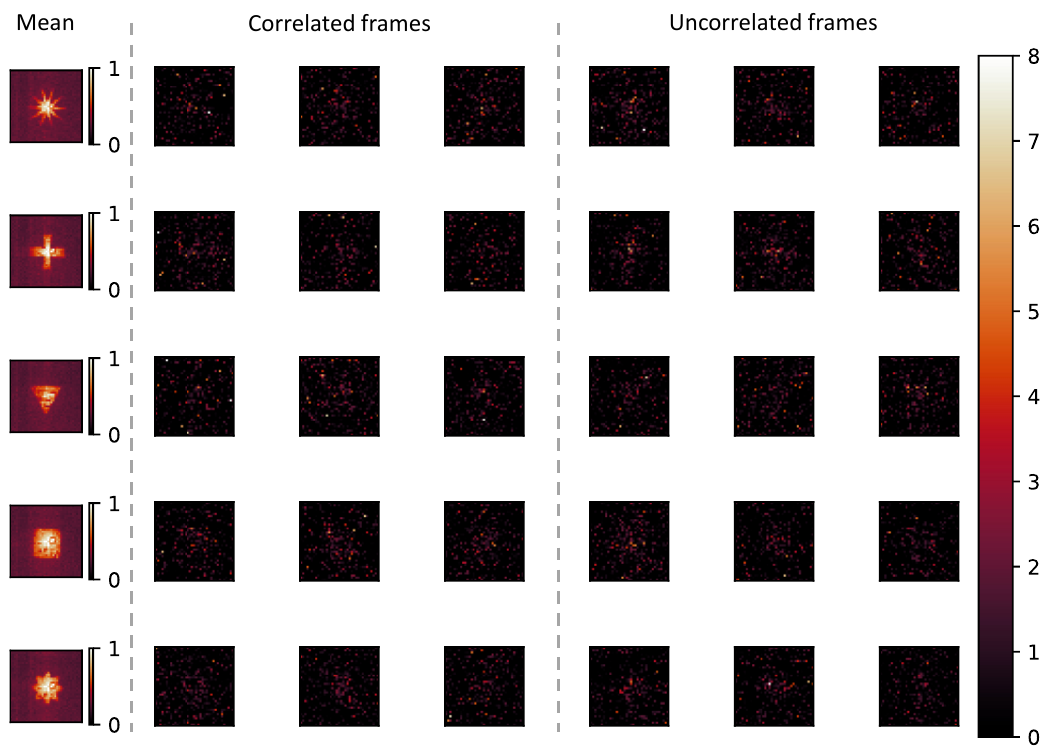}
    \caption{\textbf{Examples of single frames of the MPEG7 objects.} Examples of single frames from each class that were fed as part of the input to the set Transformer. The data is from the sixth corresponding point for the untrained correlated and uncorrelated illumination points in figure \ref{fig:spdc_training}.}
    \label{fig:S26}
\end{figure}

\begin{figure}[H]
    \centering
    \includegraphics[width = 0.75\textwidth]{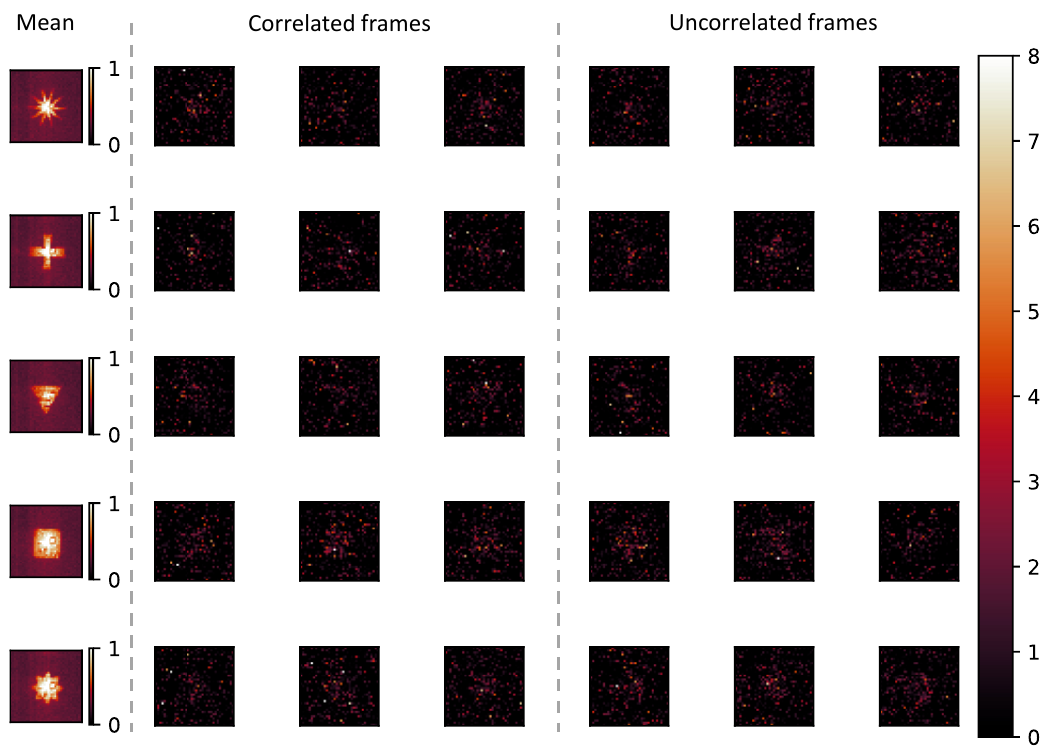}
    \caption{\textbf{Examples of single frames of the MPEG7 objects.} Examples of single frames from each class that were fed as part of the input to the set Transformer. The data is from the seventh corresponding point for the untrained correlated and uncorrelated illumination points in figure \ref{fig:spdc_training}.}
    \label{fig:S27}
\end{figure}

\begin{figure}[H]
    \centering
    \includegraphics[width = 0.75\textwidth]{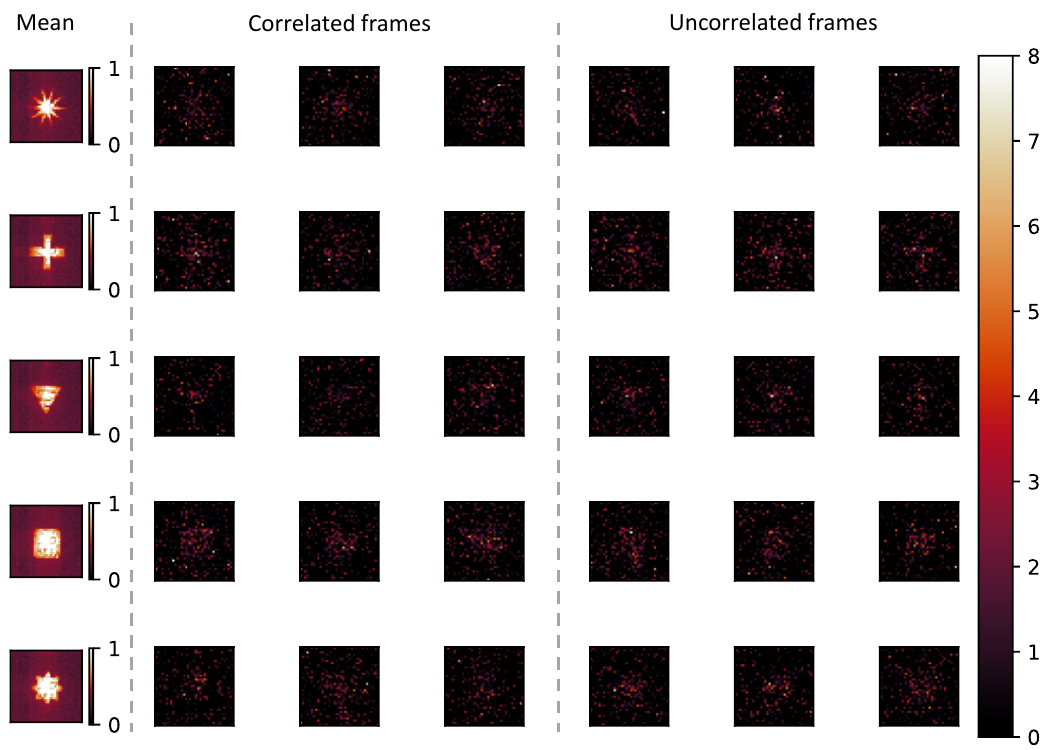}
    \caption{\textbf{Examples of single frames of the MPEG7 objects.} Examples of single frames from each class that were fed as part of the input to the set Transformer. The data is from the eighth corresponding point for the untrained correlated and uncorrelated illumination points in figure \ref{fig:spdc_training}.}
    \label{fig:S28}
\end{figure}

Figures \ref{fig:S21} through \ref{fig:S28} show examples of single frames each class that were fed as part of the input to the set Transformer. The figures are in increasing order of illumination power, again for both the 8 untrained correlated and uncorrelated illumination points in figure \ref{fig:spdc_training}. All images shown have been averaged over 2000 shots.

\subsection{MPEG7 classification results}\label{mpeg7_experiment_results_extended}
\begin{figure}[H]
    \centering
    \includegraphics[width = 0.8\textwidth]{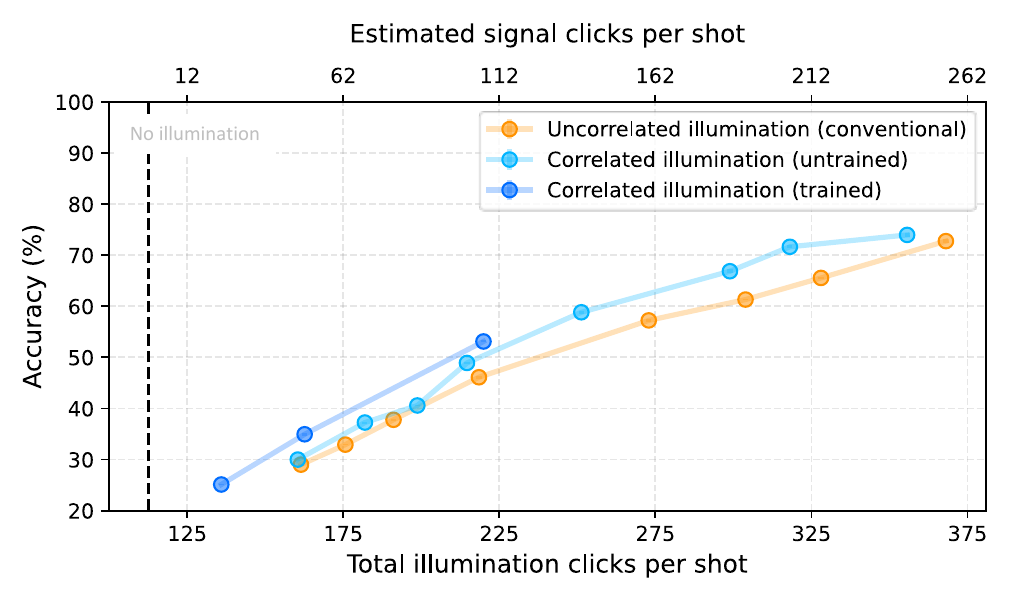}
    \caption{\textbf{Classification accuracy curve for the MPEG7 experiment.} Classification accuracy as a function of illumination power when 2 shots were used for inference}
    \label{fig:S31}
\end{figure}

\begin{figure}[H]
    \centering
    \includegraphics[width = 0.8\textwidth]{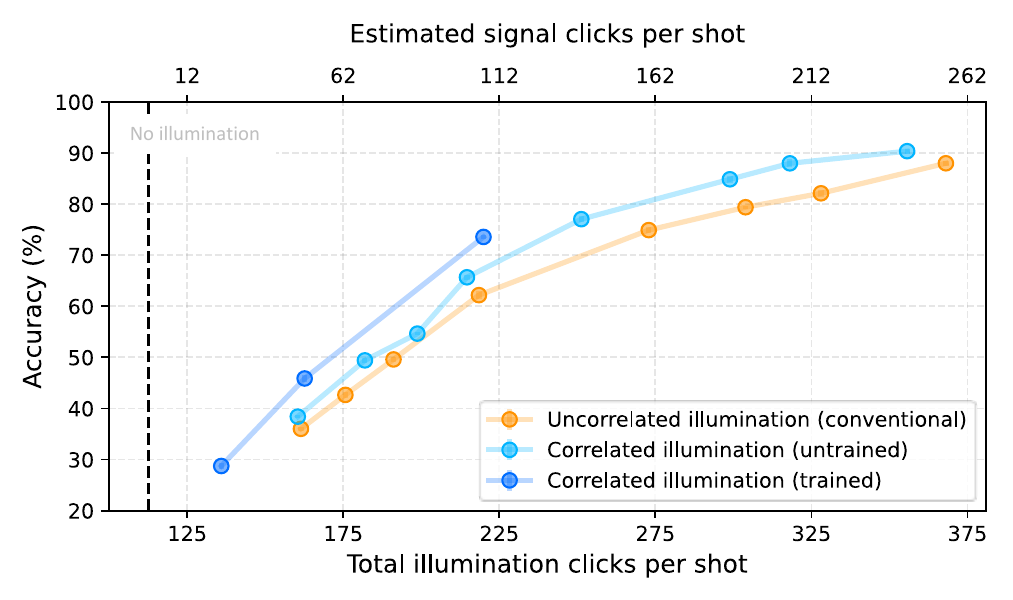}
    \caption{\textbf{Classification accuracy curve for the MPEG7 experiment.} Classification accuracy as a function of illumination power when 6 shots were used for inference}
    \label{fig:S32}
\end{figure}

\begin{figure}[H]
    \centering
    \includegraphics[width = 0.8\textwidth]{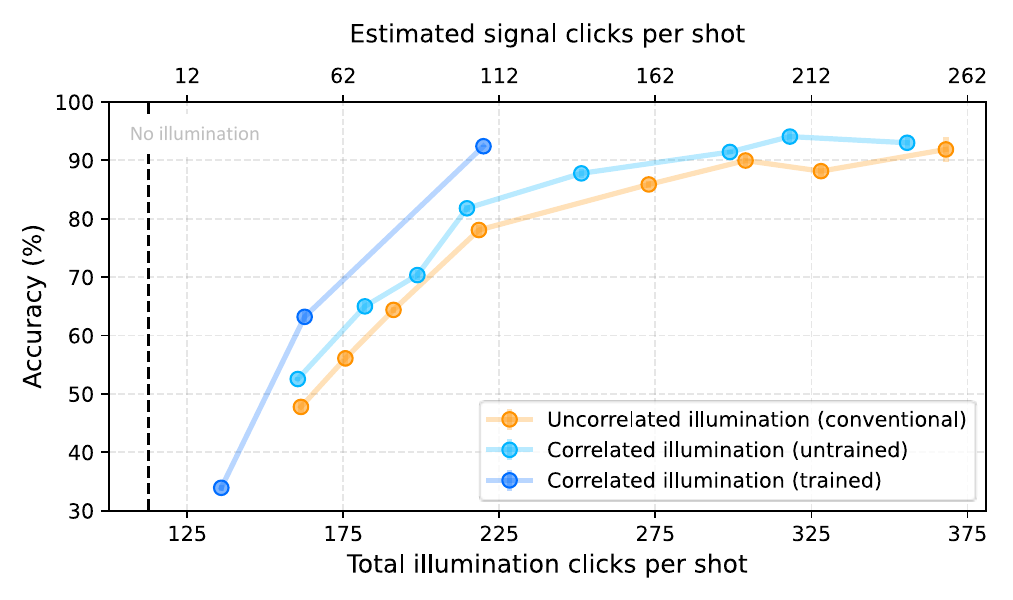}
    \caption{\textbf{Classification accuracy curve for the MPEG7 experiment.} Classification accuracy as a function of illumination power when 20 shots were used for inference}
    \label{fig:S33}
\end{figure}

\begin{figure}[H]
    \centering
    \includegraphics[width = 0.8\textwidth]{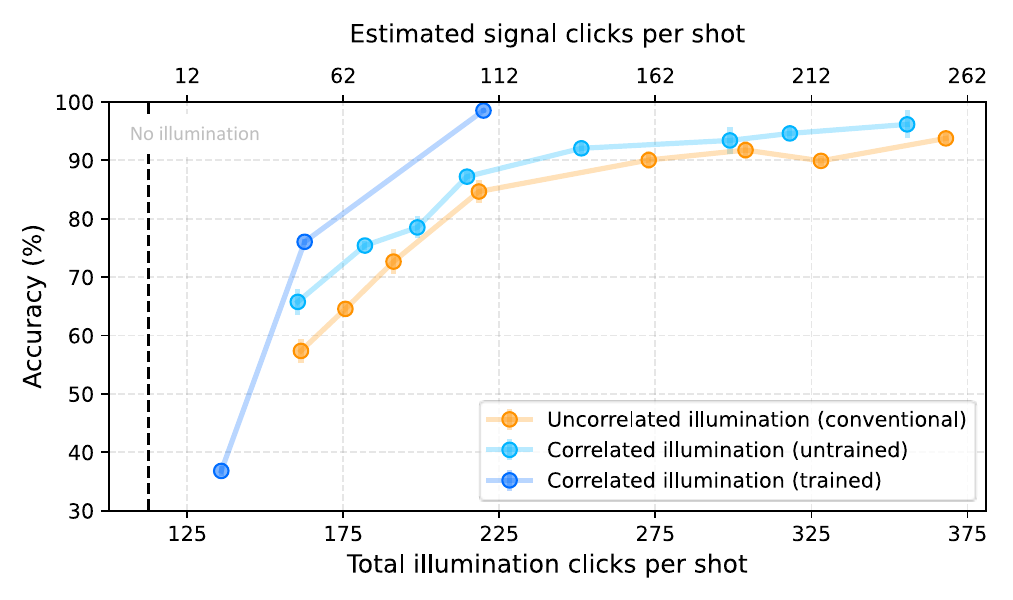}
    \caption{\textbf{Classification accuracy curve for the MPEG7 experiment.} Classification accuracy as a function of illumination power when 50 shots were used for inference}
    \label{fig:S34}
\end{figure}

\begin{figure}[H]
    \centering
    \includegraphics[width = 0.8\textwidth]{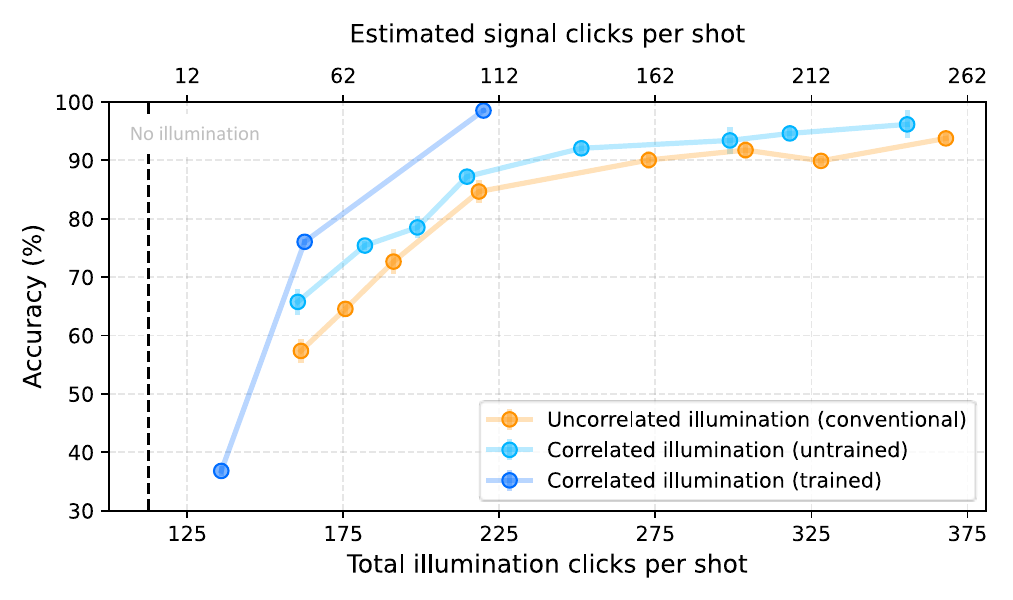}
    \caption{\textbf{Classification accuracy curve for the MPEG7 experiment.} Classification accuracy as a function of illumination power when 100 shots were used for inference}
    \label{fig:S35}
\end{figure}

As a complement to Figure \ref{fig:spdc_training}, Figures \ref{fig:S31} - \ref{fig:S35} show the classification accuracy on the test MPEG7 set as a function of the illumination power for a various number of shots, spedifically 2, 6, 20, 50, and 100. When classifying with 2 shots (see Figure \ref{fig:S31}), we see that there is no gap between the uncorrelated and untrained correlated at the lower illumination levels becuase it's hard for the transformer to extract correlation information. However as the illumination power increases, i.e., number of events per shots increases, we see a small gap emerge indicating that the transformer is able to extract more information. On the flip side, when using 100 shots for classification, we find that there is no gap between the uncorrelated and untrained correlated at the higher illumination levels because the mean-field now carries enough information to distinguish between the classes. At lower illumination levels there still is a gap, suggesting that the transformer is benefiting from the correlated illumination.

\begin{figure}[H]
    \centering
    \includegraphics[width = 0.8\textwidth]{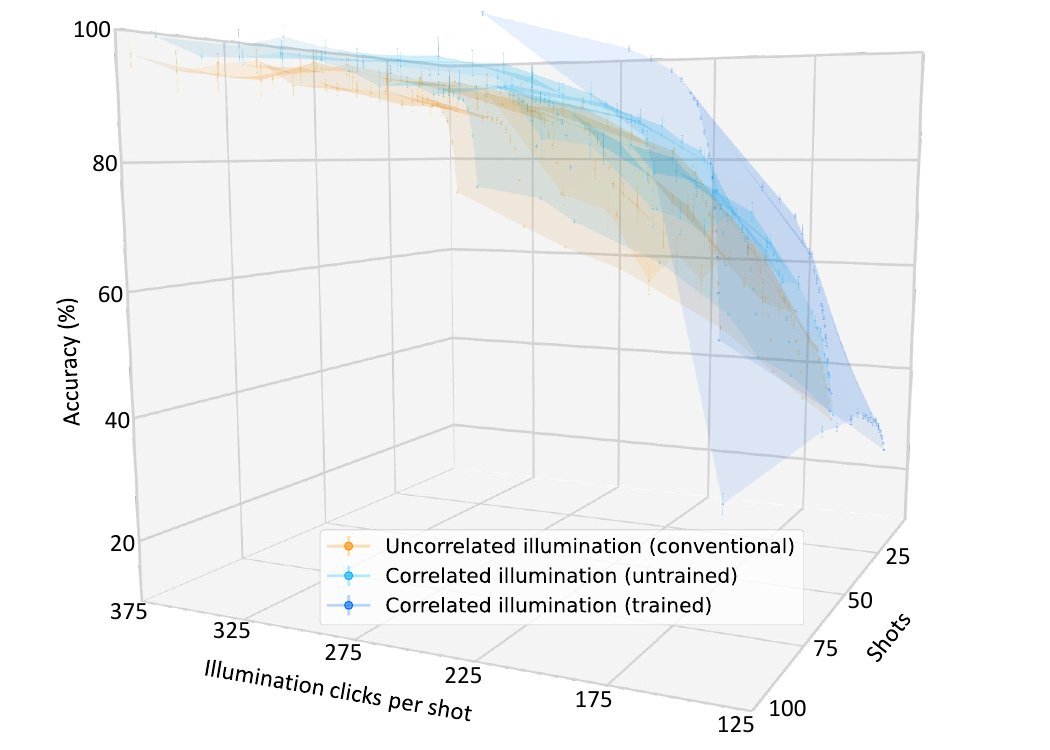}
    \caption{\textbf{Classification accuracy curves for the MPEG7 experiment.} Extended data from Figure \ref{fig:spdc_training}.}
    \label{fig:S30}
\end{figure}

Figure \ref{fig:S30} shows the classification accuracy curves (experiment) as a function of the illumination power and the number of shots used for inference. We find that the largest gaps for this particular task are in the high-illumination and low-shot or high-shot and low-illumination limits. 

\begin{figure}[H]
    \centering
    \includegraphics[width = 0.8\textwidth]{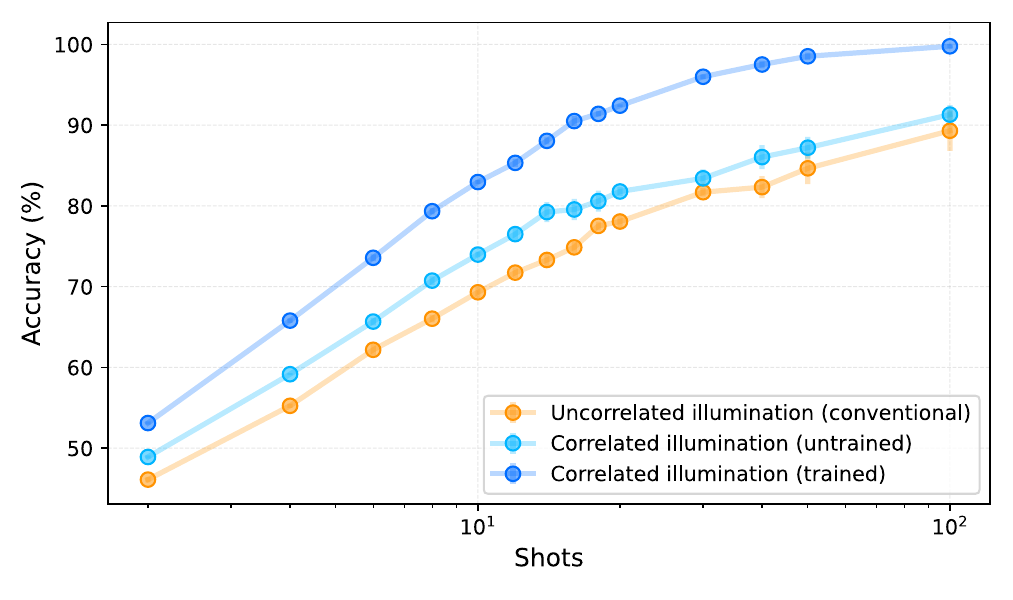}
    \caption{\textbf{Classification accuracy curves for the MPEG7 experiment.} Classification accuracy as a function of the number of shots used for inference when $\sim 100$ photons were used for illumination. Extended data from Figure \ref{fig:spdc_training}.}
    \label{fig:S38}
\end{figure}

\begin{figure}[H]
    \centering
    \includegraphics[width = 0.8\textwidth]{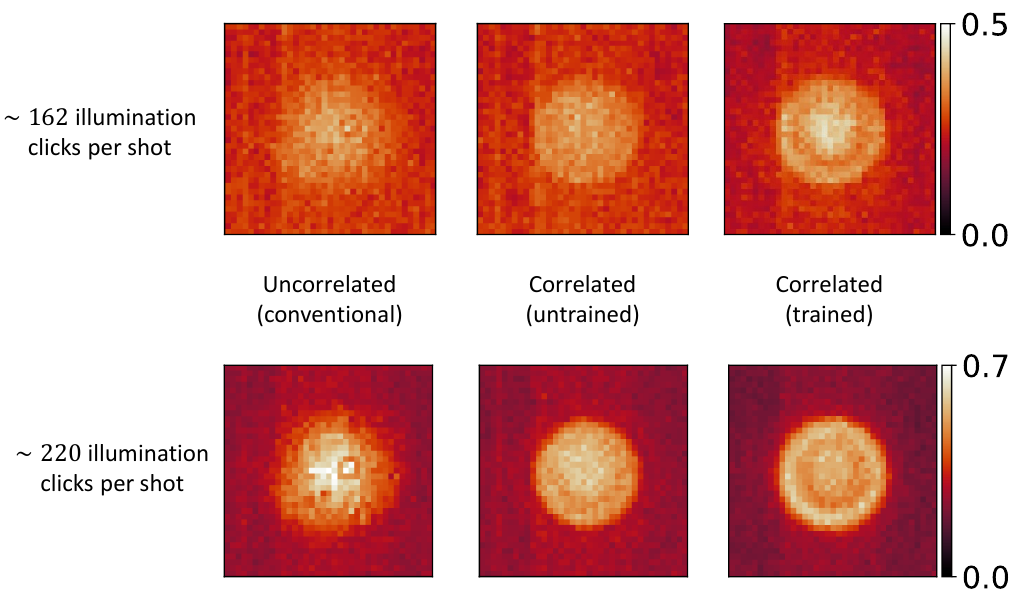}
    \caption{\textbf{Mean-field illumination patterns for the MPEG7 experiment.} Trained illumination for $\sim162$ and $\sim220$ average clicks per shot from Figure \ref{fig:spdc_training}.}
    \label{fig:S36}
\end{figure}
Appendix Figure \ref{fig:S38} shows the classification accuracy of as a function of the number of shots used for inference when $\sim 100$ photons were used for illumination. We find that the trained correlated illumination requires $>5\times$ fewer shots (16 vs 100) as compared to the conventional imaging pipeline in order to reach an accuracy of 90\%.

Appendix Figure \ref{fig:S36} shows the mean-field illumination patterns for all the three light sources at two different powers ($\sim162$ and $\sim220$ average clicks per shot). We see that the optimized illuminations differ from the unoptimized ones and even change at different power levels. At $\sim 162$ illumination clicks per frame, the training procedure increases the brightness in the center as opposed to the periphery, whereas for $\sim 220$ illumination photons per frame, the optimal pattern is brighter in the periphery.

\subsection{Simulations of cell organelle classification}\label{cell_training}
\begin{figure}[H]
    \centering
    \includegraphics[width = 0.8\textwidth]{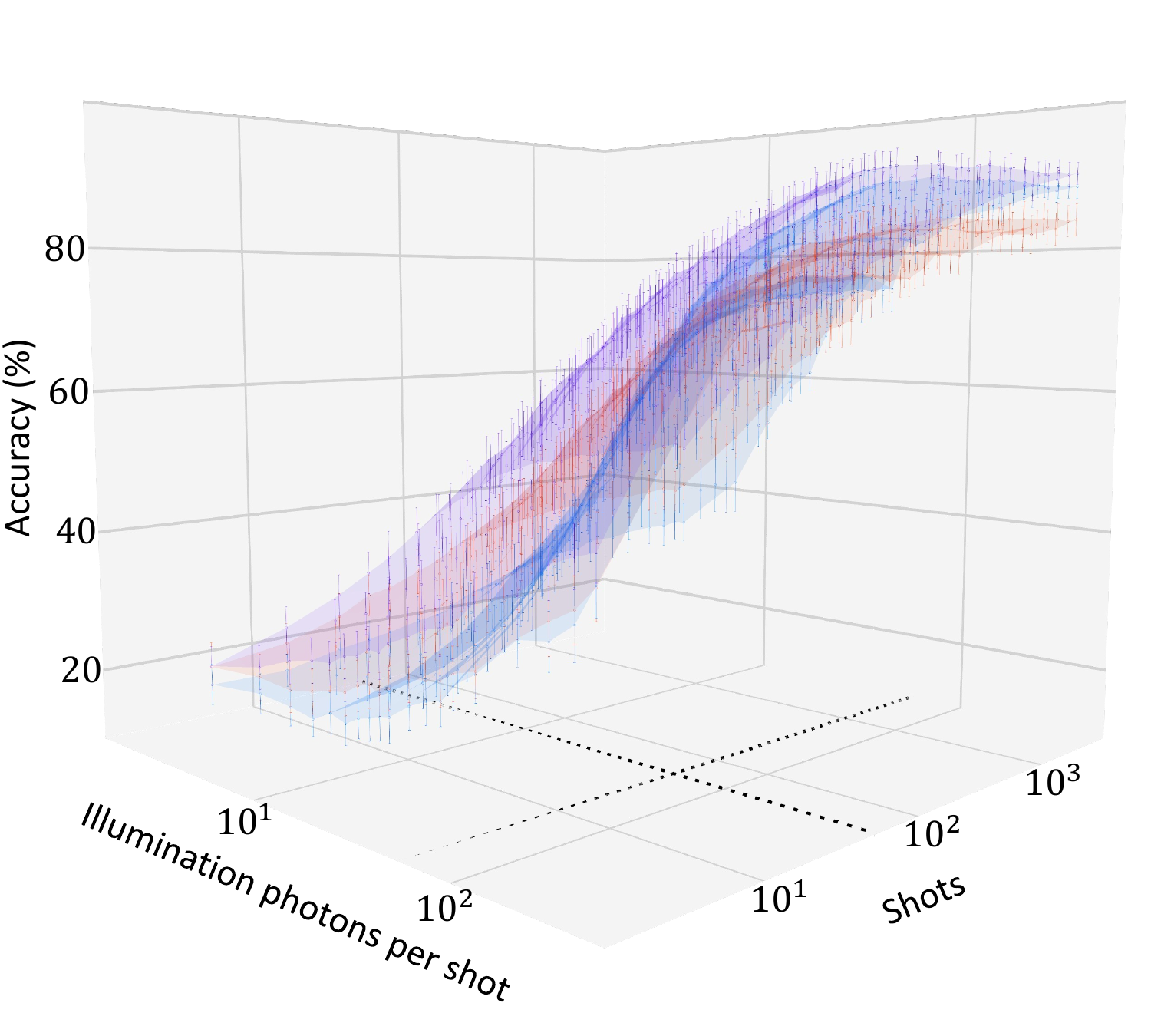}
    \caption{\textbf{Simulations of cell organelle classification with different illuminations.} Classification accuracy surfaces for the illuminations as a function of the number of illumination photons per shot and the number of shots used for classification. The dotted lines depict the cross sections of these curves that were shown in Figure \ref{fig:cell_classification}.}
    \label{fig:S39}
\end{figure}

For the simulations involving the trained correlated and uncorrelated light sources, we chose a dataset of cell organelles generated in a flow-cytometry device \cite{schraivogel2022high}. There were 5 classes of cell organelles: Mitochondria, nucleoli, nucleus, cytoplasm and cell membrane. For the correlated light, we considered two kinds of trained correlated (SPDC) light: (1) a simulation of the experiment with its digital model, and (2) an `ideal' correlated bi-photon source that has an arbitrary green's function, which, in principle is experimentally feasible by engineering the phase-matching function. The trained uncorrelated light source was modeled as a coherent Gaussian beam phase-modulated by an SLM, with the cell organelles placed in the Fourier plane of the modulator. We compared both to a conventional computer vision baseline where no training was performed for the illumination. A set-Transformer was used as the backend for each case. Figure \ref{fig:S39} shows the classification accuracy curves of the `ideal' correlated bi-photon source, the digital model of our experimental correlated photon source, and the trainable uncorrelated source.

The simulations were performed with a background noise level that was taken from the EMCCD camera used in experiment. A probability mass function was fit to photon counts of experimentally acquired background frames from the EMCCD and this was used to sample noise counts in the simulations. No photon loss was considered. All of the code for the simulations and the dataset can be found at

\subsection{The effect of loss on correlations}\label{loss_effect}
To see how photon loss affects the classification accuracy we ran simulations on the cell dataset with different values of loss for the ideal correlated biphoton source as described in \ref{corr_vs_uncorr}. 60 illumination photons and 100 shots were used for the simulations. We see that photon loss reduces the advantage of correlations. Around a transmission level of 60\% we find that the performance drops to that of the trained coherent illumination.

\begin{figure}[H]
    \centering
    \includegraphics[width = 0.75\textwidth]{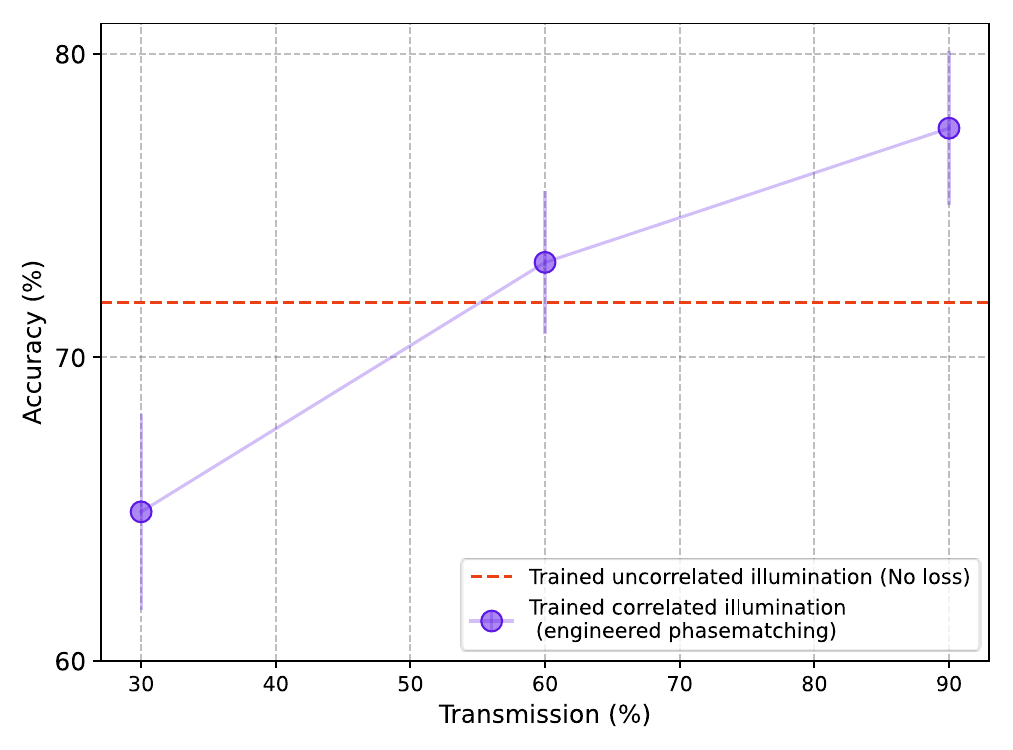}
    \caption{\textbf{Simulation: Effect of loss on correlated illumination.}}
    \label{fig:S42}
\end{figure}

\section{Ultra-low-light computer vision pipelines}\label{Comparison}

\begin{figure}[H]
    \centering
    \includegraphics[width = \textwidth]{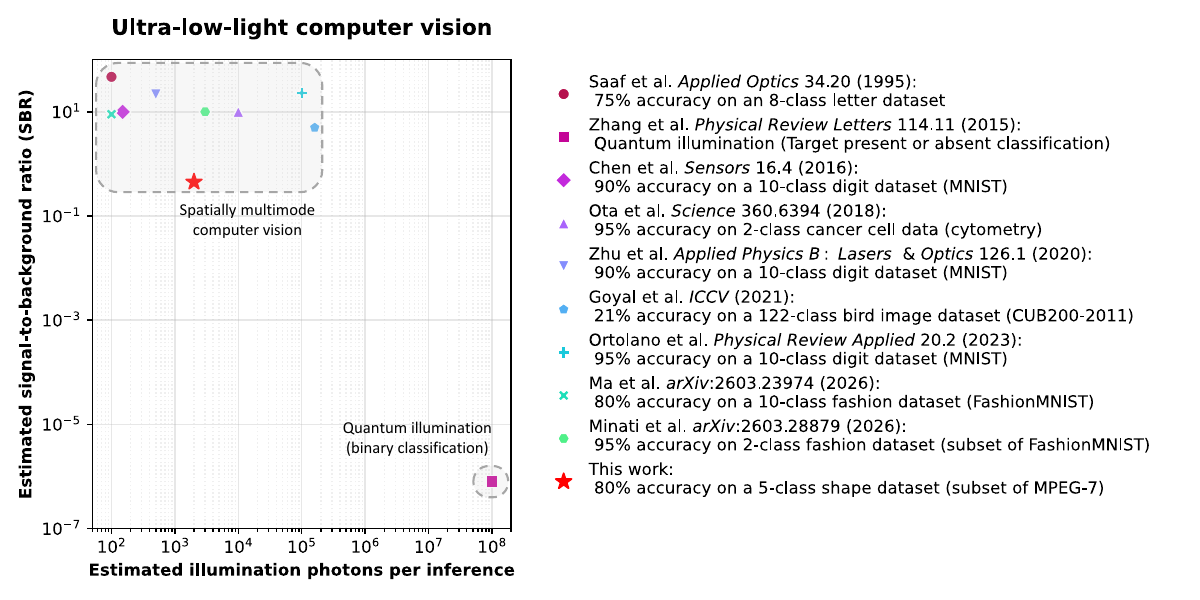}
    \caption{\textbf{Comparison between different experimental demonstrations of object/image classification in ultra-low-light conditions.}}
    \label{fig:comparison_QI}
\end{figure}

The past several decades have seen the development of computer vision and image sensing systems that have pushed towards photon-starved and noise-dominated regimes \cite{wernick1986image, ota2018ghost, goyal2021photon, ortolano2023quantum, minati2026quantum}. Early demonstrations of photon-limited pattern recognition established foundational benchmarks in ideal dark-room conditions, relying on high SBRs to extract features \cite{saaf1995photon, zhu2020photon}. As sensor technologies matured, modern inference frameworks successfully reduced the requisite photon budget down to tens or hundreds of photons per inference \cite{chen2016vision, ma2026machine}. However, pushing performance into environments where the noise floor is larger the signal---regimes where the SBR drops below unity---requires different approaches to information encoding. To contextualize our contribution, Appendix~Figure~\ref{fig:comparison_QI} presents a comparison of low-light computer vision and pattern recognition literature, mapping the total illumination photon budget per inference against the operating SBR. While foundational quantum illumination experiments have operated at very low SBRs (often below $10^{-6}$), these demonstrations are fundamentally about single-mode binary classification tasks (i.e., detecting a target's presence or absence) and require a large photon budget to outperform uncorrelated-photon techniques \cite{zhang2015entanglement}. We attempt to bridge this gap, by translating the noise resilience of engineered photon correlations into the domain of spatially multimode computer vision at ultra-low photon budgets. By training spatial biphoton correlations, our framework isolates target features from the noise floor, maintaining high-accuracy inference with less than a 100 shots even at an SBR < 1.

\begin{thebibliography}{79}%
	\makeatletter
	\providecommand \@ifxundefined [1]{%
		\@ifx{#1\undefined}
	}%
	\providecommand \@ifnum [1]{%
		\ifnum #1\expandafter \@firstoftwo
		\else \expandafter \@secondoftwo
		\fi
	}%
	\providecommand \@ifx [1]{%
		\ifx #1\expandafter \@firstoftwo
		\else \expandafter \@secondoftwo
		\fi
	}%
	\providecommand \natexlab [1]{#1}%
	\providecommand \enquote  [1]{``#1''}%
	\providecommand \bibnamefont  [1]{#1}%
	\providecommand \bibfnamefont [1]{#1}%
	\providecommand \citenamefont [1]{#1}%
	\providecommand \href@noop [0]{\@secondoftwo}%
	\providecommand \href [0]{\begingroup \@sanitize@url \@href}%
	\providecommand \@href[1]{\@@startlink{#1}\@@href}%
	\providecommand \@@href[1]{\endgroup#1\@@endlink}%
	\providecommand \@sanitize@url [0]{\catcode `\\12\catcode `\$12\catcode `\&12\catcode `\#12\catcode `\^12\catcode `\_12\catcode `\%12\relax}%
	\providecommand \@@startlink[1]{}%
	\providecommand \@@endlink[0]{}%
	\providecommand \url  [0]{\begingroup\@sanitize@url \@url }%
	\providecommand \@url [1]{\endgroup\@href {#1}{\urlprefix }}%
	\providecommand \urlprefix  [0]{URL }%
	\providecommand \Eprint [0]{\href }%
	\providecommand \doibase [0]{https://doi.org/}%
	\providecommand \selectlanguage [0]{\@gobble}%
	\providecommand \bibinfo  [0]{\@secondoftwo}%
	\providecommand \bibfield  [0]{\@secondoftwo}%
	\providecommand \translation [1]{[#1]}%
	\providecommand \BibitemOpen [0]{}%
	\providecommand \bibitemStop [0]{}%
	\providecommand \bibitemNoStop [0]{.\EOS\space}%
	\providecommand \EOS [0]{\spacefactor3000\relax}%
	\providecommand \BibitemShut  [1]{\csname bibitem#1\endcsname}%
	\let\auto@bib@innerbib\@empty
	%</preamble>
	\bibitem [{\citenamefont {Wetzstein}\ \emph {et~al.}(2020)\citenamefont {Wetzstein}, \citenamefont {Ozcan}, \citenamefont {Gigan}, \citenamefont {Fan}, \citenamefont {Englund}, \citenamefont {Solja{\v{c}}i{\'c}}, \citenamefont {Denz}, \citenamefont {Miller},\ and\ \citenamefont {Psaltis}}]{wetzstein2020inference}%
	\BibitemOpen
	\bibfield  {author} {\bibinfo {author} {\bibfnamefont {G.}~\bibnamefont {Wetzstein}}, \bibinfo {author} {\bibfnamefont {A.}~\bibnamefont {Ozcan}}, \bibinfo {author} {\bibfnamefont {S.}~\bibnamefont {Gigan}}, \bibinfo {author} {\bibfnamefont {S.}~\bibnamefont {Fan}}, \bibinfo {author} {\bibfnamefont {D.}~\bibnamefont {Englund}}, \bibinfo {author} {\bibfnamefont {M.}~\bibnamefont {Solja{\v{c}}i{\'c}}}, \bibinfo {author} {\bibfnamefont {C.}~\bibnamefont {Denz}}, \bibinfo {author} {\bibfnamefont {D.~A.}\ \bibnamefont {Miller}},\ and\ \bibinfo {author} {\bibfnamefont {D.}~\bibnamefont {Psaltis}},\ }\bibfield  {title} {\bibinfo {title} {Inference in artificial intelligence with deep optics and photonics},\ }\href@noop {} {\bibfield  {journal} {\bibinfo  {journal} {Nature}\ }\textbf {\bibinfo {volume} {588}},\ \bibinfo {pages} {39} (\bibinfo {year} {2020})}\BibitemShut {NoStop}%
	\bibitem [{\citenamefont {Pirandola}\ \emph {et~al.}(2018)\citenamefont {Pirandola}, \citenamefont {Bardhan}, \citenamefont {Gehring}, \citenamefont {Weedbrook},\ and\ \citenamefont {Lloyd}}]{pirandola2018advances}%
	\BibitemOpen
	\bibfield  {author} {\bibinfo {author} {\bibfnamefont {S.}~\bibnamefont {Pirandola}}, \bibinfo {author} {\bibfnamefont {B.~R.}\ \bibnamefont {Bardhan}}, \bibinfo {author} {\bibfnamefont {T.}~\bibnamefont {Gehring}}, \bibinfo {author} {\bibfnamefont {C.}~\bibnamefont {Weedbrook}},\ and\ \bibinfo {author} {\bibfnamefont {S.}~\bibnamefont {Lloyd}},\ }\bibfield  {title} {\bibinfo {title} {Advances in photonic quantum sensing},\ }\href@noop {} {\bibfield  {journal} {\bibinfo  {journal} {Nature Photonics}\ }\textbf {\bibinfo {volume} {12}},\ \bibinfo {pages} {724} (\bibinfo {year} {2018})}\BibitemShut {NoStop}%
	\bibitem [{\citenamefont {Magana-Loaiza}\ and\ \citenamefont {Boyd}(2019)}]{magana2019quantum}%
	\BibitemOpen
	\bibfield  {author} {\bibinfo {author} {\bibfnamefont {O.~S.}\ \bibnamefont {Magana-Loaiza}}\ and\ \bibinfo {author} {\bibfnamefont {R.~W.}\ \bibnamefont {Boyd}},\ }\bibfield  {title} {\bibinfo {title} {Quantum imaging and information},\ }\href@noop {} {\bibfield  {journal} {\bibinfo  {journal} {Reports on Progress in Physics}\ }\textbf {\bibinfo {volume} {82}},\ \bibinfo {pages} {124401} (\bibinfo {year} {2019})}\BibitemShut {NoStop}%
	\bibitem [{\citenamefont {Moreau}\ \emph {et~al.}(2019)\citenamefont {Moreau}, \citenamefont {Toninelli}, \citenamefont {Gregory},\ and\ \citenamefont {Padgett}}]{moreau2019imaging}%
	\BibitemOpen
	\bibfield  {author} {\bibinfo {author} {\bibfnamefont {P.-A.}\ \bibnamefont {Moreau}}, \bibinfo {author} {\bibfnamefont {E.}~\bibnamefont {Toninelli}}, \bibinfo {author} {\bibfnamefont {T.}~\bibnamefont {Gregory}},\ and\ \bibinfo {author} {\bibfnamefont {M.~J.}\ \bibnamefont {Padgett}},\ }\bibfield  {title} {\bibinfo {title} {Imaging with quantum states of light},\ }\href@noop {} {\bibfield  {journal} {\bibinfo  {journal} {Nature Reviews Physics}\ }\textbf {\bibinfo {volume} {1}},\ \bibinfo {pages} {367} (\bibinfo {year} {2019})}\BibitemShut {NoStop}%
	\bibitem [{\citenamefont {Defienne}\ \emph {et~al.}(2024)\citenamefont {Defienne}, \citenamefont {Bowen}, \citenamefont {Chekhova}, \citenamefont {Lemos}, \citenamefont {Oron}, \citenamefont {Ramelow}, \citenamefont {Treps},\ and\ \citenamefont {Faccio}}]{defienne2024advances}%
	\BibitemOpen
	\bibfield  {author} {\bibinfo {author} {\bibfnamefont {H.}~\bibnamefont {Defienne}}, \bibinfo {author} {\bibfnamefont {W.~P.}\ \bibnamefont {Bowen}}, \bibinfo {author} {\bibfnamefont {M.}~\bibnamefont {Chekhova}}, \bibinfo {author} {\bibfnamefont {G.~B.}\ \bibnamefont {Lemos}}, \bibinfo {author} {\bibfnamefont {D.}~\bibnamefont {Oron}}, \bibinfo {author} {\bibfnamefont {S.}~\bibnamefont {Ramelow}}, \bibinfo {author} {\bibfnamefont {N.}~\bibnamefont {Treps}},\ and\ \bibinfo {author} {\bibfnamefont {D.}~\bibnamefont {Faccio}},\ }\bibfield  {title} {\bibinfo {title} {Advances in quantum imaging},\ }\href@noop {} {\bibfield  {journal} {\bibinfo  {journal} {Nature Photonics}\ }\textbf {\bibinfo {volume} {18}},\ \bibinfo {pages} {1024} (\bibinfo {year} {2024})}\BibitemShut {NoStop}%
	\bibitem [{\citenamefont {Tsao}\ \emph {et~al.}(2025)\citenamefont {Tsao}, \citenamefont {Ling}, \citenamefont {Hinkle}, \citenamefont {Chen}, \citenamefont {Jha}, \citenamefont {Yan},\ and\ \citenamefont {Utzat}}]{tsao2025enhancing}%
	\BibitemOpen
	\bibfield  {author} {\bibinfo {author} {\bibfnamefont {C.}~\bibnamefont {Tsao}}, \bibinfo {author} {\bibfnamefont {H.}~\bibnamefont {Ling}}, \bibinfo {author} {\bibfnamefont {A.}~\bibnamefont {Hinkle}}, \bibinfo {author} {\bibfnamefont {Y.}~\bibnamefont {Chen}}, \bibinfo {author} {\bibfnamefont {K.~K.}\ \bibnamefont {Jha}}, \bibinfo {author} {\bibfnamefont {Z.-L.}\ \bibnamefont {Yan}},\ and\ \bibinfo {author} {\bibfnamefont {H.}~\bibnamefont {Utzat}},\ }\bibfield  {title} {\bibinfo {title} {Enhancing spectroscopy and microscopy with emerging methods in photon correlation and quantum illumination},\ }\href@noop {} {\bibfield  {journal} {\bibinfo  {journal} {Nature Nanotechnology}\ }\textbf {\bibinfo {volume} {20}},\ \bibinfo {pages} {1001} (\bibinfo {year} {2025})}\BibitemShut {NoStop}%
	\bibitem [{\citenamefont {Casacio}\ \emph {et~al.}(2021)\citenamefont {Casacio}, \citenamefont {Madsen}, \citenamefont {Terrasson}, \citenamefont {Waleed}, \citenamefont {Barnscheidt}, \citenamefont {Hage}, \citenamefont {Taylor},\ and\ \citenamefont {Bowen}}]{casacio2021quantum}%
	\BibitemOpen
	\bibfield  {author} {\bibinfo {author} {\bibfnamefont {C.~A.}\ \bibnamefont {Casacio}}, \bibinfo {author} {\bibfnamefont {L.~S.}\ \bibnamefont {Madsen}}, \bibinfo {author} {\bibfnamefont {A.}~\bibnamefont {Terrasson}}, \bibinfo {author} {\bibfnamefont {M.}~\bibnamefont {Waleed}}, \bibinfo {author} {\bibfnamefont {K.}~\bibnamefont {Barnscheidt}}, \bibinfo {author} {\bibfnamefont {B.}~\bibnamefont {Hage}}, \bibinfo {author} {\bibfnamefont {M.~A.}\ \bibnamefont {Taylor}},\ and\ \bibinfo {author} {\bibfnamefont {W.~P.}\ \bibnamefont {Bowen}},\ }\bibfield  {title} {\bibinfo {title} {Quantum-enhanced nonlinear microscopy},\ }\href@noop {} {\bibfield  {journal} {\bibinfo  {journal} {Nature}\ }\textbf {\bibinfo {volume} {594}},\ \bibinfo {pages} {201} (\bibinfo {year} {2021})}\BibitemShut {NoStop}%
	\bibitem [{\citenamefont {Nagata}\ \emph {et~al.}(2007)\citenamefont {Nagata}, \citenamefont {Okamoto}, \citenamefont {O'brien}, \citenamefont {Sasaki},\ and\ \citenamefont {Takeuchi}}]{nagata2007beating}%
	\BibitemOpen
	\bibfield  {author} {\bibinfo {author} {\bibfnamefont {T.}~\bibnamefont {Nagata}}, \bibinfo {author} {\bibfnamefont {R.}~\bibnamefont {Okamoto}}, \bibinfo {author} {\bibfnamefont {J.~L.}\ \bibnamefont {O'brien}}, \bibinfo {author} {\bibfnamefont {K.}~\bibnamefont {Sasaki}},\ and\ \bibinfo {author} {\bibfnamefont {S.}~\bibnamefont {Takeuchi}},\ }\bibfield  {title} {\bibinfo {title} {Beating the standard quantum limit with four-entangled photons},\ }\href@noop {} {\bibfield  {journal} {\bibinfo  {journal} {Science}\ }\textbf {\bibinfo {volume} {316}},\ \bibinfo {pages} {726} (\bibinfo {year} {2007})}\BibitemShut {NoStop}%
	\bibitem [{\citenamefont {Israel}\ \emph {et~al.}(2014)\citenamefont {Israel}, \citenamefont {Rosen},\ and\ \citenamefont {Silberberg}}]{israel2014supersensitive}%
	\BibitemOpen
	\bibfield  {author} {\bibinfo {author} {\bibfnamefont {Y.}~\bibnamefont {Israel}}, \bibinfo {author} {\bibfnamefont {S.}~\bibnamefont {Rosen}},\ and\ \bibinfo {author} {\bibfnamefont {Y.}~\bibnamefont {Silberberg}},\ }\bibfield  {title} {\bibinfo {title} {Supersensitive polarization microscopy using noon states of light},\ }\href@noop {} {\bibfield  {journal} {\bibinfo  {journal} {Physical review letters}\ }\textbf {\bibinfo {volume} {112}},\ \bibinfo {pages} {103604} (\bibinfo {year} {2014})}\BibitemShut {NoStop}%
	\bibitem [{\citenamefont {Slussarenko}\ \emph {et~al.}(2017)\citenamefont {Slussarenko}, \citenamefont {Weston}, \citenamefont {Chrzanowski}, \citenamefont {Shalm}, \citenamefont {Verma}, \citenamefont {Nam},\ and\ \citenamefont {Pryde}}]{slussarenko2017unconditional}%
	\BibitemOpen
	\bibfield  {author} {\bibinfo {author} {\bibfnamefont {S.}~\bibnamefont {Slussarenko}}, \bibinfo {author} {\bibfnamefont {M.~M.}\ \bibnamefont {Weston}}, \bibinfo {author} {\bibfnamefont {H.~M.}\ \bibnamefont {Chrzanowski}}, \bibinfo {author} {\bibfnamefont {L.~K.}\ \bibnamefont {Shalm}}, \bibinfo {author} {\bibfnamefont {V.~B.}\ \bibnamefont {Verma}}, \bibinfo {author} {\bibfnamefont {S.~W.}\ \bibnamefont {Nam}},\ and\ \bibinfo {author} {\bibfnamefont {G.~J.}\ \bibnamefont {Pryde}},\ }\bibfield  {title} {\bibinfo {title} {Unconditional violation of the shot-noise limit in photonic quantum metrology},\ }\href@noop {} {\bibfield  {journal} {\bibinfo  {journal} {Nature Photonics}\ }\textbf {\bibinfo {volume} {11}},\ \bibinfo {pages} {700} (\bibinfo {year} {2017})}\BibitemShut {NoStop}%
	\bibitem [{\citenamefont {Caves}(1981)}]{caves1981quantum}%
	\BibitemOpen
	\bibfield  {author} {\bibinfo {author} {\bibfnamefont {C.~M.}\ \bibnamefont {Caves}},\ }\bibfield  {title} {\bibinfo {title} {Quantum-mechanical noise in an interferometer},\ }\href@noop {} {\bibfield  {journal} {\bibinfo  {journal} {Physical Review D}\ }\textbf {\bibinfo {volume} {23}},\ \bibinfo {pages} {1693} (\bibinfo {year} {1981})}\BibitemShut {NoStop}%
	\bibitem [{\citenamefont {Brida}\ \emph {et~al.}(2010)\citenamefont {Brida}, \citenamefont {Genovese},\ and\ \citenamefont {Ruo~Berchera}}]{brida2010experimental}%
	\BibitemOpen
	\bibfield  {author} {\bibinfo {author} {\bibfnamefont {G.}~\bibnamefont {Brida}}, \bibinfo {author} {\bibfnamefont {M.}~\bibnamefont {Genovese}},\ and\ \bibinfo {author} {\bibfnamefont {I.}~\bibnamefont {Ruo~Berchera}},\ }\bibfield  {title} {\bibinfo {title} {Experimental realization of sub-shot-noise quantum imaging},\ }\href@noop {} {\bibfield  {journal} {\bibinfo  {journal} {Nature Photonics}\ }\textbf {\bibinfo {volume} {4}},\ \bibinfo {pages} {227} (\bibinfo {year} {2010})}\BibitemShut {NoStop}%
	\bibitem [{\citenamefont {Taylor}\ \emph {et~al.}(2013)\citenamefont {Taylor}, \citenamefont {Janousek}, \citenamefont {Daria}, \citenamefont {Knittel}, \citenamefont {Hage}, \citenamefont {Bachor},\ and\ \citenamefont {Bowen}}]{taylor2013biological}%
	\BibitemOpen
	\bibfield  {author} {\bibinfo {author} {\bibfnamefont {M.~A.}\ \bibnamefont {Taylor}}, \bibinfo {author} {\bibfnamefont {J.}~\bibnamefont {Janousek}}, \bibinfo {author} {\bibfnamefont {V.}~\bibnamefont {Daria}}, \bibinfo {author} {\bibfnamefont {J.}~\bibnamefont {Knittel}}, \bibinfo {author} {\bibfnamefont {B.}~\bibnamefont {Hage}}, \bibinfo {author} {\bibfnamefont {H.-A.}\ \bibnamefont {Bachor}},\ and\ \bibinfo {author} {\bibfnamefont {W.~P.}\ \bibnamefont {Bowen}},\ }\bibfield  {title} {\bibinfo {title} {Biological measurement beyond the quantum limit},\ }\href@noop {} {\bibfield  {journal} {\bibinfo  {journal} {Nature Photonics}\ }\textbf {\bibinfo {volume} {7}},\ \bibinfo {pages} {229} (\bibinfo {year} {2013})}\BibitemShut {NoStop}%
	\bibitem [{\citenamefont {Zhang}\ \emph {et~al.}(2024)\citenamefont {Zhang}, \citenamefont {He}, \citenamefont {Tong}, \citenamefont {Garrett}, \citenamefont {Cao},\ and\ \citenamefont {Wang}}]{zhang2024quantum}%
	\BibitemOpen
	\bibfield  {author} {\bibinfo {author} {\bibfnamefont {Y.}~\bibnamefont {Zhang}}, \bibinfo {author} {\bibfnamefont {Z.}~\bibnamefont {He}}, \bibinfo {author} {\bibfnamefont {X.}~\bibnamefont {Tong}}, \bibinfo {author} {\bibfnamefont {D.~C.}\ \bibnamefont {Garrett}}, \bibinfo {author} {\bibfnamefont {R.}~\bibnamefont {Cao}},\ and\ \bibinfo {author} {\bibfnamefont {L.~V.}\ \bibnamefont {Wang}},\ }\bibfield  {title} {\bibinfo {title} {Quantum imaging of biological organisms through spatial and polarization entanglement},\ }\href@noop {} {\bibfield  {journal} {\bibinfo  {journal} {Science Advances}\ }\textbf {\bibinfo {volume} {10}},\ \bibinfo {pages} {eadk1495} (\bibinfo {year} {2024})}\BibitemShut {NoStop}%
	\bibitem [{\citenamefont {Lloyd}(2008)}]{lloyd2008enhanced}%
	\BibitemOpen
	\bibfield  {author} {\bibinfo {author} {\bibfnamefont {S.}~\bibnamefont {Lloyd}},\ }\bibfield  {title} {\bibinfo {title} {Enhanced sensitivity of photodetection via quantum illumination},\ }\href@noop {} {\bibfield  {journal} {\bibinfo  {journal} {Science}\ }\textbf {\bibinfo {volume} {321}},\ \bibinfo {pages} {1463} (\bibinfo {year} {2008})}\BibitemShut {NoStop}%
	\bibitem [{\citenamefont {Dylov}\ \emph {et~al.}(2011)\citenamefont {Dylov}, \citenamefont {Waller},\ and\ \citenamefont {Fleischer}}]{dylov2011nonlinear}%
	\BibitemOpen
	\bibfield  {author} {\bibinfo {author} {\bibfnamefont {D.~V.}\ \bibnamefont {Dylov}}, \bibinfo {author} {\bibfnamefont {L.}~\bibnamefont {Waller}},\ and\ \bibinfo {author} {\bibfnamefont {J.~W.}\ \bibnamefont {Fleischer}},\ }\bibfield  {title} {\bibinfo {title} {Nonlinear restoration of diffused images via seeded instability},\ }\href@noop {} {\bibfield  {journal} {\bibinfo  {journal} {IEEE Journal of Selected Topics in Quantum Electronics}\ }\textbf {\bibinfo {volume} {18}},\ \bibinfo {pages} {916} (\bibinfo {year} {2011})}\BibitemShut {NoStop}%
	\bibitem [{\citenamefont {Lopaeva}\ \emph {et~al.}(2013)\citenamefont {Lopaeva}, \citenamefont {Ruo~Berchera}, \citenamefont {Degiovanni}, \citenamefont {Olivares}, \citenamefont {Brida},\ and\ \citenamefont {Genovese}}]{lopaeva2013experimental}%
	\BibitemOpen
	\bibfield  {author} {\bibinfo {author} {\bibfnamefont {E.}~\bibnamefont {Lopaeva}}, \bibinfo {author} {\bibfnamefont {I.}~\bibnamefont {Ruo~Berchera}}, \bibinfo {author} {\bibfnamefont {I.~P.}\ \bibnamefont {Degiovanni}}, \bibinfo {author} {\bibfnamefont {S.}~\bibnamefont {Olivares}}, \bibinfo {author} {\bibfnamefont {G.}~\bibnamefont {Brida}},\ and\ \bibinfo {author} {\bibfnamefont {M.}~\bibnamefont {Genovese}},\ }\bibfield  {title} {\bibinfo {title} {Experimental realization of quantum illumination},\ }\href@noop {} {\bibfield  {journal} {\bibinfo  {journal} {Physical review letters}\ }\textbf {\bibinfo {volume} {110}},\ \bibinfo {pages} {153603} (\bibinfo {year} {2013})}\BibitemShut {NoStop}%
	\bibitem [{\citenamefont {Gregory}\ \emph {et~al.}(2020)\citenamefont {Gregory}, \citenamefont {Moreau}, \citenamefont {Toninelli},\ and\ \citenamefont {Padgett}}]{gregory2020imaging}%
	\BibitemOpen
	\bibfield  {author} {\bibinfo {author} {\bibfnamefont {T.}~\bibnamefont {Gregory}}, \bibinfo {author} {\bibfnamefont {P.-A.}\ \bibnamefont {Moreau}}, \bibinfo {author} {\bibfnamefont {E.}~\bibnamefont {Toninelli}},\ and\ \bibinfo {author} {\bibfnamefont {M.~J.}\ \bibnamefont {Padgett}},\ }\bibfield  {title} {\bibinfo {title} {Imaging through noise with quantum illumination},\ }\href@noop {} {\bibfield  {journal} {\bibinfo  {journal} {Science advances}\ }\textbf {\bibinfo {volume} {6}},\ \bibinfo {pages} {eaay2652} (\bibinfo {year} {2020})}\BibitemShut {NoStop}%
	\bibitem [{\citenamefont {Pittman}\ \emph {et~al.}(1995)\citenamefont {Pittman}, \citenamefont {Shih}, \citenamefont {Strekalov},\ and\ \citenamefont {Sergienko}}]{pittman1995optical}%
	\BibitemOpen
	\bibfield  {author} {\bibinfo {author} {\bibfnamefont {T.~B.}\ \bibnamefont {Pittman}}, \bibinfo {author} {\bibfnamefont {Y.~H.}\ \bibnamefont {Shih}}, \bibinfo {author} {\bibfnamefont {D.~V.}\ \bibnamefont {Strekalov}},\ and\ \bibinfo {author} {\bibfnamefont {A.~V.}\ \bibnamefont {Sergienko}},\ }\bibfield  {title} {\bibinfo {title} {Optical imaging by means of two-photon quantum entanglement},\ }\href@noop {} {\bibfield  {journal} {\bibinfo  {journal} {Physical Review A}\ }\textbf {\bibinfo {volume} {52}},\ \bibinfo {pages} {R3429} (\bibinfo {year} {1995})}\BibitemShut {NoStop}%
	\bibitem [{\citenamefont {Strekalov}\ \emph {et~al.}(1995)\citenamefont {Strekalov}, \citenamefont {Sergienko}, \citenamefont {Klyshko},\ and\ \citenamefont {Shih}}]{strekalov1995observation}%
	\BibitemOpen
	\bibfield  {author} {\bibinfo {author} {\bibfnamefont {D.}~\bibnamefont {Strekalov}}, \bibinfo {author} {\bibfnamefont {A.}~\bibnamefont {Sergienko}}, \bibinfo {author} {\bibfnamefont {D.}~\bibnamefont {Klyshko}},\ and\ \bibinfo {author} {\bibfnamefont {Y.}~\bibnamefont {Shih}},\ }\bibfield  {title} {\bibinfo {title} {Observation of two-photon “ghost” interference and diffraction},\ }\href@noop {} {\bibfield  {journal} {\bibinfo  {journal} {Physical review letters}\ }\textbf {\bibinfo {volume} {74}},\ \bibinfo {pages} {3600} (\bibinfo {year} {1995})}\BibitemShut {NoStop}%
	\bibitem [{\citenamefont {Bennink}\ \emph {et~al.}(2002)\citenamefont {Bennink}, \citenamefont {Bentley},\ and\ \citenamefont {Boyd}}]{bennink2002two}%
	\BibitemOpen
	\bibfield  {author} {\bibinfo {author} {\bibfnamefont {R.~S.}\ \bibnamefont {Bennink}}, \bibinfo {author} {\bibfnamefont {S.~J.}\ \bibnamefont {Bentley}},\ and\ \bibinfo {author} {\bibfnamefont {R.~W.}\ \bibnamefont {Boyd}},\ }\bibfield  {title} {\bibinfo {title} {“two-photon” coincidence imaging with a classical source},\ }\href@noop {} {\bibfield  {journal} {\bibinfo  {journal} {Physical review letters}\ }\textbf {\bibinfo {volume} {89}},\ \bibinfo {pages} {113601} (\bibinfo {year} {2002})}\BibitemShut {NoStop}%
	\bibitem [{\citenamefont {Ortolano}\ \emph {et~al.}(2023)\citenamefont {Ortolano}, \citenamefont {Napoli}, \citenamefont {Harney}, \citenamefont {Pirandola}, \citenamefont {Leonetti}, \citenamefont {Boucher}, \citenamefont {Losero}, \citenamefont {Genovese},\ and\ \citenamefont {Ruo-Berchera}}]{ortolano2023quantum}%
	\BibitemOpen
	\bibfield  {author} {\bibinfo {author} {\bibfnamefont {G.}~\bibnamefont {Ortolano}}, \bibinfo {author} {\bibfnamefont {C.}~\bibnamefont {Napoli}}, \bibinfo {author} {\bibfnamefont {C.}~\bibnamefont {Harney}}, \bibinfo {author} {\bibfnamefont {S.}~\bibnamefont {Pirandola}}, \bibinfo {author} {\bibfnamefont {G.}~\bibnamefont {Leonetti}}, \bibinfo {author} {\bibfnamefont {P.}~\bibnamefont {Boucher}}, \bibinfo {author} {\bibfnamefont {E.}~\bibnamefont {Losero}}, \bibinfo {author} {\bibfnamefont {M.}~\bibnamefont {Genovese}},\ and\ \bibinfo {author} {\bibfnamefont {I.}~\bibnamefont {Ruo-Berchera}},\ }\bibfield  {title} {\bibinfo {title} {Quantum-enhanced pattern recognition},\ }\href@noop {} {\bibfield  {journal} {\bibinfo  {journal} {Physical Review Applied}\ }\textbf {\bibinfo {volume} {20}},\ \bibinfo {pages} {024072} (\bibinfo {year} {2023})}\BibitemShut {NoStop}%
	\bibitem [{\citenamefont {Giovannetti}\ \emph {et~al.}(2009)\citenamefont {Giovannetti}, \citenamefont {Lloyd}, \citenamefont {Maccone},\ and\ \citenamefont {Shapiro}}]{giovannetti2009sub}%
	\BibitemOpen
	\bibfield  {author} {\bibinfo {author} {\bibfnamefont {V.}~\bibnamefont {Giovannetti}}, \bibinfo {author} {\bibfnamefont {S.}~\bibnamefont {Lloyd}}, \bibinfo {author} {\bibfnamefont {L.}~\bibnamefont {Maccone}},\ and\ \bibinfo {author} {\bibfnamefont {J.~H.}\ \bibnamefont {Shapiro}},\ }\bibfield  {title} {\bibinfo {title} {Sub-rayleigh-diffraction-bound quantum imaging},\ }\href@noop {} {\bibfield  {journal} {\bibinfo  {journal} {Physical Review A—Atomic, Molecular, and Optical Physics}\ }\textbf {\bibinfo {volume} {79}},\ \bibinfo {pages} {013827} (\bibinfo {year} {2009})}\BibitemShut {NoStop}%
	\bibitem [{\citenamefont {Toninelli}\ \emph {et~al.}(2019)\citenamefont {Toninelli}, \citenamefont {Moreau}, \citenamefont {Gregory}, \citenamefont {Mihalyi}, \citenamefont {Edgar}, \citenamefont {Radwell},\ and\ \citenamefont {Padgett}}]{toninelli2019resolution}%
	\BibitemOpen
	\bibfield  {author} {\bibinfo {author} {\bibfnamefont {E.}~\bibnamefont {Toninelli}}, \bibinfo {author} {\bibfnamefont {P.-A.}\ \bibnamefont {Moreau}}, \bibinfo {author} {\bibfnamefont {T.}~\bibnamefont {Gregory}}, \bibinfo {author} {\bibfnamefont {A.}~\bibnamefont {Mihalyi}}, \bibinfo {author} {\bibfnamefont {M.}~\bibnamefont {Edgar}}, \bibinfo {author} {\bibfnamefont {N.}~\bibnamefont {Radwell}},\ and\ \bibinfo {author} {\bibfnamefont {M.}~\bibnamefont {Padgett}},\ }\bibfield  {title} {\bibinfo {title} {Resolution-enhanced quantum imaging by centroid estimation of biphotons},\ }\href@noop {} {\bibfield  {journal} {\bibinfo  {journal} {Optica}\ }\textbf {\bibinfo {volume} {6}},\ \bibinfo {pages} {347} (\bibinfo {year} {2019})}\BibitemShut {NoStop}%
	\bibitem [{\citenamefont {He}\ \emph {et~al.}(2023)\citenamefont {He}, \citenamefont {Zhang}, \citenamefont {Tong}, \citenamefont {Li},\ and\ \citenamefont {Wang}}]{he2023quantum}%
	\BibitemOpen
	\bibfield  {author} {\bibinfo {author} {\bibfnamefont {Z.}~\bibnamefont {He}}, \bibinfo {author} {\bibfnamefont {Y.}~\bibnamefont {Zhang}}, \bibinfo {author} {\bibfnamefont {X.}~\bibnamefont {Tong}}, \bibinfo {author} {\bibfnamefont {L.}~\bibnamefont {Li}},\ and\ \bibinfo {author} {\bibfnamefont {L.~V.}\ \bibnamefont {Wang}},\ }\bibfield  {title} {\bibinfo {title} {Quantum microscopy of cells at the heisenberg limit},\ }\href@noop {} {\bibfield  {journal} {\bibinfo  {journal} {Nature communications}\ }\textbf {\bibinfo {volume} {14}},\ \bibinfo {pages} {2441} (\bibinfo {year} {2023})}\BibitemShut {NoStop}%
	\bibitem [{\citenamefont {Grochowski}\ \emph {et~al.}(2026)\citenamefont {Grochowski}, \citenamefont {Fadel},\ and\ \citenamefont {Filip}}]{grochowski2026distributed}%
	\BibitemOpen
	\bibfield  {author} {\bibinfo {author} {\bibfnamefont {P.~T.}\ \bibnamefont {Grochowski}}, \bibinfo {author} {\bibfnamefont {M.}~\bibnamefont {Fadel}},\ and\ \bibinfo {author} {\bibfnamefont {R.}~\bibnamefont {Filip}},\ }\bibfield  {title} {\bibinfo {title} {Distributed phase-insensitive displacement sensing},\ }\href@noop {} {\bibfield  {journal} {\bibinfo  {journal} {arXiv preprint arXiv:2602.03727}\ } (\bibinfo {year} {2026})}\BibitemShut {NoStop}%
	\bibitem [{\citenamefont {El-Nouby}\ \emph {et~al.}(2021)\citenamefont {El-Nouby}, \citenamefont {Touvron}, \citenamefont {Caron}, \citenamefont {Bojanowski}, \citenamefont {Douze}, \citenamefont {Joulin}, \citenamefont {Laptev}, \citenamefont {Neverova}, \citenamefont {Synnaeve}, \citenamefont {Verbeek} \emph {et~al.}}]{el2021xcit}%
	\BibitemOpen
	\bibfield  {author} {\bibinfo {author} {\bibfnamefont {A.}~\bibnamefont {El-Nouby}}, \bibinfo {author} {\bibfnamefont {H.}~\bibnamefont {Touvron}}, \bibinfo {author} {\bibfnamefont {M.}~\bibnamefont {Caron}}, \bibinfo {author} {\bibfnamefont {P.}~\bibnamefont {Bojanowski}}, \bibinfo {author} {\bibfnamefont {M.}~\bibnamefont {Douze}}, \bibinfo {author} {\bibfnamefont {A.}~\bibnamefont {Joulin}}, \bibinfo {author} {\bibfnamefont {I.}~\bibnamefont {Laptev}}, \bibinfo {author} {\bibfnamefont {N.}~\bibnamefont {Neverova}}, \bibinfo {author} {\bibfnamefont {G.}~\bibnamefont {Synnaeve}}, \bibinfo {author} {\bibfnamefont {J.}~\bibnamefont {Verbeek}}, \emph {et~al.},\ }\bibfield  {title} {\bibinfo {title} {Xcit: Cross-covariance image transformers},\ }\href@noop {} {\bibfield  {journal} {\bibinfo  {journal} {arXiv preprint arXiv:2106.09681}\ } (\bibinfo {year} {2021})}\BibitemShut {NoStop}%
	\bibitem [{\citenamefont {Defienne}\ \emph {et~al.}(2018)\citenamefont {Defienne}, \citenamefont {Reichert},\ and\ \citenamefont {Fleischer}}]{defienne2018adaptive}%
	\BibitemOpen
	\bibfield  {author} {\bibinfo {author} {\bibfnamefont {H.}~\bibnamefont {Defienne}}, \bibinfo {author} {\bibfnamefont {M.}~\bibnamefont {Reichert}},\ and\ \bibinfo {author} {\bibfnamefont {J.~W.}\ \bibnamefont {Fleischer}},\ }\bibfield  {title} {\bibinfo {title} {Adaptive quantum optics with spatially entangled photon pairs},\ }\href@noop {} {\bibfield  {journal} {\bibinfo  {journal} {Physical review letters}\ }\textbf {\bibinfo {volume} {121}},\ \bibinfo {pages} {233601} (\bibinfo {year} {2018})}\BibitemShut {NoStop}%
	\bibitem [{\citenamefont {Lib}\ \emph {et~al.}(2020)\citenamefont {Lib}, \citenamefont {Hasson},\ and\ \citenamefont {Bromberg}}]{lib2020real}%
	\BibitemOpen
	\bibfield  {author} {\bibinfo {author} {\bibfnamefont {O.}~\bibnamefont {Lib}}, \bibinfo {author} {\bibfnamefont {G.}~\bibnamefont {Hasson}},\ and\ \bibinfo {author} {\bibfnamefont {Y.}~\bibnamefont {Bromberg}},\ }\bibfield  {title} {\bibinfo {title} {Real-time shaping of entangled photons by classical control and feedback},\ }\href@noop {} {\bibfield  {journal} {\bibinfo  {journal} {Science Advances}\ }\textbf {\bibinfo {volume} {6}},\ \bibinfo {pages} {eabb6298} (\bibinfo {year} {2020})}\BibitemShut {NoStop}%
	\bibitem [{\citenamefont {Cameron}\ \emph {et~al.}(2024{\natexlab{a}})\citenamefont {Cameron}, \citenamefont {Courme}, \citenamefont {Verni{\`e}re}, \citenamefont {Pandya}, \citenamefont {Faccio},\ and\ \citenamefont {Defienne}}]{cameron2024adaptive}%
	\BibitemOpen
	\bibfield  {author} {\bibinfo {author} {\bibfnamefont {P.}~\bibnamefont {Cameron}}, \bibinfo {author} {\bibfnamefont {B.}~\bibnamefont {Courme}}, \bibinfo {author} {\bibfnamefont {C.}~\bibnamefont {Verni{\`e}re}}, \bibinfo {author} {\bibfnamefont {R.}~\bibnamefont {Pandya}}, \bibinfo {author} {\bibfnamefont {D.}~\bibnamefont {Faccio}},\ and\ \bibinfo {author} {\bibfnamefont {H.}~\bibnamefont {Defienne}},\ }\bibfield  {title} {\bibinfo {title} {Adaptive optical imaging with entangled photons},\ }\href@noop {} {\bibfield  {journal} {\bibinfo  {journal} {Science}\ }\textbf {\bibinfo {volume} {383}},\ \bibinfo {pages} {1142} (\bibinfo {year} {2024}{\natexlab{a}})}\BibitemShut {NoStop}%
	\bibitem [{\citenamefont {Boucher}\ \emph {et~al.}(2021)\citenamefont {Boucher}, \citenamefont {Defienne},\ and\ \citenamefont {Gigan}}]{boucher2021engineering}%
	\BibitemOpen
	\bibfield  {author} {\bibinfo {author} {\bibfnamefont {P.}~\bibnamefont {Boucher}}, \bibinfo {author} {\bibfnamefont {H.}~\bibnamefont {Defienne}},\ and\ \bibinfo {author} {\bibfnamefont {S.}~\bibnamefont {Gigan}},\ }\bibfield  {title} {\bibinfo {title} {Engineering spatial correlations of entangled photon pairs by pump beam shaping},\ }\href@noop {} {\bibfield  {journal} {\bibinfo  {journal} {Optics Letters}\ }\textbf {\bibinfo {volume} {46}},\ \bibinfo {pages} {4200} (\bibinfo {year} {2021})}\BibitemShut {NoStop}%
	\bibitem [{\citenamefont {Nirala}\ \emph {et~al.}(2023)\citenamefont {Nirala}, \citenamefont {Pradyumna}, \citenamefont {Kumar},\ and\ \citenamefont {Marino}}]{nirala2023information}%
	\BibitemOpen
	\bibfield  {author} {\bibinfo {author} {\bibfnamefont {G.}~\bibnamefont {Nirala}}, \bibinfo {author} {\bibfnamefont {S.~T.}\ \bibnamefont {Pradyumna}}, \bibinfo {author} {\bibfnamefont {A.}~\bibnamefont {Kumar}},\ and\ \bibinfo {author} {\bibfnamefont {A.~M.}\ \bibnamefont {Marino}},\ }\bibfield  {title} {\bibinfo {title} {Information encoding in the spatial correlations of entangled twin beams},\ }\href@noop {} {\bibfield  {journal} {\bibinfo  {journal} {Science Advances}\ }\textbf {\bibinfo {volume} {9}},\ \bibinfo {pages} {eadf9161} (\bibinfo {year} {2023})}\BibitemShut {NoStop}%
	\bibitem [{\citenamefont {Verni{\`e}re}\ and\ \citenamefont {Defienne}(2024)}]{verniere2024hiding}%
	\BibitemOpen
	\bibfield  {author} {\bibinfo {author} {\bibfnamefont {C.}~\bibnamefont {Verni{\`e}re}}\ and\ \bibinfo {author} {\bibfnamefont {H.}~\bibnamefont {Defienne}},\ }\bibfield  {title} {\bibinfo {title} {Hiding images in quantum correlations},\ }\href@noop {} {\bibfield  {journal} {\bibinfo  {journal} {Physical Review Letters}\ }\textbf {\bibinfo {volume} {133}},\ \bibinfo {pages} {093601} (\bibinfo {year} {2024})}\BibitemShut {NoStop}%
	\bibitem [{\citenamefont {Lee}\ \emph {et~al.}(2019)\citenamefont {Lee}, \citenamefont {Lee}, \citenamefont {Kim}, \citenamefont {Kosiorek}, \citenamefont {Choi},\ and\ \citenamefont {Teh}}]{lee2019settransformer}%
	\BibitemOpen
	\bibfield  {author} {\bibinfo {author} {\bibfnamefont {J.}~\bibnamefont {Lee}}, \bibinfo {author} {\bibfnamefont {Y.}~\bibnamefont {Lee}}, \bibinfo {author} {\bibfnamefont {J.}~\bibnamefont {Kim}}, \bibinfo {author} {\bibfnamefont {A.}~\bibnamefont {Kosiorek}}, \bibinfo {author} {\bibfnamefont {S.}~\bibnamefont {Choi}},\ and\ \bibinfo {author} {\bibfnamefont {Y.~W.}\ \bibnamefont {Teh}},\ }\bibfield  {title} {\bibinfo {title} {Set transformer: A framework for attention-based permutation-invariant neural networks},\ }in\ \href {https://proceedings.mlr.press/v97/lee19d.html} {\emph {\bibinfo {booktitle} {Proceedings of the 36th International Conference on Machine Learning}}},\ \bibinfo {series} {Proceedings of Machine Learning Research}, Vol.~\bibinfo {volume} {97},\ \bibinfo {editor} {edited by\ \bibinfo {editor} {\bibfnamefont {K.}~\bibnamefont {Chaudhuri}}\ and\ \bibinfo {editor} {\bibfnamefont {R.}~\bibnamefont {Salakhutdinov}}}\ (\bibinfo  {publisher} {PMLR},\ \bibinfo {year} {2019})\ pp.\ \bibinfo
	{pages} {3744--3753}\BibitemShut {NoStop}%
	\bibitem [{\citenamefont {Kim}\ \emph {et~al.}(2024)\citenamefont {Kim}, \citenamefont {Zhou}, \citenamefont {Xu}, \citenamefont {Varma}, \citenamefont {Karamlou}, \citenamefont {Rosen}, \citenamefont {Hoke}, \citenamefont {Wan}, \citenamefont {Zhou}, \citenamefont {Oliver}, \citenamefont {Lensky}, \citenamefont {Weinberger},\ and\ \citenamefont {Kim}}]{kim2024attentionquantumcomplexity}%
	\BibitemOpen
	\bibfield  {author} {\bibinfo {author} {\bibfnamefont {H.}~\bibnamefont {Kim}}, \bibinfo {author} {\bibfnamefont {Y.}~\bibnamefont {Zhou}}, \bibinfo {author} {\bibfnamefont {Y.}~\bibnamefont {Xu}}, \bibinfo {author} {\bibfnamefont {K.}~\bibnamefont {Varma}}, \bibinfo {author} {\bibfnamefont {A.~H.}\ \bibnamefont {Karamlou}}, \bibinfo {author} {\bibfnamefont {I.~T.}\ \bibnamefont {Rosen}}, \bibinfo {author} {\bibfnamefont {J.~C.}\ \bibnamefont {Hoke}}, \bibinfo {author} {\bibfnamefont {C.}~\bibnamefont {Wan}}, \bibinfo {author} {\bibfnamefont {J.~P.}\ \bibnamefont {Zhou}}, \bibinfo {author} {\bibfnamefont {W.~D.}\ \bibnamefont {Oliver}}, \bibinfo {author} {\bibfnamefont {Y.~D.}\ \bibnamefont {Lensky}}, \bibinfo {author} {\bibfnamefont {K.~Q.}\ \bibnamefont {Weinberger}},\ and\ \bibinfo {author} {\bibfnamefont {E.-A.}\ \bibnamefont {Kim}},\ }\href {https://arxiv.org/abs/2405.11632} {\bibinfo {title} {Attention to quantum complexity}} (\bibinfo {year} {2024}),\ \Eprint {https://arxiv.org/abs/2405.11632}
	{arXiv:2405.11632 [quant-ph]} \BibitemShut {NoStop}%
	\bibitem [{\citenamefont {Cameron}\ \emph {et~al.}(2024{\natexlab{b}})\citenamefont {Cameron}, \citenamefont {Courme}, \citenamefont {Faccio},\ and\ \citenamefont {Defienne}}]{cameron2024shaping}%
	\BibitemOpen
	\bibfield  {author} {\bibinfo {author} {\bibfnamefont {P.}~\bibnamefont {Cameron}}, \bibinfo {author} {\bibfnamefont {B.}~\bibnamefont {Courme}}, \bibinfo {author} {\bibfnamefont {D.}~\bibnamefont {Faccio}},\ and\ \bibinfo {author} {\bibfnamefont {H.}~\bibnamefont {Defienne}},\ }\bibfield  {title} {\bibinfo {title} {Shaping the spatial correlations of entangled photon pairs},\ }\href@noop {} {\bibfield  {journal} {\bibinfo  {journal} {Journal of Physics: Photonics}\ }\textbf {\bibinfo {volume} {6}},\ \bibinfo {pages} {033001} (\bibinfo {year} {2024}{\natexlab{b}})}\BibitemShut {NoStop}%
	\bibitem [{\citenamefont {Walborn}\ \emph {et~al.}(2010)\citenamefont {Walborn}, \citenamefont {Monken}, \citenamefont {P{\'a}dua},\ and\ \citenamefont {Ribeiro}}]{walborn2010spatial}%
	\BibitemOpen
	\bibfield  {author} {\bibinfo {author} {\bibfnamefont {S.~P.}\ \bibnamefont {Walborn}}, \bibinfo {author} {\bibfnamefont {C.}~\bibnamefont {Monken}}, \bibinfo {author} {\bibfnamefont {S.}~\bibnamefont {P{\'a}dua}},\ and\ \bibinfo {author} {\bibfnamefont {P.~S.}\ \bibnamefont {Ribeiro}},\ }\bibfield  {title} {\bibinfo {title} {Spatial correlations in parametric down-conversion},\ }\href@noop {} {\bibfield  {journal} {\bibinfo  {journal} {Physics Reports}\ }\textbf {\bibinfo {volume} {495}},\ \bibinfo {pages} {87} (\bibinfo {year} {2010})}\BibitemShut {NoStop}%
	\bibitem [{\citenamefont {Wasilewski}\ \emph {et~al.}(2006)\citenamefont {Wasilewski}, \citenamefont {Lvovsky}, \citenamefont {Banaszek},\ and\ \citenamefont {Radzewicz}}]{wasilewski2006pulsed}%
	\BibitemOpen
	\bibfield  {author} {\bibinfo {author} {\bibfnamefont {W.}~\bibnamefont {Wasilewski}}, \bibinfo {author} {\bibfnamefont {A.~I.}\ \bibnamefont {Lvovsky}}, \bibinfo {author} {\bibfnamefont {K.}~\bibnamefont {Banaszek}},\ and\ \bibinfo {author} {\bibfnamefont {C.}~\bibnamefont {Radzewicz}},\ }\bibfield  {title} {\bibinfo {title} {Pulsed squeezed light: Simultaneous squeezing of multiple modes},\ }\href@noop {} {\bibfield  {journal} {\bibinfo  {journal} {Physical Review A—Atomic, Molecular, and Optical Physics}\ }\textbf {\bibinfo {volume} {73}},\ \bibinfo {pages} {063819} (\bibinfo {year} {2006})}\BibitemShut {NoStop}%
	\bibitem [{\citenamefont {Quesada}\ \emph {et~al.}(2022)\citenamefont {Quesada}, \citenamefont {Helt}, \citenamefont {Menotti}, \citenamefont {Liscidini},\ and\ \citenamefont {Sipe}}]{quesada2022beyond}%
	\BibitemOpen
	\bibfield  {author} {\bibinfo {author} {\bibfnamefont {N.}~\bibnamefont {Quesada}}, \bibinfo {author} {\bibfnamefont {L.}~\bibnamefont {Helt}}, \bibinfo {author} {\bibfnamefont {M.}~\bibnamefont {Menotti}}, \bibinfo {author} {\bibfnamefont {M.}~\bibnamefont {Liscidini}},\ and\ \bibinfo {author} {\bibfnamefont {J.}~\bibnamefont {Sipe}},\ }\bibfield  {title} {\bibinfo {title} {Beyond photon pairs—nonlinear quantum photonics in the high-gain regime: a tutorial},\ }\href@noop {} {\bibfield  {journal} {\bibinfo  {journal} {Advances in Optics and Photonics}\ }\textbf {\bibinfo {volume} {14}},\ \bibinfo {pages} {291} (\bibinfo {year} {2022})}\BibitemShut {NoStop}%
	\bibitem [{\citenamefont {Law}\ and\ \citenamefont {Eberly}(2004)}]{law2004analysis}%
	\BibitemOpen
	\bibfield  {author} {\bibinfo {author} {\bibfnamefont {C.}~\bibnamefont {Law}}\ and\ \bibinfo {author} {\bibfnamefont {J.}~\bibnamefont {Eberly}},\ }\bibfield  {title} {\bibinfo {title} {Analysis and interpretation of high transverse entanglement<? format?> in optical parametric down conversion},\ }\href@noop {} {\bibfield  {journal} {\bibinfo  {journal} {Physical review letters}\ }\textbf {\bibinfo {volume} {92}},\ \bibinfo {pages} {127903} (\bibinfo {year} {2004})}\BibitemShut {NoStop}%
	\bibitem [{\citenamefont {Brambilla}\ \emph {et~al.}(2004)\citenamefont {Brambilla}, \citenamefont {Gatti}, \citenamefont {Bache},\ and\ \citenamefont {Lugiato}}]{brambilla2004simultaneous}%
	\BibitemOpen
	\bibfield  {author} {\bibinfo {author} {\bibfnamefont {E.}~\bibnamefont {Brambilla}}, \bibinfo {author} {\bibfnamefont {A.}~\bibnamefont {Gatti}}, \bibinfo {author} {\bibfnamefont {M.}~\bibnamefont {Bache}},\ and\ \bibinfo {author} {\bibfnamefont {L.~A.}\ \bibnamefont {Lugiato}},\ }\bibfield  {title} {\bibinfo {title} {Simultaneous near-field and far-field spatial quantum correlations in the high-gain regime of parametric down-conversion},\ }\href@noop {} {\bibfield  {journal} {\bibinfo  {journal} {Physical Review A}\ }\textbf {\bibinfo {volume} {69}},\ \bibinfo {pages} {023802} (\bibinfo {year} {2004})}\BibitemShut {NoStop}%
	\bibitem [{\citenamefont {Gatti}\ \emph {et~al.}(1999)\citenamefont {Gatti}, \citenamefont {Brambilla}, \citenamefont {Lugiato},\ and\ \citenamefont {Kolobov}}]{gatti1999quantum}%
	\BibitemOpen
	\bibfield  {author} {\bibinfo {author} {\bibfnamefont {A.}~\bibnamefont {Gatti}}, \bibinfo {author} {\bibfnamefont {E.}~\bibnamefont {Brambilla}}, \bibinfo {author} {\bibfnamefont {L.}~\bibnamefont {Lugiato}},\ and\ \bibinfo {author} {\bibfnamefont {M.}~\bibnamefont {Kolobov}},\ }\bibfield  {title} {\bibinfo {title} {Quantum entangled images},\ }\href@noop {} {\bibfield  {journal} {\bibinfo  {journal} {Physical review letters}\ }\textbf {\bibinfo {volume} {83}},\ \bibinfo {pages} {1763} (\bibinfo {year} {1999})}\BibitemShut {NoStop}%
	\bibitem [{\citenamefont {Gatti}\ \emph {et~al.}(2004)\citenamefont {Gatti}, \citenamefont {Brambilla}, \citenamefont {Bache},\ and\ \citenamefont {Lugiato}}]{gatti2004ghost}%
	\BibitemOpen
	\bibfield  {author} {\bibinfo {author} {\bibfnamefont {A.}~\bibnamefont {Gatti}}, \bibinfo {author} {\bibfnamefont {E.}~\bibnamefont {Brambilla}}, \bibinfo {author} {\bibfnamefont {M.}~\bibnamefont {Bache}},\ and\ \bibinfo {author} {\bibfnamefont {L.~A.}\ \bibnamefont {Lugiato}},\ }\bibfield  {title} {\bibinfo {title} {Ghost imaging with thermal light: comparing entanglement and classical correlation},\ }\href@noop {} {\bibfield  {journal} {\bibinfo  {journal} {Physical review letters}\ }\textbf {\bibinfo {volume} {93}},\ \bibinfo {pages} {093602} (\bibinfo {year} {2004})}\BibitemShut {NoStop}%
	\bibitem [{\citenamefont {Shapiro}(2008)}]{shapiro2008computational}%
	\BibitemOpen
	\bibfield  {author} {\bibinfo {author} {\bibfnamefont {J.~H.}\ \bibnamefont {Shapiro}},\ }\bibfield  {title} {\bibinfo {title} {Computational ghost imaging},\ }\href@noop {} {\bibfield  {journal} {\bibinfo  {journal} {Physical Review A—Atomic, Molecular, and Optical Physics}\ }\textbf {\bibinfo {volume} {78}},\ \bibinfo {pages} {061802} (\bibinfo {year} {2008})}\BibitemShut {NoStop}%
	\bibitem [{\citenamefont {Wright}\ \emph {et~al.}(2022)\citenamefont {Wright}, \citenamefont {Onodera}, \citenamefont {Stein}, \citenamefont {Wang}, \citenamefont {Schachter}, \citenamefont {Hu},\ and\ \citenamefont {McMahon}}]{wright2022deep}%
	\BibitemOpen
	\bibfield  {author} {\bibinfo {author} {\bibfnamefont {L.~G.}\ \bibnamefont {Wright}}, \bibinfo {author} {\bibfnamefont {T.}~\bibnamefont {Onodera}}, \bibinfo {author} {\bibfnamefont {M.~M.}\ \bibnamefont {Stein}}, \bibinfo {author} {\bibfnamefont {T.}~\bibnamefont {Wang}}, \bibinfo {author} {\bibfnamefont {D.~T.}\ \bibnamefont {Schachter}}, \bibinfo {author} {\bibfnamefont {Z.}~\bibnamefont {Hu}},\ and\ \bibinfo {author} {\bibfnamefont {P.~L.}\ \bibnamefont {McMahon}},\ }\bibfield  {title} {\bibinfo {title} {Deep physical neural networks trained with backpropagation},\ }\href@noop {} {\bibfield  {journal} {\bibinfo  {journal} {Nature}\ }\textbf {\bibinfo {volume} {601}},\ \bibinfo {pages} {549} (\bibinfo {year} {2022})}\BibitemShut {NoStop}%
	\bibitem [{\citenamefont {Jang}\ \emph {et~al.}(2017)\citenamefont {Jang}, \citenamefont {Gu},\ and\ \citenamefont {Poole}}]{jang2017categorical}%
	\BibitemOpen
	\bibfield  {author} {\bibinfo {author} {\bibfnamefont {E.}~\bibnamefont {Jang}}, \bibinfo {author} {\bibfnamefont {S.}~\bibnamefont {Gu}},\ and\ \bibinfo {author} {\bibfnamefont {B.}~\bibnamefont {Poole}},\ }\bibfield  {title} {\bibinfo {title} {Categorical reparameterization with gumbel-softmax},\ }in\ \href {https://openreview.net/forum?id=rkE3y85ee} {\emph {\bibinfo {booktitle} {International Conference on Learning Representations (ICLR)}}}\ (\bibinfo {year} {2017})\BibitemShut {NoStop}%
	\bibitem [{\citenamefont {Maddison}\ \emph {et~al.}(2017)\citenamefont {Maddison}, \citenamefont {Mnih},\ and\ \citenamefont {Teh}}]{maddison2017concrete}%
	\BibitemOpen
	\bibfield  {author} {\bibinfo {author} {\bibfnamefont {C.~J.}\ \bibnamefont {Maddison}}, \bibinfo {author} {\bibfnamefont {A.}~\bibnamefont {Mnih}},\ and\ \bibinfo {author} {\bibfnamefont {Y.~W.}\ \bibnamefont {Teh}},\ }\bibfield  {title} {\bibinfo {title} {The concrete distribution: A continuous relaxation of discrete random variables},\ }in\ \href {https://openreview.net/forum?id=S1jE5L5gl} {\emph {\bibinfo {booktitle} {International Conference on Learning Representations (ICLR)}}}\ (\bibinfo {year} {2017})\BibitemShut {NoStop}%
	\bibitem [{\citenamefont {Bengio}\ \emph {et~al.}(2013)\citenamefont {Bengio}, \citenamefont {L{\'e}onard},\ and\ \citenamefont {Courville}}]{bengio2013estimating}%
	\BibitemOpen
	\bibfield  {author} {\bibinfo {author} {\bibfnamefont {Y.}~\bibnamefont {Bengio}}, \bibinfo {author} {\bibfnamefont {N.}~\bibnamefont {L{\'e}onard}},\ and\ \bibinfo {author} {\bibfnamefont {A.}~\bibnamefont {Courville}},\ }\bibfield  {title} {\bibinfo {title} {Estimating or propagating gradients through stochastic neurons for conditional computation},\ }\href {https://arxiv.org/abs/1308.3432} {\bibfield  {journal} {\bibinfo  {journal} {arXiv preprint arXiv:1308.3432}\ } (\bibinfo {year} {2013})}\BibitemShut {NoStop}%
	\bibitem [{\citenamefont {Yang}\ \emph {et~al.}(2012)\citenamefont {Yang}, \citenamefont {Prasad},\ and\ \citenamefont {Latecki}}]{yang2012affinity}%
	\BibitemOpen
	\bibfield  {author} {\bibinfo {author} {\bibfnamefont {X.}~\bibnamefont {Yang}}, \bibinfo {author} {\bibfnamefont {L.}~\bibnamefont {Prasad}},\ and\ \bibinfo {author} {\bibfnamefont {L.~J.}\ \bibnamefont {Latecki}},\ }\bibfield  {title} {\bibinfo {title} {Affinity learning with diffusion on tensor product graph},\ }\href@noop {} {\bibfield  {journal} {\bibinfo  {journal} {IEEE transactions on pattern analysis and machine intelligence}\ }\textbf {\bibinfo {volume} {35}},\ \bibinfo {pages} {28} (\bibinfo {year} {2012})}\BibitemShut {NoStop}%
	\bibitem [{\citenamefont {Guerra}(1995)}]{guerra1995super}%
	\BibitemOpen
	\bibfield  {author} {\bibinfo {author} {\bibfnamefont {J.~M.}\ \bibnamefont {Guerra}},\ }\bibfield  {title} {\bibinfo {title} {Super-resolution through illumination by diffraction-born evanescent waves},\ }\href@noop {} {\bibfield  {journal} {\bibinfo  {journal} {Applied physics letters}\ }\textbf {\bibinfo {volume} {66}},\ \bibinfo {pages} {3555} (\bibinfo {year} {1995})}\BibitemShut {NoStop}%
	\bibitem [{\citenamefont {Gustafsson}(2000)}]{gustafsson2000surpassing}%
	\BibitemOpen
	\bibfield  {author} {\bibinfo {author} {\bibfnamefont {M.~G.}\ \bibnamefont {Gustafsson}},\ }\bibfield  {title} {\bibinfo {title} {Surpassing the lateral resolution limit by a factor of two using structured illumination microscopy},\ }\href@noop {} {\bibfield  {journal} {\bibinfo  {journal} {Journal of microscopy}\ }\textbf {\bibinfo {volume} {198}},\ \bibinfo {pages} {82} (\bibinfo {year} {2000})}\BibitemShut {NoStop}%
	\bibitem [{\citenamefont {Rego}\ and\ \citenamefont {Shao}(2014)}]{rego2014practical}%
	\BibitemOpen
	\bibfield  {author} {\bibinfo {author} {\bibfnamefont {E.~H.}\ \bibnamefont {Rego}}\ and\ \bibinfo {author} {\bibfnamefont {L.}~\bibnamefont {Shao}},\ }\bibfield  {title} {\bibinfo {title} {Practical structured illumination microscopy},\ }in\ \href@noop {} {\emph {\bibinfo {booktitle} {Advanced Fluorescence Microscopy: Methods and Protocols}}}\ (\bibinfo  {publisher} {Springer},\ \bibinfo {year} {2014})\ pp.\ \bibinfo {pages} {175--192}\BibitemShut {NoStop}%
	\bibitem [{\citenamefont {Schraivogel}\ \emph {et~al.}(2022)\citenamefont {Schraivogel}, \citenamefont {Kuhn}, \citenamefont {Rauscher}, \citenamefont {Rodr{\'\i}guez-Mart{\'\i}nez}, \citenamefont {Paulsen}, \citenamefont {Owsley}, \citenamefont {Middlebrook}, \citenamefont {Tischer}, \citenamefont {Ramasz}, \citenamefont {Ordo{\~n}ez-Rueda} \emph {et~al.}}]{schraivogel2022high}%
	\BibitemOpen
	\bibfield  {author} {\bibinfo {author} {\bibfnamefont {D.}~\bibnamefont {Schraivogel}}, \bibinfo {author} {\bibfnamefont {T.~M.}\ \bibnamefont {Kuhn}}, \bibinfo {author} {\bibfnamefont {B.}~\bibnamefont {Rauscher}}, \bibinfo {author} {\bibfnamefont {M.}~\bibnamefont {Rodr{\'\i}guez-Mart{\'\i}nez}}, \bibinfo {author} {\bibfnamefont {M.}~\bibnamefont {Paulsen}}, \bibinfo {author} {\bibfnamefont {K.}~\bibnamefont {Owsley}}, \bibinfo {author} {\bibfnamefont {A.}~\bibnamefont {Middlebrook}}, \bibinfo {author} {\bibfnamefont {C.}~\bibnamefont {Tischer}}, \bibinfo {author} {\bibfnamefont {B.}~\bibnamefont {Ramasz}}, \bibinfo {author} {\bibfnamefont {D.}~\bibnamefont {Ordo{\~n}ez-Rueda}}, \emph {et~al.},\ }\bibfield  {title} {\bibinfo {title} {High-speed fluorescence image--enabled cell sorting},\ }\href@noop {} {\bibfield  {journal} {\bibinfo  {journal} {Science}\ }\textbf {\bibinfo {volume} {375}},\ \bibinfo {pages} {315} (\bibinfo {year} {2022})}\BibitemShut {NoStop}%
	\bibitem [{\citenamefont {Rozenberg}\ \emph {et~al.}(2022)\citenamefont {Rozenberg}, \citenamefont {Karnieli}, \citenamefont {Yesharim}, \citenamefont {Foley-Comer}, \citenamefont {Trajtenberg-Mills}, \citenamefont {Freedman}, \citenamefont {Bronstein},\ and\ \citenamefont {Arie}}]{rozenberg2022inverse}%
	\BibitemOpen
	\bibfield  {author} {\bibinfo {author} {\bibfnamefont {E.}~\bibnamefont {Rozenberg}}, \bibinfo {author} {\bibfnamefont {A.}~\bibnamefont {Karnieli}}, \bibinfo {author} {\bibfnamefont {O.}~\bibnamefont {Yesharim}}, \bibinfo {author} {\bibfnamefont {J.}~\bibnamefont {Foley-Comer}}, \bibinfo {author} {\bibfnamefont {S.}~\bibnamefont {Trajtenberg-Mills}}, \bibinfo {author} {\bibfnamefont {D.}~\bibnamefont {Freedman}}, \bibinfo {author} {\bibfnamefont {A.~M.}\ \bibnamefont {Bronstein}},\ and\ \bibinfo {author} {\bibfnamefont {A.}~\bibnamefont {Arie}},\ }\bibfield  {title} {\bibinfo {title} {Inverse design of spontaneous parametric downconversion for generation of high-dimensional qudits},\ }\href@noop {} {\bibfield  {journal} {\bibinfo  {journal} {Optica}\ }\textbf {\bibinfo {volume} {9}},\ \bibinfo {pages} {602} (\bibinfo {year} {2022})}\BibitemShut {NoStop}%
	\bibitem [{\citenamefont {Roberts}\ \emph {et~al.}(2024)\citenamefont {Roberts}, \citenamefont {Wolley}, \citenamefont {Gregory},\ and\ \citenamefont {Padgett}}]{roberts2024comparison}%
	\BibitemOpen
	\bibfield  {author} {\bibinfo {author} {\bibfnamefont {K.}~\bibnamefont {Roberts}}, \bibinfo {author} {\bibfnamefont {O.}~\bibnamefont {Wolley}}, \bibinfo {author} {\bibfnamefont {T.}~\bibnamefont {Gregory}},\ and\ \bibinfo {author} {\bibfnamefont {M.}~\bibnamefont {Padgett}},\ }\bibfield  {title} {\bibinfo {title} {A comparison between the measurement of quantum spatial correlations using qcmos photon-number resolving and electron multiplying ccd camera technologies},\ }\href@noop {} {\bibfield  {journal} {\bibinfo  {journal} {Scientific Reports}\ }\textbf {\bibinfo {volume} {14}},\ \bibinfo {pages} {14687} (\bibinfo {year} {2024})}\BibitemShut {NoStop}%
	\bibitem [{\citenamefont {Resta}\ \emph {et~al.}(2023)\citenamefont {Resta}, \citenamefont {Stasi}, \citenamefont {Perrenoud}, \citenamefont {El-Khoury}, \citenamefont {Brydges}, \citenamefont {Thew}, \citenamefont {Zbinden},\ and\ \citenamefont {Bussi{\`e}res}}]{resta2023gigahertz}%
	\BibitemOpen
	\bibfield  {author} {\bibinfo {author} {\bibfnamefont {G.~V.}\ \bibnamefont {Resta}}, \bibinfo {author} {\bibfnamefont {L.}~\bibnamefont {Stasi}}, \bibinfo {author} {\bibfnamefont {M.}~\bibnamefont {Perrenoud}}, \bibinfo {author} {\bibfnamefont {S.}~\bibnamefont {El-Khoury}}, \bibinfo {author} {\bibfnamefont {T.}~\bibnamefont {Brydges}}, \bibinfo {author} {\bibfnamefont {R.}~\bibnamefont {Thew}}, \bibinfo {author} {\bibfnamefont {H.}~\bibnamefont {Zbinden}},\ and\ \bibinfo {author} {\bibfnamefont {F.}~\bibnamefont {Bussi{\`e}res}},\ }\bibfield  {title} {\bibinfo {title} {Gigahertz detection rates and dynamic photon-number resolution with superconducting nanowire arrays},\ }\href@noop {} {\bibfield  {journal} {\bibinfo  {journal} {Nano Letters}\ }\textbf {\bibinfo {volume} {23}},\ \bibinfo {pages} {6018} (\bibinfo {year} {2023})}\BibitemShut {NoStop}%
	\bibitem [{\citenamefont {Sloan}\ \emph {et~al.}(2025)\citenamefont {Sloan}, \citenamefont {Horodynski}, \citenamefont {Uddin}, \citenamefont {Salamin}, \citenamefont {Birk}, \citenamefont {Sidorenko}, \citenamefont {Kaminer}, \citenamefont {Solja{\v{c}}i{\'c}},\ and\ \citenamefont {Rivera}}]{sloan2025programmable}%
	\BibitemOpen
	\bibfield  {author} {\bibinfo {author} {\bibfnamefont {J.}~\bibnamefont {Sloan}}, \bibinfo {author} {\bibfnamefont {M.}~\bibnamefont {Horodynski}}, \bibinfo {author} {\bibfnamefont {S.~Z.}\ \bibnamefont {Uddin}}, \bibinfo {author} {\bibfnamefont {Y.}~\bibnamefont {Salamin}}, \bibinfo {author} {\bibfnamefont {M.}~\bibnamefont {Birk}}, \bibinfo {author} {\bibfnamefont {P.}~\bibnamefont {Sidorenko}}, \bibinfo {author} {\bibfnamefont {I.}~\bibnamefont {Kaminer}}, \bibinfo {author} {\bibfnamefont {M.}~\bibnamefont {Solja{\v{c}}i{\'c}}},\ and\ \bibinfo {author} {\bibfnamefont {N.}~\bibnamefont {Rivera}},\ }\bibfield  {title} {\bibinfo {title} {Programmable control of the spatiotemporal quantum noise of light},\ }\href@noop {} {\bibfield  {journal} {\bibinfo  {journal} {arXiv preprint arXiv:2509.03482}\ } (\bibinfo {year} {2025})}\BibitemShut {NoStop}%
	\bibitem [{\citenamefont {Yanagimoto}\ \emph {et~al.}(2025)\citenamefont {Yanagimoto}, \citenamefont {Ash}, \citenamefont {Sohoni}, \citenamefont {Stein}, \citenamefont {Zhao}, \citenamefont {Presutti}, \citenamefont {Jankowski}, \citenamefont {Wright}, \citenamefont {Onodera},\ and\ \citenamefont {McMahon}}]{yanagimoto2025programmable}%
	\BibitemOpen
	\bibfield  {author} {\bibinfo {author} {\bibfnamefont {R.}~\bibnamefont {Yanagimoto}}, \bibinfo {author} {\bibfnamefont {B.~A.}\ \bibnamefont {Ash}}, \bibinfo {author} {\bibfnamefont {M.~M.}\ \bibnamefont {Sohoni}}, \bibinfo {author} {\bibfnamefont {M.~M.}\ \bibnamefont {Stein}}, \bibinfo {author} {\bibfnamefont {Y.}~\bibnamefont {Zhao}}, \bibinfo {author} {\bibfnamefont {F.}~\bibnamefont {Presutti}}, \bibinfo {author} {\bibfnamefont {M.}~\bibnamefont {Jankowski}}, \bibinfo {author} {\bibfnamefont {L.~G.}\ \bibnamefont {Wright}}, \bibinfo {author} {\bibfnamefont {T.}~\bibnamefont {Onodera}},\ and\ \bibinfo {author} {\bibfnamefont {P.~L.}\ \bibnamefont {McMahon}},\ }\bibfield  {title} {\bibinfo {title} {Programmable on-chip nonlinear photonics},\ }\href@noop {} {\bibfield  {journal} {\bibinfo  {journal} {arXiv preprint arXiv:2503.19861}\ } (\bibinfo {year} {2025})}\BibitemShut {NoStop}%
	\bibitem [{\citenamefont {Figurnov}\ \emph {et~al.}(2018)\citenamefont {Figurnov}, \citenamefont {Mohamed},\ and\ \citenamefont {Mnih}}]{figurnov2018implicit}%
	\BibitemOpen
	\bibfield  {author} {\bibinfo {author} {\bibfnamefont {M.}~\bibnamefont {Figurnov}}, \bibinfo {author} {\bibfnamefont {S.}~\bibnamefont {Mohamed}},\ and\ \bibinfo {author} {\bibfnamefont {A.}~\bibnamefont {Mnih}},\ }\bibfield  {title} {\bibinfo {title} {Implicit reparameterization gradients},\ }\href@noop {} {\bibfield  {journal} {\bibinfo  {journal} {Advances in neural information processing systems}\ }\textbf {\bibinfo {volume} {31}} (\bibinfo {year} {2018})}\BibitemShut {NoStop}%
	\bibitem [{\citenamefont {Jankowiak}\ and\ \citenamefont {Karaletsos}(2019)}]{jankowiak2019pathwise}%
	\BibitemOpen
	\bibfield  {author} {\bibinfo {author} {\bibfnamefont {M.}~\bibnamefont {Jankowiak}}\ and\ \bibinfo {author} {\bibfnamefont {T.}~\bibnamefont {Karaletsos}},\ }\bibfield  {title} {\bibinfo {title} {Pathwise derivatives for multivariate distributions},\ }in\ \href@noop {} {\emph {\bibinfo {booktitle} {The 22nd International Conference on Artificial Intelligence and Statistics}}}\ (\bibinfo {organization} {PMLR},\ \bibinfo {year} {2019})\ pp.\ \bibinfo {pages} {333--342}\BibitemShut {NoStop}%
	\bibitem [{\citenamefont {Wernick}\ and\ \citenamefont {Morris}(1986)}]{wernick1986image}%
	\BibitemOpen
	\bibfield  {author} {\bibinfo {author} {\bibfnamefont {M.~N.}\ \bibnamefont {Wernick}}\ and\ \bibinfo {author} {\bibfnamefont {G.~M.}\ \bibnamefont {Morris}},\ }\bibfield  {title} {\bibinfo {title} {Image classification at low light levels},\ }\href@noop {} {\bibfield  {journal} {\bibinfo  {journal} {Journal of the Optical Society of America A}\ }\textbf {\bibinfo {volume} {3}},\ \bibinfo {pages} {2179} (\bibinfo {year} {1986})}\BibitemShut {NoStop}%
	\bibitem [{\citenamefont {Ota}\ \emph {et~al.}(2018)\citenamefont {Ota}, \citenamefont {Horisaki}, \citenamefont {Kawamura}, \citenamefont {Ugawa}, \citenamefont {Sato}, \citenamefont {Hashimoto}, \citenamefont {Kamesawa}, \citenamefont {Setoyama}, \citenamefont {Yamaguchi}, \citenamefont {Fujiu} \emph {et~al.}}]{ota2018ghost}%
	\BibitemOpen
	\bibfield  {author} {\bibinfo {author} {\bibfnamefont {S.}~\bibnamefont {Ota}}, \bibinfo {author} {\bibfnamefont {R.}~\bibnamefont {Horisaki}}, \bibinfo {author} {\bibfnamefont {Y.}~\bibnamefont {Kawamura}}, \bibinfo {author} {\bibfnamefont {M.}~\bibnamefont {Ugawa}}, \bibinfo {author} {\bibfnamefont {I.}~\bibnamefont {Sato}}, \bibinfo {author} {\bibfnamefont {K.}~\bibnamefont {Hashimoto}}, \bibinfo {author} {\bibfnamefont {R.}~\bibnamefont {Kamesawa}}, \bibinfo {author} {\bibfnamefont {K.}~\bibnamefont {Setoyama}}, \bibinfo {author} {\bibfnamefont {S.}~\bibnamefont {Yamaguchi}}, \bibinfo {author} {\bibfnamefont {K.}~\bibnamefont {Fujiu}}, \emph {et~al.},\ }\bibfield  {title} {\bibinfo {title} {Ghost cytometry},\ }\href@noop {} {\bibfield  {journal} {\bibinfo  {journal} {Science}\ }\textbf {\bibinfo {volume} {360}},\ \bibinfo {pages} {1246} (\bibinfo {year} {2018})}\BibitemShut {NoStop}%
	\bibitem [{\citenamefont {Goyal}\ and\ \citenamefont {Gupta}(2021)}]{goyal2021photon}%
	\BibitemOpen
	\bibfield  {author} {\bibinfo {author} {\bibfnamefont {B.}~\bibnamefont {Goyal}}\ and\ \bibinfo {author} {\bibfnamefont {M.}~\bibnamefont {Gupta}},\ }\bibfield  {title} {\bibinfo {title} {Photon-starved scene inference using single photon cameras},\ }in\ \href@noop {} {\emph {\bibinfo {booktitle} {Proceedings of the IEEE/CVF International Conference on Computer Vision}}}\ (\bibinfo {year} {2021})\ pp.\ \bibinfo {pages} {2512--2521}\BibitemShut {NoStop}%
	\bibitem [{\citenamefont {Minati}\ \emph {et~al.}(2026)\citenamefont {Minati}, \citenamefont {Roncallo}, \citenamefont {Scrofana}, \citenamefont {Morgillo}, \citenamefont {Spagnolo}, \citenamefont {Macchiavello}, \citenamefont {Maccone}, \citenamefont {Cimini},\ and\ \citenamefont {Sciarrino}}]{minati2026quantum}%
	\BibitemOpen
	\bibfield  {author} {\bibinfo {author} {\bibfnamefont {G.}~\bibnamefont {Minati}}, \bibinfo {author} {\bibfnamefont {S.}~\bibnamefont {Roncallo}}, \bibinfo {author} {\bibfnamefont {S.}~\bibnamefont {Scrofana}}, \bibinfo {author} {\bibfnamefont {A.~R.}\ \bibnamefont {Morgillo}}, \bibinfo {author} {\bibfnamefont {N.}~\bibnamefont {Spagnolo}}, \bibinfo {author} {\bibfnamefont {C.}~\bibnamefont {Macchiavello}}, \bibinfo {author} {\bibfnamefont {L.}~\bibnamefont {Maccone}}, \bibinfo {author} {\bibfnamefont {V.}~\bibnamefont {Cimini}},\ and\ \bibinfo {author} {\bibfnamefont {F.}~\bibnamefont {Sciarrino}},\ }\bibfield  {title} {\bibinfo {title} {Quantum optical neuron for image classification via multiphoton interference},\ }\href@noop {} {\bibfield  {journal} {\bibinfo  {journal} {arXiv preprint arXiv:2603.28879}\ } (\bibinfo {year} {2026})}\BibitemShut {NoStop}%
	\bibitem [{\citenamefont {Saaf}\ and\ \citenamefont {Morris}(1995)}]{saaf1995photon}%
	\BibitemOpen
	\bibfield  {author} {\bibinfo {author} {\bibfnamefont {L.~A.}\ \bibnamefont {Saaf}}\ and\ \bibinfo {author} {\bibfnamefont {G.~M.}\ \bibnamefont {Morris}},\ }\bibfield  {title} {\bibinfo {title} {Photon-limited image classification with a feedforward neural network},\ }\href@noop {} {\bibfield  {journal} {\bibinfo  {journal} {Applied optics}\ }\textbf {\bibinfo {volume} {34}},\ \bibinfo {pages} {3963} (\bibinfo {year} {1995})}\BibitemShut {NoStop}%
	\bibitem [{\citenamefont {Zhu}\ \emph {et~al.}(2020)\citenamefont {Zhu}, \citenamefont {Shi}, \citenamefont {Wu}, \citenamefont {Liu}, \citenamefont {Zeng}, \citenamefont {Sun}, \citenamefont {Tian},\ and\ \citenamefont {Su}}]{zhu2020photon}%
	\BibitemOpen
	\bibfield  {author} {\bibinfo {author} {\bibfnamefont {Y.}~\bibnamefont {Zhu}}, \bibinfo {author} {\bibfnamefont {J.}~\bibnamefont {Shi}}, \bibinfo {author} {\bibfnamefont {X.}~\bibnamefont {Wu}}, \bibinfo {author} {\bibfnamefont {X.}~\bibnamefont {Liu}}, \bibinfo {author} {\bibfnamefont {G.}~\bibnamefont {Zeng}}, \bibinfo {author} {\bibfnamefont {J.}~\bibnamefont {Sun}}, \bibinfo {author} {\bibfnamefont {L.}~\bibnamefont {Tian}},\ and\ \bibinfo {author} {\bibfnamefont {F.}~\bibnamefont {Su}},\ }\bibfield  {title} {\bibinfo {title} {Photon-limited non-imaging object detection and classification based on single-pixel imaging system.},\ }\href@noop {} {\bibfield  {journal} {\bibinfo  {journal} {Applied Physics B: Lasers \& Optics}\ }\textbf {\bibinfo {volume} {126}} (\bibinfo {year} {2020})}\BibitemShut {NoStop}%
	\bibitem [{\citenamefont {Chen}\ and\ \citenamefont {Perona}(2016)}]{chen2016vision}%
	\BibitemOpen
	\bibfield  {author} {\bibinfo {author} {\bibfnamefont {B.}~\bibnamefont {Chen}}\ and\ \bibinfo {author} {\bibfnamefont {P.}~\bibnamefont {Perona}},\ }\bibfield  {title} {\bibinfo {title} {Vision without the image},\ }\href@noop {} {\bibfield  {journal} {\bibinfo  {journal} {Sensors}\ }\textbf {\bibinfo {volume} {16}},\ \bibinfo {pages} {484} (\bibinfo {year} {2016})}\BibitemShut {NoStop}%
	\bibitem [{\citenamefont {Ma}\ \emph {et~al.}(2026)\citenamefont {Ma}, \citenamefont {Laydevant}, \citenamefont {Sohoni}, \citenamefont {Wright}, \citenamefont {Wang},\ and\ \citenamefont {McMahon}}]{ma2026machine}%
	\BibitemOpen
	\bibfield  {author} {\bibinfo {author} {\bibfnamefont {S.-Y.}\ \bibnamefont {Ma}}, \bibinfo {author} {\bibfnamefont {J.}~\bibnamefont {Laydevant}}, \bibinfo {author} {\bibfnamefont {M.~M.}\ \bibnamefont {Sohoni}}, \bibinfo {author} {\bibfnamefont {L.~G.}\ \bibnamefont {Wright}}, \bibinfo {author} {\bibfnamefont {T.}~\bibnamefont {Wang}},\ and\ \bibinfo {author} {\bibfnamefont {P.~L.}\ \bibnamefont {McMahon}},\ }\bibfield  {title} {\bibinfo {title} {Machine vision with small numbers of detected photons per inference},\ }\href@noop {} {\bibfield  {journal} {\bibinfo  {journal} {arXiv preprint arXiv:2603.23974}\ } (\bibinfo {year} {2026})}\BibitemShut {NoStop}%
	\bibitem [{\citenamefont {Zhang}\ \emph {et~al.}(2015)\citenamefont {Zhang}, \citenamefont {Mouradian}, \citenamefont {Wong},\ and\ \citenamefont {Shapiro}}]{zhang2015entanglement}%
	\BibitemOpen
	\bibfield  {author} {\bibinfo {author} {\bibfnamefont {Z.}~\bibnamefont {Zhang}}, \bibinfo {author} {\bibfnamefont {S.}~\bibnamefont {Mouradian}}, \bibinfo {author} {\bibfnamefont {F.~N.}\ \bibnamefont {Wong}},\ and\ \bibinfo {author} {\bibfnamefont {J.~H.}\ \bibnamefont {Shapiro}},\ }\bibfield  {title} {\bibinfo {title} {Entanglement-enhanced sensing in a lossy and noisy environment},\ }\href@noop {} {\bibfield  {journal} {\bibinfo  {journal} {Physical review letters}\ }\textbf {\bibinfo {volume} {114}},\ \bibinfo {pages} {110506} (\bibinfo {year} {2015})}\BibitemShut {NoStop}%
	\bibitem [{\citenamefont {Sarovar}(2023)}]{sarovar2023quantum}%
	\BibitemOpen
	\bibfield  {author} {\bibinfo {author} {\bibfnamefont {M.}~\bibnamefont {Sarovar}},\ }\bibfield  {title} {\bibinfo {title} {Quantum computational imaging and sensing},\ }in\ \href@noop {} {\emph {\bibinfo {booktitle} {Quantum Nanophotonic Materials, Devices, and Systems 2023}}},\ Vol.\ \bibinfo {volume} {12657}\ (\bibinfo {organization} {SPIE},\ \bibinfo {year} {2023})\ pp.\ \bibinfo {pages} {6--9}\BibitemShut {NoStop}%
	\bibitem [{\citenamefont {Khan}\ \emph {et~al.}(2025)\citenamefont {Khan}, \citenamefont {Prabhu}, \citenamefont {Wright},\ and\ \citenamefont {McMahon}}]{khan2025quantum}%
	\BibitemOpen
	\bibfield  {author} {\bibinfo {author} {\bibfnamefont {S.~A.}\ \bibnamefont {Khan}}, \bibinfo {author} {\bibfnamefont {S.}~\bibnamefont {Prabhu}}, \bibinfo {author} {\bibfnamefont {L.~G.}\ \bibnamefont {Wright}},\ and\ \bibinfo {author} {\bibfnamefont {P.~L.}\ \bibnamefont {McMahon}},\ }\bibfield  {title} {\bibinfo {title} {Quantum computational-sensing advantage},\ }\href@noop {} {\bibfield  {journal} {\bibinfo  {journal} {arXiv preprint arXiv:2507.16918}\ } (\bibinfo {year} {2025})}\BibitemShut {NoStop}%
	\bibitem [{\citenamefont {Weedbrook}\ \emph {et~al.}(2016)\citenamefont {Weedbrook}, \citenamefont {Pirandola}, \citenamefont {Thompson}, \citenamefont {Vedral},\ and\ \citenamefont {Gu}}]{weedbrook2016discord}%
	\BibitemOpen
	\bibfield  {author} {\bibinfo {author} {\bibfnamefont {C.}~\bibnamefont {Weedbrook}}, \bibinfo {author} {\bibfnamefont {S.}~\bibnamefont {Pirandola}}, \bibinfo {author} {\bibfnamefont {J.}~\bibnamefont {Thompson}}, \bibinfo {author} {\bibfnamefont {V.}~\bibnamefont {Vedral}},\ and\ \bibinfo {author} {\bibfnamefont {M.}~\bibnamefont {Gu}},\ }\bibfield  {title} {\bibinfo {title} {How discord underlies the noise resilience of quantum illumination},\ }\href@noop {} {\bibfield  {journal} {\bibinfo  {journal} {New Journal of Physics}\ }\textbf {\bibinfo {volume} {18}},\ \bibinfo {pages} {043027} (\bibinfo {year} {2016})}\BibitemShut {NoStop}%
	\bibitem [{\citenamefont {Couteau}(2018)}]{couteau2018spontaneous}%
	\BibitemOpen
	\bibfield  {author} {\bibinfo {author} {\bibfnamefont {C.}~\bibnamefont {Couteau}},\ }\bibfield  {title} {\bibinfo {title} {Spontaneous parametric down-conversion},\ }\href@noop {} {\bibfield  {journal} {\bibinfo  {journal} {Contemporary Physics}\ }\textbf {\bibinfo {volume} {59}},\ \bibinfo {pages} {291} (\bibinfo {year} {2018})}\BibitemShut {NoStop}%
	\bibitem [{\citenamefont {Ono}\ \emph {et~al.}(2013)\citenamefont {Ono}, \citenamefont {Okamoto},\ and\ \citenamefont {Takeuchi}}]{ono2013entanglement}%
	\BibitemOpen
	\bibfield  {author} {\bibinfo {author} {\bibfnamefont {T.}~\bibnamefont {Ono}}, \bibinfo {author} {\bibfnamefont {R.}~\bibnamefont {Okamoto}},\ and\ \bibinfo {author} {\bibfnamefont {S.}~\bibnamefont {Takeuchi}},\ }\bibfield  {title} {\bibinfo {title} {An entanglement-enhanced microscope},\ }\href@noop {} {\bibfield  {journal} {\bibinfo  {journal} {Nature communications}\ }\textbf {\bibinfo {volume} {4}},\ \bibinfo {pages} {2426} (\bibinfo {year} {2013})}\BibitemShut {NoStop}%
	\bibitem [{\citenamefont {Defienne}\ \emph {et~al.}(2021)\citenamefont {Defienne}, \citenamefont {Ndagano}, \citenamefont {Lyons},\ and\ \citenamefont {Faccio}}]{defienne2021polarization}%
	\BibitemOpen
	\bibfield  {author} {\bibinfo {author} {\bibfnamefont {H.}~\bibnamefont {Defienne}}, \bibinfo {author} {\bibfnamefont {B.}~\bibnamefont {Ndagano}}, \bibinfo {author} {\bibfnamefont {A.}~\bibnamefont {Lyons}},\ and\ \bibinfo {author} {\bibfnamefont {D.}~\bibnamefont {Faccio}},\ }\bibfield  {title} {\bibinfo {title} {Polarization entanglement-enabled quantum holography},\ }\href@noop {} {\bibfield  {journal} {\bibinfo  {journal} {Nature Physics}\ }\textbf {\bibinfo {volume} {17}},\ \bibinfo {pages} {591} (\bibinfo {year} {2021})}\BibitemShut {NoStop}%
	\bibitem [{\citenamefont {Hu}\ \emph {et~al.}(2023)\citenamefont {Hu}, \citenamefont {Angelatos}, \citenamefont {Khan}, \citenamefont {Vives}, \citenamefont {T{\"u}reci}, \citenamefont {Bello}, \citenamefont {Rowlands}, \citenamefont {Ribeill},\ and\ \citenamefont {T{\"u}reci}}]{hu2023tackling}%
	\BibitemOpen
	\bibfield  {author} {\bibinfo {author} {\bibfnamefont {F.}~\bibnamefont {Hu}}, \bibinfo {author} {\bibfnamefont {G.}~\bibnamefont {Angelatos}}, \bibinfo {author} {\bibfnamefont {S.~A.}\ \bibnamefont {Khan}}, \bibinfo {author} {\bibfnamefont {M.}~\bibnamefont {Vives}}, \bibinfo {author} {\bibfnamefont {E.}~\bibnamefont {T{\"u}reci}}, \bibinfo {author} {\bibfnamefont {L.}~\bibnamefont {Bello}}, \bibinfo {author} {\bibfnamefont {G.~E.}\ \bibnamefont {Rowlands}}, \bibinfo {author} {\bibfnamefont {G.~J.}\ \bibnamefont {Ribeill}},\ and\ \bibinfo {author} {\bibfnamefont {H.~E.}\ \bibnamefont {T{\"u}reci}},\ }\bibfield  {title} {\bibinfo {title} {Tackling sampling noise in physical systems for machine learning applications: Fundamental limits and eigentasks},\ }\href@noop {} {\bibfield  {journal} {\bibinfo  {journal} {Physical Review X}\ }\textbf {\bibinfo {volume} {13}},\ \bibinfo {pages} {041020} (\bibinfo {year} {2023})}\BibitemShut {NoStop}%
	\bibitem [{\citenamefont {Choi}\ and\ \citenamefont {Majumdar}(2025)}]{choi2025free}%
	\BibitemOpen
	\bibfield  {author} {\bibinfo {author} {\bibfnamefont {M.}~\bibnamefont {Choi}}\ and\ \bibinfo {author} {\bibfnamefont {A.}~\bibnamefont {Majumdar}},\ }\bibfield  {title} {\bibinfo {title} {Free-space optical encoder for computer vision},\ }\href@noop {} {\bibfield  {journal} {\bibinfo  {journal} {npj Nanophotonics}\ }\textbf {\bibinfo {volume} {2}},\ \bibinfo {pages} {36} (\bibinfo {year} {2025})}\BibitemShut {NoStop}%
	\bibitem [{\citenamefont {Steinbrecher}\ \emph {et~al.}(2019)\citenamefont {Steinbrecher}, \citenamefont {Olson}, \citenamefont {Englund},\ and\ \citenamefont {Carolan}}]{steinbrecher2019quantum}%
	\BibitemOpen
	\bibfield  {author} {\bibinfo {author} {\bibfnamefont {G.~R.}\ \bibnamefont {Steinbrecher}}, \bibinfo {author} {\bibfnamefont {J.~P.}\ \bibnamefont {Olson}}, \bibinfo {author} {\bibfnamefont {D.}~\bibnamefont {Englund}},\ and\ \bibinfo {author} {\bibfnamefont {J.}~\bibnamefont {Carolan}},\ }\bibfield  {title} {\bibinfo {title} {Quantum optical neural networks},\ }\href@noop {} {\bibfield  {journal} {\bibinfo  {journal} {npj Quantum Information}\ }\textbf {\bibinfo {volume} {5}},\ \bibinfo {pages} {60} (\bibinfo {year} {2019})}\BibitemShut {NoStop}%
	\bibitem [{\citenamefont {Avella}\ \emph {et~al.}(2016)\citenamefont {Avella}, \citenamefont {Ruo-Berchera}, \citenamefont {Degiovanni}, \citenamefont {Brida},\ and\ \citenamefont {Genovese}}]{avella2016absolute}%
	\BibitemOpen
	\bibfield  {author} {\bibinfo {author} {\bibfnamefont {A.}~\bibnamefont {Avella}}, \bibinfo {author} {\bibfnamefont {I.}~\bibnamefont {Ruo-Berchera}}, \bibinfo {author} {\bibfnamefont {I.~P.}\ \bibnamefont {Degiovanni}}, \bibinfo {author} {\bibfnamefont {G.}~\bibnamefont {Brida}},\ and\ \bibinfo {author} {\bibfnamefont {M.}~\bibnamefont {Genovese}},\ }\bibfield  {title} {\bibinfo {title} {Absolute calibration of an emccd camera by quantum correlation, linking photon counting to the analog regime},\ }\href@noop {} {\bibfield  {journal} {\bibinfo  {journal} {Optics letters}\ }\textbf {\bibinfo {volume} {41}},\ \bibinfo {pages} {1841} (\bibinfo {year} {2016})}\BibitemShut {NoStop}%
\end{thebibliography}
\end{document}